\documentclass[10pt,twocolumn,letterpaper]{article}
\usepackage[pagenumbers]{cvpr}
\usepackage[T1]{fontenc}
\usepackage{amsthm,mathtools,array,multirow,tabularx}
\usepackage{algorithm,algorithmic}
\usepackage{tikz}
\usetikzlibrary{arrows.meta,positioning,fit,calc}
\usepackage{placeins}
\usepackage{microtype}
\newtheorem{proposition}{Proposition}
\newtheorem{lemma}{Lemma}
\definecolor{capeteal}{HTML}{176B70}
\definecolor{capeamber}{HTML}{A86322}
\definecolor{capeblue}{HTML}{456D9A}
\newcommand{\calD}{\mathcal D}
\newcommand{\calF}{\mathcal F}

\newcommand{\supp}{Supplementary Material}
\newsavebox{\tablebox}
\newcommand{\fitinput}[1]{%
  \sbox{\tablebox}{#1}%
  \ifdim\wd\tablebox>\linewidth
    \resizebox{\linewidth}{!}{\usebox{\tablebox}}%
  \else\usebox{\tablebox}\fi}
\newcommand{\JVal}[1]{\csname joint@#1\endcsname}
\expandafter\def\csname joint@csv/3/partial_native/success_rate\endcsname{85.16}
\expandafter\def\csname joint@csv/3/partial_native/calls\endcsname{2.196}
\expandafter\def\csname joint@csv/3/partial_native/queries\endcsname{2980.5}
\expandafter\def\csname joint@csv/3/partial_native/lifecycle_seconds\endcsname{10.774}
\expandafter\def\csname joint@csv/3/partial_native/certified\endcsname{98.44}
\expandafter\def\csname joint@csv/3/partial_native/optimal\endcsname{100.00}
\expandafter\def\csname joint@csv/3/partial_native/initial_success_rate\endcsname{69.97}
\expandafter\def\csname joint@csv/3/partial_native/initial_calls\endcsname{2.113}
\expandafter\def\csname joint@csv/3/eager_optimized/success_rate\endcsname{85.16}
\expandafter\def\csname joint@csv/3/eager_optimized/calls\endcsname{2.196}
\expandafter\def\csname joint@csv/3/eager_optimized/queries\endcsname{3450.9}
\expandafter\def\csname joint@csv/3/eager_optimized/lifecycle_seconds\endcsname{11.173}
\expandafter\def\csname joint@csv/3/eager_optimized/certified\endcsname{98.44}
\expandafter\def\csname joint@csv/3/eager_optimized/optimal\endcsname{100.00}
\expandafter\def\csname joint@csv/3/eager_optimized/initial_success_rate\endcsname{69.97}
\expandafter\def\csname joint@csv/3/eager_optimized/initial_calls\endcsname{2.113}
\expandafter\def\csname joint@csv/3/native_program_only/success_rate\endcsname{76.76}
\expandafter\def\csname joint@csv/3/native_program_only/calls\endcsname{2.022}
\expandafter\def\csname joint@csv/3/native_program_only/queries\endcsname{624.0}
\expandafter\def\csname joint@csv/3/native_program_only/lifecycle_seconds\endcsname{3.631}
\expandafter\def\csname joint@csv/3/native_program_only/certified\endcsname{34.38}
\expandafter\def\csname joint@csv/3/native_program_only/optimal\endcsname{35.04}
\expandafter\def\csname joint@csv/3/native_program_only/initial_success_rate\endcsname{69.97}
\expandafter\def\csname joint@csv/3/native_program_only/initial_calls\endcsname{2.113}
\expandafter\def\csname joint@csv/3/initial_safe_two_step/success_rate\endcsname{83.98}
\expandafter\def\csname joint@csv/3/initial_safe_two_step/calls\endcsname{2.474}
\expandafter\def\csname joint@csv/3/initial_safe_two_step/queries\endcsname{3450.9}
\expandafter\def\csname joint@csv/3/initial_safe_two_step/lifecycle_seconds\endcsname{11.121}
\expandafter\def\csname joint@csv/3/initial_safe_two_step/certified\endcsname{34.38}
\expandafter\def\csname joint@csv/3/initial_safe_two_step/optimal\endcsname{35.04}
\expandafter\def\csname joint@csv/3/initial_safe_two_step/initial_success_rate\endcsname{69.97}
\expandafter\def\csname joint@csv/3/initial_safe_two_step/initial_calls\endcsname{2.113}
\expandafter\def\csname joint@csv/3/lazy_quotient/success_rate\endcsname{83.74}
\expandafter\def\csname joint@csv/3/lazy_quotient/calls\endcsname{2.198}
\expandafter\def\csname joint@csv/3/lazy_quotient/queries\endcsname{3332.5}
\expandafter\def\csname joint@csv/3/lazy_quotient/lifecycle_seconds\endcsname{17.089}
\expandafter\def\csname joint@csv/3/lazy_quotient/certified\endcsname{93.75}
\expandafter\def\csname joint@csv/3/lazy_quotient/optimal\endcsname{95.31}
\expandafter\def\csname joint@csv/3/lazy_quotient/initial_success_rate\endcsname{69.97}
\expandafter\def\csname joint@csv/3/lazy_quotient/initial_calls\endcsname{2.113}
\expandafter\def\csname joint@csv/3/native_frontier/success_rate\endcsname{85.16}
\expandafter\def\csname joint@csv/3/native_frontier/calls\endcsname{2.196}
\expandafter\def\csname joint@csv/3/native_frontier/queries\endcsname{3307.3}
\expandafter\def\csname joint@csv/3/native_frontier/lifecycle_seconds\endcsname{11.402}
\expandafter\def\csname joint@csv/3/native_frontier/certified\endcsname{98.44}
\expandafter\def\csname joint@csv/3/native_frontier/optimal\endcsname{100.00}
\expandafter\def\csname joint@csv/3/native_frontier/initial_success_rate\endcsname{69.97}
\expandafter\def\csname joint@csv/3/native_frontier/initial_calls\endcsname{2.113}
\expandafter\def\csname joint@csv/3/partial_native_full_rows/success_rate\endcsname{85.16}
\expandafter\def\csname joint@csv/3/partial_native_full_rows/calls\endcsname{2.196}
\expandafter\def\csname joint@csv/3/partial_native_full_rows/queries\endcsname{3243.6}
\expandafter\def\csname joint@csv/3/partial_native_full_rows/lifecycle_seconds\endcsname{11.311}
\expandafter\def\csname joint@csv/3/partial_native_full_rows/certified\endcsname{98.44}
\expandafter\def\csname joint@csv/3/partial_native_full_rows/optimal\endcsname{100.00}
\expandafter\def\csname joint@csv/3/partial_native_full_rows/initial_success_rate\endcsname{69.97}
\expandafter\def\csname joint@csv/3/partial_native_full_rows/initial_calls\endcsname{2.113}
\expandafter\def\csname joint@csv/3/partial_native_no_tightening/success_rate\endcsname{85.16}
\expandafter\def\csname joint@csv/3/partial_native_no_tightening/calls\endcsname{2.196}
\expandafter\def\csname joint@csv/3/partial_native_no_tightening/queries\endcsname{3330.3}
\expandafter\def\csname joint@csv/3/partial_native_no_tightening/lifecycle_seconds\endcsname{11.374}
\expandafter\def\csname joint@csv/3/partial_native_no_tightening/certified\endcsname{98.44}
\expandafter\def\csname joint@csv/3/partial_native_no_tightening/optimal\endcsname{100.00}
\expandafter\def\csname joint@csv/3/partial_native_no_tightening/initial_success_rate\endcsname{69.97}
\expandafter\def\csname joint@csv/3/partial_native_no_tightening/initial_calls\endcsname{2.113}
\expandafter\def\csname joint@csv/5/partial_native/success_rate\endcsname{98.44}
\expandafter\def\csname joint@csv/5/partial_native/calls\endcsname{2.348}
\expandafter\def\csname joint@csv/5/partial_native/queries\endcsname{3161.1}
\expandafter\def\csname joint@csv/5/partial_native/lifecycle_seconds\endcsname{11.201}
\expandafter\def\csname joint@csv/5/partial_native/certified\endcsname{98.44}
\expandafter\def\csname joint@csv/5/partial_native/optimal\endcsname{100.00}
\expandafter\def\csname joint@csv/5/partial_native/initial_success_rate\endcsname{83.01}
\expandafter\def\csname joint@csv/5/partial_native/initial_calls\endcsname{2.605}
\expandafter\def\csname joint@csv/5/eager_optimized/success_rate\endcsname{98.44}
\expandafter\def\csname joint@csv/5/eager_optimized/calls\endcsname{2.348}
\expandafter\def\csname joint@csv/5/eager_optimized/queries\endcsname{3450.9}
\expandafter\def\csname joint@csv/5/eager_optimized/lifecycle_seconds\endcsname{11.417}
\expandafter\def\csname joint@csv/5/eager_optimized/certified\endcsname{98.44}
\expandafter\def\csname joint@csv/5/eager_optimized/optimal\endcsname{100.00}
\expandafter\def\csname joint@csv/5/eager_optimized/initial_success_rate\endcsname{83.01}
\expandafter\def\csname joint@csv/5/eager_optimized/initial_calls\endcsname{2.605}
\expandafter\def\csname joint@csv/5/native_program_only/success_rate\endcsname{92.19}
\expandafter\def\csname joint@csv/5/native_program_only/calls\endcsname{2.365}
\expandafter\def\csname joint@csv/5/native_program_only/queries\endcsname{624.0}
\expandafter\def\csname joint@csv/5/native_program_only/lifecycle_seconds\endcsname{3.624}
\expandafter\def\csname joint@csv/5/native_program_only/certified\endcsname{34.38}
\expandafter\def\csname joint@csv/5/native_program_only/optimal\endcsname{75.00}
\expandafter\def\csname joint@csv/5/native_program_only/initial_success_rate\endcsname{83.01}
\expandafter\def\csname joint@csv/5/native_program_only/initial_calls\endcsname{2.605}
\expandafter\def\csname joint@csv/5/initial_safe_two_step/success_rate\endcsname{98.44}
\expandafter\def\csname joint@csv/5/initial_safe_two_step/calls\endcsname{2.838}
\expandafter\def\csname joint@csv/5/initial_safe_two_step/queries\endcsname{3450.9}
\expandafter\def\csname joint@csv/5/initial_safe_two_step/lifecycle_seconds\endcsname{11.192}
\expandafter\def\csname joint@csv/5/initial_safe_two_step/certified\endcsname{34.38}
\expandafter\def\csname joint@csv/5/initial_safe_two_step/optimal\endcsname{70.31}
\expandafter\def\csname joint@csv/5/initial_safe_two_step/initial_success_rate\endcsname{83.01}
\expandafter\def\csname joint@csv/5/initial_safe_two_step/initial_calls\endcsname{2.605}
\expandafter\def\csname joint@csv/5/lazy_quotient/success_rate\endcsname{84.18}
\expandafter\def\csname joint@csv/5/lazy_quotient/calls\endcsname{2.534}
\expandafter\def\csname joint@csv/5/lazy_quotient/queries\endcsname{2382.0}
\expandafter\def\csname joint@csv/5/lazy_quotient/lifecycle_seconds\endcsname{13.700}
\expandafter\def\csname joint@csv/5/lazy_quotient/certified\endcsname{73.44}
\expandafter\def\csname joint@csv/5/lazy_quotient/optimal\endcsname{75.00}
\expandafter\def\csname joint@csv/5/lazy_quotient/initial_success_rate\endcsname{83.01}
\expandafter\def\csname joint@csv/5/lazy_quotient/initial_calls\endcsname{2.605}
\expandafter\def\csname joint@csv/5/native_frontier/success_rate\endcsname{98.44}
\expandafter\def\csname joint@csv/5/native_frontier/calls\endcsname{2.348}
\expandafter\def\csname joint@csv/5/native_frontier/queries\endcsname{3191.6}
\expandafter\def\csname joint@csv/5/native_frontier/lifecycle_seconds\endcsname{11.572}
\expandafter\def\csname joint@csv/5/native_frontier/certified\endcsname{98.44}
\expandafter\def\csname joint@csv/5/native_frontier/optimal\endcsname{100.00}
\expandafter\def\csname joint@csv/5/native_frontier/initial_success_rate\endcsname{83.01}
\expandafter\def\csname joint@csv/5/native_frontier/initial_calls\endcsname{2.605}
\expandafter\def\csname joint@csv/5/partial_native_full_rows/success_rate\endcsname{98.44}
\expandafter\def\csname joint@csv/5/partial_native_full_rows/calls\endcsname{2.348}
\expandafter\def\csname joint@csv/5/partial_native_full_rows/queries\endcsname{3161.0}
\expandafter\def\csname joint@csv/5/partial_native_full_rows/lifecycle_seconds\endcsname{11.177}
\expandafter\def\csname joint@csv/5/partial_native_full_rows/certified\endcsname{98.44}
\expandafter\def\csname joint@csv/5/partial_native_full_rows/optimal\endcsname{100.00}
\expandafter\def\csname joint@csv/5/partial_native_full_rows/initial_success_rate\endcsname{83.01}
\expandafter\def\csname joint@csv/5/partial_native_full_rows/initial_calls\endcsname{2.605}
\expandafter\def\csname joint@csv/5/partial_native_no_tightening/success_rate\endcsname{98.44}
\expandafter\def\csname joint@csv/5/partial_native_no_tightening/calls\endcsname{2.348}
\expandafter\def\csname joint@csv/5/partial_native_no_tightening/queries\endcsname{3161.1}
\expandafter\def\csname joint@csv/5/partial_native_no_tightening/lifecycle_seconds\endcsname{11.217}
\expandafter\def\csname joint@csv/5/partial_native_no_tightening/certified\endcsname{98.44}
\expandafter\def\csname joint@csv/5/partial_native_no_tightening/optimal\endcsname{100.00}
\expandafter\def\csname joint@csv/5/partial_native_no_tightening/initial_success_rate\endcsname{83.01}
\expandafter\def\csname joint@csv/5/partial_native_no_tightening/initial_calls\endcsname{2.605}
\expandafter\def\csname joint@json/3/partial_native/success_rate\endcsname{94.74}
\expandafter\def\csname joint@json/3/partial_native/calls\endcsname{1.476}
\expandafter\def\csname joint@json/3/partial_native/queries\endcsname{1506.2}
\expandafter\def\csname joint@json/3/partial_native/lifecycle_seconds\endcsname{11.919}
\expandafter\def\csname joint@json/3/partial_native/certified\endcsname{95.31}
\expandafter\def\csname joint@json/3/partial_native/optimal\endcsname{98.44}
\expandafter\def\csname joint@json/3/partial_native/initial_success_rate\endcsname{60.09}
\expandafter\def\csname joint@json/3/partial_native/initial_calls\endcsname{1.957}
\expandafter\def\csname joint@json/3/eager_optimized/success_rate\endcsname{94.60}
\expandafter\def\csname joint@json/3/eager_optimized/calls\endcsname{1.474}
\expandafter\def\csname joint@json/3/eager_optimized/queries\endcsname{2309.0}
\expandafter\def\csname joint@json/3/eager_optimized/lifecycle_seconds\endcsname{18.021}
\expandafter\def\csname joint@json/3/eager_optimized/certified\endcsname{84.38}
\expandafter\def\csname joint@json/3/eager_optimized/optimal\endcsname{98.44}
\expandafter\def\csname joint@json/3/eager_optimized/initial_success_rate\endcsname{60.09}
\expandafter\def\csname joint@json/3/eager_optimized/initial_calls\endcsname{1.957}
\expandafter\def\csname joint@json/3/native_program_only/success_rate\endcsname{94.60}
\expandafter\def\csname joint@json/3/native_program_only/calls\endcsname{1.474}
\expandafter\def\csname joint@json/3/native_program_only/queries\endcsname{2075.7}
\expandafter\def\csname joint@json/3/native_program_only/lifecycle_seconds\endcsname{17.792}
\expandafter\def\csname joint@json/3/native_program_only/certified\endcsname{77.34}
\expandafter\def\csname joint@json/3/native_program_only/optimal\endcsname{98.44}
\expandafter\def\csname joint@json/3/native_program_only/initial_success_rate\endcsname{60.09}
\expandafter\def\csname joint@json/3/native_program_only/initial_calls\endcsname{1.957}
\expandafter\def\csname joint@json/3/initial_safe_two_step/success_rate\endcsname{94.60}
\expandafter\def\csname joint@json/3/initial_safe_two_step/calls\endcsname{1.493}
\expandafter\def\csname joint@json/3/initial_safe_two_step/queries\endcsname{2307.2}
\expandafter\def\csname joint@json/3/initial_safe_two_step/lifecycle_seconds\endcsname{18.008}
\expandafter\def\csname joint@json/3/initial_safe_two_step/certified\endcsname{77.34}
\expandafter\def\csname joint@json/3/initial_safe_two_step/optimal\endcsname{91.96}
\expandafter\def\csname joint@json/3/initial_safe_two_step/initial_success_rate\endcsname{60.09}
\expandafter\def\csname joint@json/3/initial_safe_two_step/initial_calls\endcsname{1.957}
\expandafter\def\csname joint@json/3/lazy_quotient/success_rate\endcsname{92.90}
\expandafter\def\csname joint@json/3/lazy_quotient/calls\endcsname{1.495}
\expandafter\def\csname joint@json/3/lazy_quotient/queries\endcsname{2029.5}
\expandafter\def\csname joint@json/3/lazy_quotient/lifecycle_seconds\endcsname{16.200}
\expandafter\def\csname joint@json/3/lazy_quotient/certified\endcsname{93.75}
\expandafter\def\csname joint@json/3/lazy_quotient/optimal\endcsname{95.31}
\expandafter\def\csname joint@json/3/lazy_quotient/initial_success_rate\endcsname{60.09}
\expandafter\def\csname joint@json/3/lazy_quotient/initial_calls\endcsname{1.957}
\expandafter\def\csname joint@json/3/native_frontier/success_rate\endcsname{94.60}
\expandafter\def\csname joint@json/3/native_frontier/calls\endcsname{1.474}
\expandafter\def\csname joint@json/3/native_frontier/queries\endcsname{2170.0}
\expandafter\def\csname joint@json/3/native_frontier/lifecycle_seconds\endcsname{17.878}
\expandafter\def\csname joint@json/3/native_frontier/certified\endcsname{85.94}
\expandafter\def\csname joint@json/3/native_frontier/optimal\endcsname{98.44}
\expandafter\def\csname joint@json/3/native_frontier/initial_success_rate\endcsname{60.09}
\expandafter\def\csname joint@json/3/native_frontier/initial_calls\endcsname{1.957}
\expandafter\def\csname joint@json/3/partial_native_full_rows/success_rate\endcsname{90.48}
\expandafter\def\csname joint@json/3/partial_native_full_rows/calls\endcsname{1.430}
\expandafter\def\csname joint@json/3/partial_native_full_rows/queries\endcsname{2354.3}
\expandafter\def\csname joint@json/3/partial_native_full_rows/lifecycle_seconds\endcsname{18.003}
\expandafter\def\csname joint@json/3/partial_native_full_rows/certified\endcsname{85.16}
\expandafter\def\csname joint@json/3/partial_native_full_rows/optimal\endcsname{86.72}
\expandafter\def\csname joint@json/3/partial_native_full_rows/initial_success_rate\endcsname{60.09}
\expandafter\def\csname joint@json/3/partial_native_full_rows/initial_calls\endcsname{1.957}
\expandafter\def\csname joint@json/3/partial_native_no_tightening/success_rate\endcsname{94.74}
\expandafter\def\csname joint@json/3/partial_native_no_tightening/calls\endcsname{1.476}
\expandafter\def\csname joint@json/3/partial_native_no_tightening/queries\endcsname{1516.4}
\expandafter\def\csname joint@json/3/partial_native_no_tightening/lifecycle_seconds\endcsname{11.921}
\expandafter\def\csname joint@json/3/partial_native_no_tightening/certified\endcsname{95.31}
\expandafter\def\csname joint@json/3/partial_native_no_tightening/optimal\endcsname{98.44}
\expandafter\def\csname joint@json/3/partial_native_no_tightening/initial_success_rate\endcsname{60.09}
\expandafter\def\csname joint@json/3/partial_native_no_tightening/initial_calls\endcsname{1.957}
\expandafter\def\csname joint@json/5/partial_native/success_rate\endcsname{94.74}
\expandafter\def\csname joint@json/5/partial_native/calls\endcsname{1.475}
\expandafter\def\csname joint@json/5/partial_native/queries\endcsname{1485.9}
\expandafter\def\csname joint@json/5/partial_native/lifecycle_seconds\endcsname{11.903}
\expandafter\def\csname joint@json/5/partial_native/certified\endcsname{96.88}
\expandafter\def\csname joint@json/5/partial_native/optimal\endcsname{100.00}
\expandafter\def\csname joint@json/5/partial_native/initial_success_rate\endcsname{67.90}
\expandafter\def\csname joint@json/5/partial_native/initial_calls\endcsname{2.731}
\expandafter\def\csname joint@json/5/eager_optimized/success_rate\endcsname{94.60}
\expandafter\def\csname joint@json/5/eager_optimized/calls\endcsname{1.474}
\expandafter\def\csname joint@json/5/eager_optimized/queries\endcsname{2317.8}
\expandafter\def\csname joint@json/5/eager_optimized/lifecycle_seconds\endcsname{18.079}
\expandafter\def\csname joint@json/5/eager_optimized/certified\endcsname{84.38}
\expandafter\def\csname joint@json/5/eager_optimized/optimal\endcsname{98.44}
\expandafter\def\csname joint@json/5/eager_optimized/initial_success_rate\endcsname{67.90}
\expandafter\def\csname joint@json/5/eager_optimized/initial_calls\endcsname{2.731}
\expandafter\def\csname joint@json/5/native_program_only/success_rate\endcsname{94.60}
\expandafter\def\csname joint@json/5/native_program_only/calls\endcsname{1.474}
\expandafter\def\csname joint@json/5/native_program_only/queries\endcsname{2077.9}
\expandafter\def\csname joint@json/5/native_program_only/lifecycle_seconds\endcsname{17.744}
\expandafter\def\csname joint@json/5/native_program_only/certified\endcsname{76.56}
\expandafter\def\csname joint@json/5/native_program_only/optimal\endcsname{98.44}
\expandafter\def\csname joint@json/5/native_program_only/initial_success_rate\endcsname{67.90}
\expandafter\def\csname joint@json/5/native_program_only/initial_calls\endcsname{2.731}
\expandafter\def\csname joint@json/5/initial_safe_two_step/success_rate\endcsname{94.60}
\expandafter\def\csname joint@json/5/initial_safe_two_step/calls\endcsname{1.493}
\expandafter\def\csname joint@json/5/initial_safe_two_step/queries\endcsname{2314.1}
\expandafter\def\csname joint@json/5/initial_safe_two_step/lifecycle_seconds\endcsname{18.005}
\expandafter\def\csname joint@json/5/initial_safe_two_step/certified\endcsname{77.34}
\expandafter\def\csname joint@json/5/initial_safe_two_step/optimal\endcsname{91.96}
\expandafter\def\csname joint@json/5/initial_safe_two_step/initial_success_rate\endcsname{67.90}
\expandafter\def\csname joint@json/5/initial_safe_two_step/initial_calls\endcsname{2.731}
\expandafter\def\csname joint@json/5/lazy_quotient/success_rate\endcsname{92.90}
\expandafter\def\csname joint@json/5/lazy_quotient/calls\endcsname{1.546}
\expandafter\def\csname joint@json/5/lazy_quotient/queries\endcsname{1954.9}
\expandafter\def\csname joint@json/5/lazy_quotient/lifecycle_seconds\endcsname{16.008}
\expandafter\def\csname joint@json/5/lazy_quotient/certified\endcsname{93.75}
\expandafter\def\csname joint@json/5/lazy_quotient/optimal\endcsname{95.31}
\expandafter\def\csname joint@json/5/lazy_quotient/initial_success_rate\endcsname{67.90}
\expandafter\def\csname joint@json/5/lazy_quotient/initial_calls\endcsname{2.731}
\expandafter\def\csname joint@json/5/native_frontier/success_rate\endcsname{94.60}
\expandafter\def\csname joint@json/5/native_frontier/calls\endcsname{1.474}
\expandafter\def\csname joint@json/5/native_frontier/queries\endcsname{2174.5}
\expandafter\def\csname joint@json/5/native_frontier/lifecycle_seconds\endcsname{17.919}
\expandafter\def\csname joint@json/5/native_frontier/certified\endcsname{83.59}
\expandafter\def\csname joint@json/5/native_frontier/optimal\endcsname{98.44}
\expandafter\def\csname joint@json/5/native_frontier/initial_success_rate\endcsname{67.90}
\expandafter\def\csname joint@json/5/native_frontier/initial_calls\endcsname{2.731}
\expandafter\def\csname joint@json/5/partial_native_full_rows/success_rate\endcsname{91.05}
\expandafter\def\csname joint@json/5/partial_native_full_rows/calls\endcsname{1.435}
\expandafter\def\csname joint@json/5/partial_native_full_rows/queries\endcsname{2049.3}
\expandafter\def\csname joint@json/5/partial_native_full_rows/lifecycle_seconds\endcsname{17.036}
\expandafter\def\csname joint@json/5/partial_native_full_rows/certified\endcsname{86.72}
\expandafter\def\csname joint@json/5/partial_native_full_rows/optimal\endcsname{89.84}
\expandafter\def\csname joint@json/5/partial_native_full_rows/initial_success_rate\endcsname{67.90}
\expandafter\def\csname joint@json/5/partial_native_full_rows/initial_calls\endcsname{2.731}
\expandafter\def\csname joint@json/5/partial_native_no_tightening/success_rate\endcsname{94.74}
\expandafter\def\csname joint@json/5/partial_native_no_tightening/calls\endcsname{1.475}
\expandafter\def\csname joint@json/5/partial_native_no_tightening/queries\endcsname{1486.2}
\expandafter\def\csname joint@json/5/partial_native_no_tightening/lifecycle_seconds\endcsname{11.862}
\expandafter\def\csname joint@json/5/partial_native_no_tightening/certified\endcsname{96.88}
\expandafter\def\csname joint@json/5/partial_native_no_tightening/optimal\endcsname{100.00}
\expandafter\def\csname joint@json/5/partial_native_no_tightening/initial_success_rate\endcsname{67.90}
\expandafter\def\csname joint@json/5/partial_native_no_tightening/initial_calls\endcsname{2.731}
\expandafter\def\csname joint@csv/3/eager_optimized/success_gain\endcsname{0.00}
\expandafter\def\csname joint@csv/3/eager_optimized/calls_saved\endcsname{0.000}
\expandafter\def\csname joint@csv/3/eager_optimized/seconds_saved\endcsname{0.399}
\expandafter\def\csname joint@csv/3/eager_optimized/queries_saved\endcsname{470.4}
\expandafter\def\csname joint@csv/3/eager_optimized/time_reduction\endcsname{3.6}
\expandafter\def\csname joint@csv/3/eager_optimized/query_reduction\endcsname{13.6}
\expandafter\def\csname joint@csv/3/eager_optimized/better\endcsname{0}
\expandafter\def\csname joint@csv/3/eager_optimized/tie\endcsname{126}
\expandafter\def\csname joint@csv/3/eager_optimized/worse\endcsname{0}
\expandafter\def\csname joint@csv/3/eager_optimized/invalid\endcsname{2}
\expandafter\def\csname joint@csv/3/eager_optimized/seconds_interval\endcsname{0.041, 0.768}
\expandafter\def\csname joint@csv/3/native_program_only/success_gain\endcsname{8.40}
\expandafter\def\csname joint@csv/3/native_program_only/calls_saved\endcsname{-0.174}
\expandafter\def\csname joint@csv/3/native_program_only/seconds_saved\endcsname{-7.143}
\expandafter\def\csname joint@csv/3/native_program_only/queries_saved\endcsname{-2356.5}
\expandafter\def\csname joint@csv/3/native_program_only/time_reduction\endcsname{-196.7}
\expandafter\def\csname joint@csv/3/native_program_only/query_reduction\endcsname{-377.6}
\expandafter\def\csname joint@csv/3/native_program_only/better\endcsname{82}
\expandafter\def\csname joint@csv/3/native_program_only/tie\endcsname{44}
\expandafter\def\csname joint@csv/3/native_program_only/worse\endcsname{0}
\expandafter\def\csname joint@csv/3/native_program_only/invalid\endcsname{2}
\expandafter\def\csname joint@csv/3/native_program_only/seconds_interval\endcsname{-9.004, -4.786}
\expandafter\def\csname joint@csv/3/initial_safe_two_step/success_gain\endcsname{1.17}
\expandafter\def\csname joint@csv/3/initial_safe_two_step/calls_saved\endcsname{0.278}
\expandafter\def\csname joint@csv/3/initial_safe_two_step/seconds_saved\endcsname{0.348}
\expandafter\def\csname joint@csv/3/initial_safe_two_step/queries_saved\endcsname{470.4}
\expandafter\def\csname joint@csv/3/initial_safe_two_step/time_reduction\endcsname{3.1}
\expandafter\def\csname joint@csv/3/initial_safe_two_step/query_reduction\endcsname{13.6}
\expandafter\def\csname joint@csv/3/initial_safe_two_step/better\endcsname{82}
\expandafter\def\csname joint@csv/3/initial_safe_two_step/tie\endcsname{44}
\expandafter\def\csname joint@csv/3/initial_safe_two_step/worse\endcsname{0}
\expandafter\def\csname joint@csv/3/initial_safe_two_step/invalid\endcsname{2}
\expandafter\def\csname joint@csv/3/initial_safe_two_step/seconds_interval\endcsname{-0.054, 0.713}
\expandafter\def\csname joint@csv/3/lazy_quotient/success_gain\endcsname{1.42}
\expandafter\def\csname joint@csv/3/lazy_quotient/calls_saved\endcsname{0.001}
\expandafter\def\csname joint@csv/3/lazy_quotient/seconds_saved\endcsname{6.316}
\expandafter\def\csname joint@csv/3/lazy_quotient/queries_saved\endcsname{352.0}
\expandafter\def\csname joint@csv/3/lazy_quotient/time_reduction\endcsname{37.0}
\expandafter\def\csname joint@csv/3/lazy_quotient/query_reduction\endcsname{10.6}
\expandafter\def\csname joint@csv/3/lazy_quotient/better\endcsname{6}
\expandafter\def\csname joint@csv/3/lazy_quotient/tie\endcsname{120}
\expandafter\def\csname joint@csv/3/lazy_quotient/worse\endcsname{0}
\expandafter\def\csname joint@csv/3/lazy_quotient/invalid\endcsname{2}
\expandafter\def\csname joint@csv/3/lazy_quotient/seconds_interval\endcsname{3.972, 8.385}
\expandafter\def\csname joint@csv/3/native_frontier/success_gain\endcsname{0.00}
\expandafter\def\csname joint@csv/3/native_frontier/calls_saved\endcsname{0.000}
\expandafter\def\csname joint@csv/3/native_frontier/seconds_saved\endcsname{0.628}
\expandafter\def\csname joint@csv/3/native_frontier/queries_saved\endcsname{326.8}
\expandafter\def\csname joint@csv/3/native_frontier/time_reduction\endcsname{5.5}
\expandafter\def\csname joint@csv/3/native_frontier/query_reduction\endcsname{9.9}
\expandafter\def\csname joint@csv/3/native_frontier/better\endcsname{0}
\expandafter\def\csname joint@csv/3/native_frontier/tie\endcsname{126}
\expandafter\def\csname joint@csv/3/native_frontier/worse\endcsname{0}
\expandafter\def\csname joint@csv/3/native_frontier/invalid\endcsname{2}
\expandafter\def\csname joint@csv/3/native_frontier/seconds_interval\endcsname{-0.076, 1.255}
\expandafter\def\csname joint@csv/3/partial_native_full_rows/success_gain\endcsname{0.00}
\expandafter\def\csname joint@csv/3/partial_native_full_rows/calls_saved\endcsname{0.000}
\expandafter\def\csname joint@csv/3/partial_native_full_rows/seconds_saved\endcsname{0.538}
\expandafter\def\csname joint@csv/3/partial_native_full_rows/queries_saved\endcsname{263.1}
\expandafter\def\csname joint@csv/3/partial_native_full_rows/time_reduction\endcsname{4.8}
\expandafter\def\csname joint@csv/3/partial_native_full_rows/query_reduction\endcsname{8.1}
\expandafter\def\csname joint@csv/3/partial_native_full_rows/better\endcsname{0}
\expandafter\def\csname joint@csv/3/partial_native_full_rows/tie\endcsname{126}
\expandafter\def\csname joint@csv/3/partial_native_full_rows/worse\endcsname{0}
\expandafter\def\csname joint@csv/3/partial_native_full_rows/invalid\endcsname{2}
\expandafter\def\csname joint@csv/3/partial_native_full_rows/seconds_interval\endcsname{-0.071, 1.050}
\expandafter\def\csname joint@csv/3/partial_native_no_tightening/success_gain\endcsname{0.00}
\expandafter\def\csname joint@csv/3/partial_native_no_tightening/calls_saved\endcsname{0.000}
\expandafter\def\csname joint@csv/3/partial_native_no_tightening/seconds_saved\endcsname{0.601}
\expandafter\def\csname joint@csv/3/partial_native_no_tightening/queries_saved\endcsname{349.8}
\expandafter\def\csname joint@csv/3/partial_native_no_tightening/time_reduction\endcsname{5.3}
\expandafter\def\csname joint@csv/3/partial_native_no_tightening/query_reduction\endcsname{10.5}
\expandafter\def\csname joint@csv/3/partial_native_no_tightening/better\endcsname{0}
\expandafter\def\csname joint@csv/3/partial_native_no_tightening/tie\endcsname{126}
\expandafter\def\csname joint@csv/3/partial_native_no_tightening/worse\endcsname{0}
\expandafter\def\csname joint@csv/3/partial_native_no_tightening/invalid\endcsname{2}
\expandafter\def\csname joint@csv/3/partial_native_no_tightening/seconds_interval\endcsname{0.113, 1.076}
\expandafter\def\csname joint@json/3/eager_optimized/success_gain\endcsname{0.14}
\expandafter\def\csname joint@json/3/eager_optimized/calls_saved\endcsname{-0.002}
\expandafter\def\csname joint@json/3/eager_optimized/seconds_saved\endcsname{6.102}
\expandafter\def\csname joint@json/3/eager_optimized/queries_saved\endcsname{802.8}
\expandafter\def\csname joint@json/3/eager_optimized/time_reduction\endcsname{33.9}
\expandafter\def\csname joint@json/3/eager_optimized/query_reduction\endcsname{34.8}
\expandafter\def\csname joint@json/3/eager_optimized/better\endcsname{2}
\expandafter\def\csname joint@json/3/eager_optimized/tie\endcsname{122}
\expandafter\def\csname joint@json/3/eager_optimized/worse\endcsname{2}
\expandafter\def\csname joint@json/3/eager_optimized/invalid\endcsname{2}
\expandafter\def\csname joint@json/3/eager_optimized/seconds_interval\endcsname{4.074, 8.107}
\expandafter\def\csname joint@json/3/native_program_only/success_gain\endcsname{0.14}
\expandafter\def\csname joint@json/3/native_program_only/calls_saved\endcsname{-0.002}
\expandafter\def\csname joint@json/3/native_program_only/seconds_saved\endcsname{5.873}
\expandafter\def\csname joint@json/3/native_program_only/queries_saved\endcsname{569.5}
\expandafter\def\csname joint@json/3/native_program_only/time_reduction\endcsname{33.0}
\expandafter\def\csname joint@json/3/native_program_only/query_reduction\endcsname{27.4}
\expandafter\def\csname joint@json/3/native_program_only/better\endcsname{2}
\expandafter\def\csname joint@json/3/native_program_only/tie\endcsname{122}
\expandafter\def\csname joint@json/3/native_program_only/worse\endcsname{2}
\expandafter\def\csname joint@json/3/native_program_only/invalid\endcsname{2}
\expandafter\def\csname joint@json/3/native_program_only/seconds_interval\endcsname{3.659, 8.080}
\expandafter\def\csname joint@json/3/initial_safe_two_step/success_gain\endcsname{0.14}
\expandafter\def\csname joint@json/3/initial_safe_two_step/calls_saved\endcsname{0.017}
\expandafter\def\csname joint@json/3/initial_safe_two_step/seconds_saved\endcsname{6.088}
\expandafter\def\csname joint@json/3/initial_safe_two_step/queries_saved\endcsname{801.0}
\expandafter\def\csname joint@json/3/initial_safe_two_step/time_reduction\endcsname{33.8}
\expandafter\def\csname joint@json/3/initial_safe_two_step/query_reduction\endcsname{34.7}
\expandafter\def\csname joint@json/3/initial_safe_two_step/better\endcsname{10}
\expandafter\def\csname joint@json/3/initial_safe_two_step/tie\endcsname{114}
\expandafter\def\csname joint@json/3/initial_safe_two_step/worse\endcsname{2}
\expandafter\def\csname joint@json/3/initial_safe_two_step/invalid\endcsname{2}
\expandafter\def\csname joint@json/3/initial_safe_two_step/seconds_interval\endcsname{4.061, 8.108}
\expandafter\def\csname joint@json/3/lazy_quotient/success_gain\endcsname{1.85}
\expandafter\def\csname joint@json/3/lazy_quotient/calls_saved\endcsname{0.019}
\expandafter\def\csname joint@json/3/lazy_quotient/seconds_saved\endcsname{4.281}
\expandafter\def\csname joint@json/3/lazy_quotient/queries_saved\endcsname{523.3}
\expandafter\def\csname joint@json/3/lazy_quotient/time_reduction\endcsname{26.4}
\expandafter\def\csname joint@json/3/lazy_quotient/query_reduction\endcsname{25.8}
\expandafter\def\csname joint@json/3/lazy_quotient/better\endcsname{6}
\expandafter\def\csname joint@json/3/lazy_quotient/tie\endcsname{120}
\expandafter\def\csname joint@json/3/lazy_quotient/worse\endcsname{0}
\expandafter\def\csname joint@json/3/lazy_quotient/invalid\endcsname{2}
\expandafter\def\csname joint@json/3/lazy_quotient/seconds_interval\endcsname{2.827, 5.757}
\expandafter\def\csname joint@json/3/native_frontier/success_gain\endcsname{0.14}
\expandafter\def\csname joint@json/3/native_frontier/calls_saved\endcsname{-0.002}
\expandafter\def\csname joint@json/3/native_frontier/seconds_saved\endcsname{5.959}
\expandafter\def\csname joint@json/3/native_frontier/queries_saved\endcsname{663.8}
\expandafter\def\csname joint@json/3/native_frontier/time_reduction\endcsname{33.3}
\expandafter\def\csname joint@json/3/native_frontier/query_reduction\endcsname{30.6}
\expandafter\def\csname joint@json/3/native_frontier/better\endcsname{2}
\expandafter\def\csname joint@json/3/native_frontier/tie\endcsname{122}
\expandafter\def\csname joint@json/3/native_frontier/worse\endcsname{2}
\expandafter\def\csname joint@json/3/native_frontier/invalid\endcsname{2}
\expandafter\def\csname joint@json/3/native_frontier/seconds_interval\endcsname{3.876, 8.132}
\expandafter\def\csname joint@json/3/partial_native_full_rows/success_gain\endcsname{4.26}
\expandafter\def\csname joint@json/3/partial_native_full_rows/calls_saved\endcsname{-0.046}
\expandafter\def\csname joint@json/3/partial_native_full_rows/seconds_saved\endcsname{6.084}
\expandafter\def\csname joint@json/3/partial_native_full_rows/queries_saved\endcsname{848.1}
\expandafter\def\csname joint@json/3/partial_native_full_rows/time_reduction\endcsname{33.8}
\expandafter\def\csname joint@json/3/partial_native_full_rows/query_reduction\endcsname{36.0}
\expandafter\def\csname joint@json/3/partial_native_full_rows/better\endcsname{17}
\expandafter\def\csname joint@json/3/partial_native_full_rows/tie\endcsname{109}
\expandafter\def\csname joint@json/3/partial_native_full_rows/worse\endcsname{0}
\expandafter\def\csname joint@json/3/partial_native_full_rows/invalid\endcsname{2}
\expandafter\def\csname joint@json/3/partial_native_full_rows/seconds_interval\endcsname{4.122, 8.051}
\expandafter\def\csname joint@json/3/partial_native_no_tightening/success_gain\endcsname{0.00}
\expandafter\def\csname joint@json/3/partial_native_no_tightening/calls_saved\endcsname{0.000}
\expandafter\def\csname joint@json/3/partial_native_no_tightening/seconds_saved\endcsname{0.002}
\expandafter\def\csname joint@json/3/partial_native_no_tightening/queries_saved\endcsname{10.2}
\expandafter\def\csname joint@json/3/partial_native_no_tightening/time_reduction\endcsname{0.0}
\expandafter\def\csname joint@json/3/partial_native_no_tightening/query_reduction\endcsname{0.7}
\expandafter\def\csname joint@json/3/partial_native_no_tightening/better\endcsname{0}
\expandafter\def\csname joint@json/3/partial_native_no_tightening/tie\endcsname{126}
\expandafter\def\csname joint@json/3/partial_native_no_tightening/worse\endcsname{0}
\expandafter\def\csname joint@json/3/partial_native_no_tightening/invalid\endcsname{2}
\expandafter\def\csname joint@json/3/partial_native_no_tightening/seconds_interval\endcsname{-0.022, 0.026}
\expandafter\def\csname joint@csv/5/eager_optimized/success_gain\endcsname{0.00}
\expandafter\def\csname joint@csv/5/eager_optimized/calls_saved\endcsname{0.000}
\expandafter\def\csname joint@csv/5/eager_optimized/seconds_saved\endcsname{0.216}
\expandafter\def\csname joint@csv/5/eager_optimized/queries_saved\endcsname{289.7}
\expandafter\def\csname joint@csv/5/eager_optimized/time_reduction\endcsname{1.9}
\expandafter\def\csname joint@csv/5/eager_optimized/query_reduction\endcsname{8.4}
\expandafter\def\csname joint@csv/5/eager_optimized/better\endcsname{0}
\expandafter\def\csname joint@csv/5/eager_optimized/tie\endcsname{126}
\expandafter\def\csname joint@csv/5/eager_optimized/worse\endcsname{0}
\expandafter\def\csname joint@csv/5/eager_optimized/invalid\endcsname{2}
\expandafter\def\csname joint@csv/5/eager_optimized/seconds_interval\endcsname{-0.125, 0.693}
\expandafter\def\csname joint@csv/5/native_program_only/success_gain\endcsname{6.25}
\expandafter\def\csname joint@csv/5/native_program_only/calls_saved\endcsname{0.017}
\expandafter\def\csname joint@csv/5/native_program_only/seconds_saved\endcsname{-7.577}
\expandafter\def\csname joint@csv/5/native_program_only/queries_saved\endcsname{-2537.1}
\expandafter\def\csname joint@csv/5/native_program_only/time_reduction\endcsname{-209.1}
\expandafter\def\csname joint@csv/5/native_program_only/query_reduction\endcsname{-406.6}
\expandafter\def\csname joint@csv/5/native_program_only/better\endcsname{32}
\expandafter\def\csname joint@csv/5/native_program_only/tie\endcsname{94}
\expandafter\def\csname joint@csv/5/native_program_only/worse\endcsname{0}
\expandafter\def\csname joint@csv/5/native_program_only/invalid\endcsname{2}
\expandafter\def\csname joint@csv/5/native_program_only/seconds_interval\endcsname{-9.510, -5.041}
\expandafter\def\csname joint@csv/5/initial_safe_two_step/success_gain\endcsname{0.00}
\expandafter\def\csname joint@csv/5/initial_safe_two_step/calls_saved\endcsname{0.490}
\expandafter\def\csname joint@csv/5/initial_safe_two_step/seconds_saved\endcsname{-0.010}
\expandafter\def\csname joint@csv/5/initial_safe_two_step/queries_saved\endcsname{289.7}
\expandafter\def\csname joint@csv/5/initial_safe_two_step/time_reduction\endcsname{-0.1}
\expandafter\def\csname joint@csv/5/initial_safe_two_step/query_reduction\endcsname{8.4}
\expandafter\def\csname joint@csv/5/initial_safe_two_step/better\endcsname{38}
\expandafter\def\csname joint@csv/5/initial_safe_two_step/tie\endcsname{88}
\expandafter\def\csname joint@csv/5/initial_safe_two_step/worse\endcsname{0}
\expandafter\def\csname joint@csv/5/initial_safe_two_step/invalid\endcsname{2}
\expandafter\def\csname joint@csv/5/initial_safe_two_step/seconds_interval\endcsname{-0.558, 0.562}
\expandafter\def\csname joint@csv/5/lazy_quotient/success_gain\endcsname{14.26}
\expandafter\def\csname joint@csv/5/lazy_quotient/calls_saved\endcsname{0.187}
\expandafter\def\csname joint@csv/5/lazy_quotient/seconds_saved\endcsname{2.498}
\expandafter\def\csname joint@csv/5/lazy_quotient/queries_saved\endcsname{-779.1}
\expandafter\def\csname joint@csv/5/lazy_quotient/time_reduction\endcsname{18.2}
\expandafter\def\csname joint@csv/5/lazy_quotient/query_reduction\endcsname{-32.7}
\expandafter\def\csname joint@csv/5/lazy_quotient/better\endcsname{32}
\expandafter\def\csname joint@csv/5/lazy_quotient/tie\endcsname{94}
\expandafter\def\csname joint@csv/5/lazy_quotient/worse\endcsname{0}
\expandafter\def\csname joint@csv/5/lazy_quotient/invalid\endcsname{2}
\expandafter\def\csname joint@csv/5/lazy_quotient/seconds_interval\endcsname{0.342, 4.938}
\expandafter\def\csname joint@csv/5/native_frontier/success_gain\endcsname{0.00}
\expandafter\def\csname joint@csv/5/native_frontier/calls_saved\endcsname{0.000}
\expandafter\def\csname joint@csv/5/native_frontier/seconds_saved\endcsname{0.371}
\expandafter\def\csname joint@csv/5/native_frontier/queries_saved\endcsname{30.5}
\expandafter\def\csname joint@csv/5/native_frontier/time_reduction\endcsname{3.2}
\expandafter\def\csname joint@csv/5/native_frontier/query_reduction\endcsname{1.0}
\expandafter\def\csname joint@csv/5/native_frontier/better\endcsname{0}
\expandafter\def\csname joint@csv/5/native_frontier/tie\endcsname{126}
\expandafter\def\csname joint@csv/5/native_frontier/worse\endcsname{0}
\expandafter\def\csname joint@csv/5/native_frontier/invalid\endcsname{2}
\expandafter\def\csname joint@csv/5/native_frontier/seconds_interval\endcsname{0.146, 0.729}
\expandafter\def\csname joint@csv/5/partial_native_full_rows/success_gain\endcsname{0.00}
\expandafter\def\csname joint@csv/5/partial_native_full_rows/calls_saved\endcsname{0.000}
\expandafter\def\csname joint@csv/5/partial_native_full_rows/seconds_saved\endcsname{-0.024}
\expandafter\def\csname joint@csv/5/partial_native_full_rows/queries_saved\endcsname{-0.2}
\expandafter\def\csname joint@csv/5/partial_native_full_rows/time_reduction\endcsname{-0.2}
\expandafter\def\csname joint@csv/5/partial_native_full_rows/query_reduction\endcsname{-0.0}
\expandafter\def\csname joint@csv/5/partial_native_full_rows/better\endcsname{0}
\expandafter\def\csname joint@csv/5/partial_native_full_rows/tie\endcsname{126}
\expandafter\def\csname joint@csv/5/partial_native_full_rows/worse\endcsname{0}
\expandafter\def\csname joint@csv/5/partial_native_full_rows/invalid\endcsname{2}
\expandafter\def\csname joint@csv/5/partial_native_full_rows/seconds_interval\endcsname{-0.124, 0.080}
\expandafter\def\csname joint@csv/5/partial_native_no_tightening/success_gain\endcsname{0.00}
\expandafter\def\csname joint@csv/5/partial_native_no_tightening/calls_saved\endcsname{0.000}
\expandafter\def\csname joint@csv/5/partial_native_no_tightening/seconds_saved\endcsname{0.015}
\expandafter\def\csname joint@csv/5/partial_native_no_tightening/queries_saved\endcsname{0.0}
\expandafter\def\csname joint@csv/5/partial_native_no_tightening/time_reduction\endcsname{0.1}
\expandafter\def\csname joint@csv/5/partial_native_no_tightening/query_reduction\endcsname{0.0}
\expandafter\def\csname joint@csv/5/partial_native_no_tightening/better\endcsname{0}
\expandafter\def\csname joint@csv/5/partial_native_no_tightening/tie\endcsname{126}
\expandafter\def\csname joint@csv/5/partial_native_no_tightening/worse\endcsname{0}
\expandafter\def\csname joint@csv/5/partial_native_no_tightening/invalid\endcsname{2}
\expandafter\def\csname joint@csv/5/partial_native_no_tightening/seconds_interval\endcsname{-0.033, 0.065}
\expandafter\def\csname joint@json/5/eager_optimized/success_gain\endcsname{0.14}
\expandafter\def\csname joint@json/5/eager_optimized/calls_saved\endcsname{-0.001}
\expandafter\def\csname joint@json/5/eager_optimized/seconds_saved\endcsname{6.176}
\expandafter\def\csname joint@json/5/eager_optimized/queries_saved\endcsname{831.9}
\expandafter\def\csname joint@json/5/eager_optimized/time_reduction\endcsname{34.2}
\expandafter\def\csname joint@json/5/eager_optimized/query_reduction\endcsname{35.9}
\expandafter\def\csname joint@json/5/eager_optimized/better\endcsname{2}
\expandafter\def\csname joint@json/5/eager_optimized/tie\endcsname{124}
\expandafter\def\csname joint@json/5/eager_optimized/worse\endcsname{0}
\expandafter\def\csname joint@json/5/eager_optimized/invalid\endcsname{2}
\expandafter\def\csname joint@json/5/eager_optimized/seconds_interval\endcsname{4.131, 8.157}
\expandafter\def\csname joint@json/5/native_program_only/success_gain\endcsname{0.14}
\expandafter\def\csname joint@json/5/native_program_only/calls_saved\endcsname{-0.001}
\expandafter\def\csname joint@json/5/native_program_only/seconds_saved\endcsname{5.842}
\expandafter\def\csname joint@json/5/native_program_only/queries_saved\endcsname{591.9}
\expandafter\def\csname joint@json/5/native_program_only/time_reduction\endcsname{32.9}
\expandafter\def\csname joint@json/5/native_program_only/query_reduction\endcsname{28.5}
\expandafter\def\csname joint@json/5/native_program_only/better\endcsname{2}
\expandafter\def\csname joint@json/5/native_program_only/tie\endcsname{124}
\expandafter\def\csname joint@json/5/native_program_only/worse\endcsname{0}
\expandafter\def\csname joint@json/5/native_program_only/invalid\endcsname{2}
\expandafter\def\csname joint@json/5/native_program_only/seconds_interval\endcsname{3.735, 8.026}
\expandafter\def\csname joint@json/5/initial_safe_two_step/success_gain\endcsname{0.14}
\expandafter\def\csname joint@json/5/initial_safe_two_step/calls_saved\endcsname{0.017}
\expandafter\def\csname joint@json/5/initial_safe_two_step/seconds_saved\endcsname{6.103}
\expandafter\def\csname joint@json/5/initial_safe_two_step/queries_saved\endcsname{828.2}
\expandafter\def\csname joint@json/5/initial_safe_two_step/time_reduction\endcsname{33.9}
\expandafter\def\csname joint@json/5/initial_safe_two_step/query_reduction\endcsname{35.8}
\expandafter\def\csname joint@json/5/initial_safe_two_step/better\endcsname{10}
\expandafter\def\csname joint@json/5/initial_safe_two_step/tie\endcsname{116}
\expandafter\def\csname joint@json/5/initial_safe_two_step/worse\endcsname{0}
\expandafter\def\csname joint@json/5/initial_safe_two_step/invalid\endcsname{2}
\expandafter\def\csname joint@json/5/initial_safe_two_step/seconds_interval\endcsname{4.136, 8.141}
\expandafter\def\csname joint@json/5/lazy_quotient/success_gain\endcsname{1.85}
\expandafter\def\csname joint@json/5/lazy_quotient/calls_saved\endcsname{0.071}
\expandafter\def\csname joint@json/5/lazy_quotient/seconds_saved\endcsname{4.105}
\expandafter\def\csname joint@json/5/lazy_quotient/queries_saved\endcsname{468.9}
\expandafter\def\csname joint@json/5/lazy_quotient/time_reduction\endcsname{25.6}
\expandafter\def\csname joint@json/5/lazy_quotient/query_reduction\endcsname{24.0}
\expandafter\def\csname joint@json/5/lazy_quotient/better\endcsname{6}
\expandafter\def\csname joint@json/5/lazy_quotient/tie\endcsname{120}
\expandafter\def\csname joint@json/5/lazy_quotient/worse\endcsname{0}
\expandafter\def\csname joint@json/5/lazy_quotient/invalid\endcsname{2}
\expandafter\def\csname joint@json/5/lazy_quotient/seconds_interval\endcsname{2.774, 5.458}
\expandafter\def\csname joint@json/5/native_frontier/success_gain\endcsname{0.14}
\expandafter\def\csname joint@json/5/native_frontier/calls_saved\endcsname{-0.001}
\expandafter\def\csname joint@json/5/native_frontier/seconds_saved\endcsname{6.016}
\expandafter\def\csname joint@json/5/native_frontier/queries_saved\endcsname{688.6}
\expandafter\def\csname joint@json/5/native_frontier/time_reduction\endcsname{33.6}
\expandafter\def\csname joint@json/5/native_frontier/query_reduction\endcsname{31.7}
\expandafter\def\csname joint@json/5/native_frontier/better\endcsname{2}
\expandafter\def\csname joint@json/5/native_frontier/tie\endcsname{124}
\expandafter\def\csname joint@json/5/native_frontier/worse\endcsname{0}
\expandafter\def\csname joint@json/5/native_frontier/invalid\endcsname{2}
\expandafter\def\csname joint@json/5/native_frontier/seconds_interval\endcsname{3.899, 8.168}
\expandafter\def\csname joint@json/5/partial_native_full_rows/success_gain\endcsname{3.69}
\expandafter\def\csname joint@json/5/partial_native_full_rows/calls_saved\endcsname{-0.040}
\expandafter\def\csname joint@json/5/partial_native_full_rows/seconds_saved\endcsname{5.134}
\expandafter\def\csname joint@json/5/partial_native_full_rows/queries_saved\endcsname{563.3}
\expandafter\def\csname joint@json/5/partial_native_full_rows/time_reduction\endcsname{30.1}
\expandafter\def\csname joint@json/5/partial_native_full_rows/query_reduction\endcsname{27.5}
\expandafter\def\csname joint@json/5/partial_native_full_rows/better\endcsname{13}
\expandafter\def\csname joint@json/5/partial_native_full_rows/tie\endcsname{113}
\expandafter\def\csname joint@json/5/partial_native_full_rows/worse\endcsname{0}
\expandafter\def\csname joint@json/5/partial_native_full_rows/invalid\endcsname{2}
\expandafter\def\csname joint@json/5/partial_native_full_rows/seconds_interval\endcsname{2.801, 7.523}
\expandafter\def\csname joint@json/5/partial_native_no_tightening/success_gain\endcsname{0.00}
\expandafter\def\csname joint@json/5/partial_native_no_tightening/calls_saved\endcsname{0.000}
\expandafter\def\csname joint@json/5/partial_native_no_tightening/seconds_saved\endcsname{-0.041}
\expandafter\def\csname joint@json/5/partial_native_no_tightening/queries_saved\endcsname{0.3}
\expandafter\def\csname joint@json/5/partial_native_no_tightening/time_reduction\endcsname{-0.3}
\expandafter\def\csname joint@json/5/partial_native_no_tightening/query_reduction\endcsname{0.0}
\expandafter\def\csname joint@json/5/partial_native_no_tightening/better\endcsname{0}
\expandafter\def\csname joint@json/5/partial_native_no_tightening/tie\endcsname{126}
\expandafter\def\csname joint@json/5/partial_native_no_tightening/worse\endcsname{0}
\expandafter\def\csname joint@json/5/partial_native_no_tightening/invalid\endcsname{2}
\expandafter\def\csname joint@json/5/partial_native_no_tightening/seconds_interval\endcsname{-0.100, 0.016}
\expandafter\def\csname joint@pipeline_records\endcsname{4096}
\expandafter\def\csname joint@measurements\endcsname{4032}
\expandafter\def\csname joint@native_measurements\endcsname{4032}
\expandafter\def\csname joint@native_inputs\endcsname{126}
\expandafter\def\csname joint@structures\endcsname{24}
\expandafter\def\csname joint@policy_pairs\endcsname{586}
\expandafter\def\csname joint@native_episodes\endcsname{83,744}
\expandafter\def\csname joint@native_calls\endcsname{195,798}
\expandafter\def\csname joint@objectives\endcsname{252}
\expandafter\def\csname joint@resolved_objectives\endcsname{252}
\expandafter\def\csname joint@history_witnesses\endcsname{8970}
\expandafter\def\csname joint@acquisition/acquisitions\endcsname{41626}
\expandafter\def\csname joint@acquisition/local_acquisition_attempts\endcsname{23180}
\expandafter\def\csname joint@acquisition/local_acquisitions\endcsname{23180}
\expandafter\def\csname joint@acquisition/local_effect_queries\endcsname{308858}
\expandafter\def\csname joint@acquisition/runs_with_local_acquisition\endcsname{380}
\expandafter\def\csname joint@candidate/native_records\endcsname{504}
\expandafter\def\csname joint@candidate/optimal_records\endcsname{502}
\expandafter\def\csname joint@candidate/certified_records\endcsname{498}
\expandafter\def\csname joint@dev/csv/candidate_queries\endcsname{3860}
\expandafter\def\csname joint@dev/csv/control_queries\endcsname{4672}
\expandafter\def\csname joint@dev/json/candidate_queries\endcsname{632}
\expandafter\def\csname joint@dev/json/control_queries\endcsname{848}
\expandafter\def\csname joint@overall/partial_native/lifecycle_seconds\endcsname{11.449}
\expandafter\def\csname joint@overall/partial_native/queries\endcsname{2283.4}
\expandafter\def\csname joint@overall/partial_native/success_rate\endcsname{93.27}
\expandafter\def\csname joint@overall/partial_native/calls\endcsname{1.874}
\expandafter\def\csname joint@overall/eager_optimized/lifecycle_seconds\endcsname{14.673}
\expandafter\def\csname joint@overall/eager_optimized/queries\endcsname{2882.1}
\expandafter\def\csname joint@overall/eager_optimized/success_rate\endcsname{93.20}
\expandafter\def\csname joint@overall/eager_optimized/calls\endcsname{1.873}
\expandafter\def\csname joint@overall/native_program_only/lifecycle_seconds\endcsname{10.698}
\expandafter\def\csname joint@overall/native_program_only/queries\endcsname{1350.4}
\expandafter\def\csname joint@overall/native_program_only/success_rate\endcsname{89.54}
\expandafter\def\csname joint@overall/native_program_only/calls\endcsname{1.834}
\expandafter\def\csname joint@overall/initial_safe_two_step/lifecycle_seconds\endcsname{14.582}
\expandafter\def\csname joint@overall/initial_safe_two_step/queries\endcsname{2880.7}
\expandafter\def\csname joint@overall/initial_safe_two_step/success_rate\endcsname{92.91}
\expandafter\def\csname joint@overall/initial_safe_two_step/calls\endcsname{2.074}
\expandafter\def\csname joint@overall/lazy_quotient/lifecycle_seconds\endcsname{15.749}
\expandafter\def\csname joint@overall/lazy_quotient/queries\endcsname{2424.7}
\expandafter\def\csname joint@overall/lazy_quotient/success_rate\endcsname{88.43}
\expandafter\def\csname joint@overall/lazy_quotient/calls\endcsname{1.944}
\expandafter\def\csname joint@overall/native_frontier/lifecycle_seconds\endcsname{14.693}
\expandafter\def\csname joint@overall/native_frontier/queries\endcsname{2710.9}
\expandafter\def\csname joint@overall/native_frontier/success_rate\endcsname{93.20}
\expandafter\def\csname joint@overall/native_frontier/calls\endcsname{1.873}
\expandafter\def\csname joint@overall/partial_native_full_rows/lifecycle_seconds\endcsname{14.382}
\expandafter\def\csname joint@overall/partial_native_full_rows/queries\endcsname{2702.0}
\expandafter\def\csname joint@overall/partial_native_full_rows/success_rate\endcsname{91.28}
\expandafter\def\csname joint@overall/partial_native_full_rows/calls\endcsname{1.852}
\expandafter\def\csname joint@overall/partial_native_no_tightening/lifecycle_seconds\endcsname{11.594}
\expandafter\def\csname joint@overall/partial_native_no_tightening/queries\endcsname{2373.5}
\expandafter\def\csname joint@overall/partial_native_no_tightening/success_rate\endcsname{93.27}
\expandafter\def\csname joint@overall/partial_native_no_tightening/calls\endcsname{1.874}
\expandafter\def\csname joint@overall/eager_optimized/time_reduction\endcsname{22.0}
\expandafter\def\csname joint@overall/eager_optimized/query_reduction\endcsname{20.8}
\expandafter\def\csname joint@overall/native_program_only/time_reduction\endcsname{-7.0}
\expandafter\def\csname joint@overall/native_program_only/query_reduction\endcsname{-69.1}
\expandafter\def\csname joint@overall/initial_safe_two_step/time_reduction\endcsname{21.5}
\expandafter\def\csname joint@overall/initial_safe_two_step/query_reduction\endcsname{20.7}
\expandafter\def\csname joint@overall/lazy_quotient/time_reduction\endcsname{27.3}
\expandafter\def\csname joint@overall/lazy_quotient/query_reduction\endcsname{5.8}
\expandafter\def\csname joint@overall/native_frontier/time_reduction\endcsname{22.1}
\expandafter\def\csname joint@overall/native_frontier/query_reduction\endcsname{15.8}
\expandafter\def\csname joint@overall/partial_native_full_rows/time_reduction\endcsname{20.4}
\expandafter\def\csname joint@overall/partial_native_full_rows/query_reduction\endcsname{15.5}
\expandafter\def\csname joint@overall/partial_native_no_tightening/time_reduction\endcsname{1.2}
\expandafter\def\csname joint@overall/partial_native_no_tightening/query_reduction\endcsname{3.8}

\newcommand{\MFVal}[1]{\csname matched@#1\endcsname}
\expandafter\def\csname matched@primary_records\endcsname{3648}
\expandafter\def\csname matched@diagnostic_records\endcsname{1824}
\expandafter\def\csname matched@input_groups\endcsname{304}
\expandafter\def\csname matched@native_models\endcsname{152}
\expandafter\def\csname matched@structures\endcsname{33}
\expandafter\def\csname matched@optimal_objectives\endcsname{304}
\expandafter\def\csname matched@profile_value_agreement\endcsname{1805}
\expandafter\def\csname matched@csv/3/cape_none/success_rate\endcsname{85.156}
\expandafter\def\csname matched@csv/3/cape_none/calls\endcsname{2.196}
\expandafter\def\csname matched@csv/3/cape_none/queries\endcsname{3330.308}
\expandafter\def\csname matched@csv/3/cape_none/lifecycle_seconds\endcsname{11.489}
\expandafter\def\csname matched@csv/3/cape_selective/success_rate\endcsname{85.156}
\expandafter\def\csname matched@csv/3/cape_selective/calls\endcsname{2.196}
\expandafter\def\csname matched@csv/3/cape_selective/queries\endcsname{2979.933}
\expandafter\def\csname matched@csv/3/cape_selective/lifecycle_seconds\endcsname{10.918}
\expandafter\def\csname matched@csv/3/cape_always/success_rate\endcsname{85.156}
\expandafter\def\csname matched@csv/3/cape_always/calls\endcsname{2.196}
\expandafter\def\csname matched@csv/3/cape_always/queries\endcsname{2546.355}
\expandafter\def\csname matched@csv/3/cape_always/lifecycle_seconds\endcsname{10.177}
\expandafter\def\csname matched@csv/3/lazy_andor_none/success_rate\endcsname{85.156}
\expandafter\def\csname matched@csv/3/lazy_andor_none/calls\endcsname{2.196}
\expandafter\def\csname matched@csv/3/lazy_andor_none/queries\endcsname{3278.605}
\expandafter\def\csname matched@csv/3/lazy_andor_none/lifecycle_seconds\endcsname{11.315}
\expandafter\def\csname matched@csv/3/lazy_andor_selective/success_rate\endcsname{85.156}
\expandafter\def\csname matched@csv/3/lazy_andor_selective/calls\endcsname{2.196}
\expandafter\def\csname matched@csv/3/lazy_andor_selective/queries\endcsname{3250.518}
\expandafter\def\csname matched@csv/3/lazy_andor_selective/lifecycle_seconds\endcsname{11.916}
\expandafter\def\csname matched@csv/3/lazy_andor_always/success_rate\endcsname{85.156}
\expandafter\def\csname matched@csv/3/lazy_andor_always/calls\endcsname{2.196}
\expandafter\def\csname matched@csv/3/lazy_andor_always/queries\endcsname{3249.721}
\expandafter\def\csname matched@csv/3/lazy_andor_always/lifecycle_seconds\endcsname{12.654}
\expandafter\def\csname matched@csv/3/cape_none/lazy_andor_none/success_gain\endcsname{0.000}
\expandafter\def\csname matched@csv/3/cape_none/lazy_andor_none/success_gain_ci\endcsname{0.000, 0.000}
\expandafter\def\csname matched@csv/3/cape_none/lazy_andor_none/calls_saved\endcsname{0.000}
\expandafter\def\csname matched@csv/3/cape_none/lazy_andor_none/calls_saved_ci\endcsname{0.000, 0.000}
\expandafter\def\csname matched@csv/3/cape_none/lazy_andor_none/queries_saved\endcsname{-51.703}
\expandafter\def\csname matched@csv/3/cape_none/lazy_andor_none/queries_saved_ci\endcsname{-160.875, 5.766}
\expandafter\def\csname matched@csv/3/cape_none/lazy_andor_none/seconds_saved\endcsname{-0.173}
\expandafter\def\csname matched@csv/3/cape_none/lazy_andor_none/seconds_saved_ci\endcsname{-0.522, 0.061}
\expandafter\def\csname matched@csv/3/cape_selective/lazy_andor_selective/success_gain\endcsname{0.000}
\expandafter\def\csname matched@csv/3/cape_selective/lazy_andor_selective/success_gain_ci\endcsname{0.000, 0.000}
\expandafter\def\csname matched@csv/3/cape_selective/lazy_andor_selective/calls_saved\endcsname{0.000}
\expandafter\def\csname matched@csv/3/cape_selective/lazy_andor_selective/calls_saved_ci\endcsname{0.000, 0.000}
\expandafter\def\csname matched@csv/3/cape_selective/lazy_andor_selective/queries_saved\endcsname{270.585}
\expandafter\def\csname matched@csv/3/cape_selective/lazy_andor_selective/queries_saved_ci\endcsname{93.719, 438.217}
\expandafter\def\csname matched@csv/3/cape_selective/lazy_andor_selective/seconds_saved\endcsname{0.998}
\expandafter\def\csname matched@csv/3/cape_selective/lazy_andor_selective/seconds_saved_ci\endcsname{0.358, 1.596}
\expandafter\def\csname matched@csv/3/cape_always/lazy_andor_always/success_gain\endcsname{0.000}
\expandafter\def\csname matched@csv/3/cape_always/lazy_andor_always/success_gain_ci\endcsname{0.000, 0.000}
\expandafter\def\csname matched@csv/3/cape_always/lazy_andor_always/calls_saved\endcsname{0.000}
\expandafter\def\csname matched@csv/3/cape_always/lazy_andor_always/calls_saved_ci\endcsname{0.000, 0.000}
\expandafter\def\csname matched@csv/3/cape_always/lazy_andor_always/queries_saved\endcsname{703.366}
\expandafter\def\csname matched@csv/3/cape_always/lazy_andor_always/queries_saved_ci\endcsname{392.710, 982.935}
\expandafter\def\csname matched@csv/3/cape_always/lazy_andor_always/seconds_saved\endcsname{2.477}
\expandafter\def\csname matched@csv/3/cape_always/lazy_andor_always/seconds_saved_ci\endcsname{1.420, 3.479}
\expandafter\def\csname matched@csv/3/cape_selective/cape_none/success_gain\endcsname{0.000}
\expandafter\def\csname matched@csv/3/cape_selective/cape_none/success_gain_ci\endcsname{0.000, 0.000}
\expandafter\def\csname matched@csv/3/cape_selective/cape_none/calls_saved\endcsname{0.000}
\expandafter\def\csname matched@csv/3/cape_selective/cape_none/calls_saved_ci\endcsname{0.000, 0.000}
\expandafter\def\csname matched@csv/3/cape_selective/cape_none/queries_saved\endcsname{350.375}
\expandafter\def\csname matched@csv/3/cape_selective/cape_none/queries_saved_ci\endcsname{213.094, 473.812}
\expandafter\def\csname matched@csv/3/cape_selective/cape_none/seconds_saved\endcsname{0.571}
\expandafter\def\csname matched@csv/3/cape_selective/cape_none/seconds_saved_ci\endcsname{0.009, 1.130}
\expandafter\def\csname matched@csv/3/cape_selective/cape_always/success_gain\endcsname{0.000}
\expandafter\def\csname matched@csv/3/cape_selective/cape_always/success_gain_ci\endcsname{0.000, 0.000}
\expandafter\def\csname matched@csv/3/cape_selective/cape_always/calls_saved\endcsname{0.000}
\expandafter\def\csname matched@csv/3/cape_selective/cape_always/calls_saved_ci\endcsname{0.000, 0.000}
\expandafter\def\csname matched@csv/3/cape_selective/cape_always/queries_saved\endcsname{-433.578}
\expandafter\def\csname matched@csv/3/cape_selective/cape_always/queries_saved_ci\endcsname{-633.500, -226.797}
\expandafter\def\csname matched@csv/3/cape_selective/cape_always/seconds_saved\endcsname{-0.741}
\expandafter\def\csname matched@csv/3/cape_selective/cape_always/seconds_saved_ci\endcsname{-1.255, -0.242}
\expandafter\def\csname matched@csv/3/lazy_andor_selective/lazy_andor_none/success_gain\endcsname{0.000}
\expandafter\def\csname matched@csv/3/lazy_andor_selective/lazy_andor_none/success_gain_ci\endcsname{0.000, 0.000}
\expandafter\def\csname matched@csv/3/lazy_andor_selective/lazy_andor_none/calls_saved\endcsname{0.000}
\expandafter\def\csname matched@csv/3/lazy_andor_selective/lazy_andor_none/calls_saved_ci\endcsname{0.000, 0.000}
\expandafter\def\csname matched@csv/3/lazy_andor_selective/lazy_andor_none/queries_saved\endcsname{28.087}
\expandafter\def\csname matched@csv/3/lazy_andor_selective/lazy_andor_none/queries_saved_ci\endcsname{9.071, 59.150}
\expandafter\def\csname matched@csv/3/lazy_andor_selective/lazy_andor_none/seconds_saved\endcsname{-0.600}
\expandafter\def\csname matched@csv/3/lazy_andor_selective/lazy_andor_none/seconds_saved_ci\endcsname{-0.869, -0.291}
\expandafter\def\csname matched@csv/3/lazy_andor_selective/lazy_andor_always/success_gain\endcsname{0.000}
\expandafter\def\csname matched@csv/3/lazy_andor_selective/lazy_andor_always/success_gain_ci\endcsname{0.000, 0.000}
\expandafter\def\csname matched@csv/3/lazy_andor_selective/lazy_andor_always/calls_saved\endcsname{0.000}
\expandafter\def\csname matched@csv/3/lazy_andor_selective/lazy_andor_always/calls_saved_ci\endcsname{0.000, 0.000}
\expandafter\def\csname matched@csv/3/lazy_andor_selective/lazy_andor_always/queries_saved\endcsname{-0.797}
\expandafter\def\csname matched@csv/3/lazy_andor_selective/lazy_andor_always/queries_saved_ci\endcsname{-2.547, 0.250}
\expandafter\def\csname matched@csv/3/lazy_andor_selective/lazy_andor_always/seconds_saved\endcsname{0.738}
\expandafter\def\csname matched@csv/3/lazy_andor_selective/lazy_andor_always/seconds_saved_ci\endcsname{0.512, 0.925}
\expandafter\def\csname matched@csv/3/scheduler_reduction_min\endcsname{7.53}
\expandafter\def\csname matched@csv/3/scheduler_reduction_max\endcsname{10.83}
\expandafter\def\csname matched@csv/5/cape_none/success_rate\endcsname{98.438}
\expandafter\def\csname matched@csv/5/cape_none/calls\endcsname{2.348}
\expandafter\def\csname matched@csv/5/cape_none/queries\endcsname{3161.147}
\expandafter\def\csname matched@csv/5/cape_none/lifecycle_seconds\endcsname{11.374}
\expandafter\def\csname matched@csv/5/cape_selective/success_rate\endcsname{98.438}
\expandafter\def\csname matched@csv/5/cape_selective/calls\endcsname{2.348}
\expandafter\def\csname matched@csv/5/cape_selective/queries\endcsname{3161.147}
\expandafter\def\csname matched@csv/5/cape_selective/lifecycle_seconds\endcsname{11.319}
\expandafter\def\csname matched@csv/5/cape_always/success_rate\endcsname{98.438}
\expandafter\def\csname matched@csv/5/cape_always/calls\endcsname{2.348}
\expandafter\def\csname matched@csv/5/cape_always/queries\endcsname{3148.147}
\expandafter\def\csname matched@csv/5/cape_always/lifecycle_seconds\endcsname{12.464}
\expandafter\def\csname matched@csv/5/lazy_andor_none/success_rate\endcsname{98.438}
\expandafter\def\csname matched@csv/5/lazy_andor_none/calls\endcsname{2.348}
\expandafter\def\csname matched@csv/5/lazy_andor_none/queries\endcsname{3163.382}
\expandafter\def\csname matched@csv/5/lazy_andor_none/lifecycle_seconds\endcsname{11.148}
\expandafter\def\csname matched@csv/5/lazy_andor_selective/success_rate\endcsname{98.438}
\expandafter\def\csname matched@csv/5/lazy_andor_selective/calls\endcsname{2.348}
\expandafter\def\csname matched@csv/5/lazy_andor_selective/queries\endcsname{3163.382}
\expandafter\def\csname matched@csv/5/lazy_andor_selective/lifecycle_seconds\endcsname{11.265}
\expandafter\def\csname matched@csv/5/lazy_andor_always/success_rate\endcsname{98.438}
\expandafter\def\csname matched@csv/5/lazy_andor_always/calls\endcsname{2.348}
\expandafter\def\csname matched@csv/5/lazy_andor_always/queries\endcsname{3163.382}
\expandafter\def\csname matched@csv/5/lazy_andor_always/lifecycle_seconds\endcsname{11.381}
\expandafter\def\csname matched@csv/5/cape_none/lazy_andor_none/success_gain\endcsname{0.000}
\expandafter\def\csname matched@csv/5/cape_none/lazy_andor_none/success_gain_ci\endcsname{0.000, 0.000}
\expandafter\def\csname matched@csv/5/cape_none/lazy_andor_none/calls_saved\endcsname{0.000}
\expandafter\def\csname matched@csv/5/cape_none/lazy_andor_none/calls_saved_ci\endcsname{0.000, 0.000}
\expandafter\def\csname matched@csv/5/cape_none/lazy_andor_none/queries_saved\endcsname{2.234}
\expandafter\def\csname matched@csv/5/cape_none/lazy_andor_none/queries_saved_ci\endcsname{0.000, 6.703}
\expandafter\def\csname matched@csv/5/cape_none/lazy_andor_none/seconds_saved\endcsname{-0.226}
\expandafter\def\csname matched@csv/5/cape_none/lazy_andor_none/seconds_saved_ci\endcsname{-0.357, -0.095}
\expandafter\def\csname matched@csv/5/cape_selective/lazy_andor_selective/success_gain\endcsname{0.000}
\expandafter\def\csname matched@csv/5/cape_selective/lazy_andor_selective/success_gain_ci\endcsname{0.000, 0.000}
\expandafter\def\csname matched@csv/5/cape_selective/lazy_andor_selective/calls_saved\endcsname{0.000}
\expandafter\def\csname matched@csv/5/cape_selective/lazy_andor_selective/calls_saved_ci\endcsname{0.000, 0.000}
\expandafter\def\csname matched@csv/5/cape_selective/lazy_andor_selective/queries_saved\endcsname{2.234}
\expandafter\def\csname matched@csv/5/cape_selective/lazy_andor_selective/queries_saved_ci\endcsname{0.000, 6.703}
\expandafter\def\csname matched@csv/5/cape_selective/lazy_andor_selective/seconds_saved\endcsname{-0.054}
\expandafter\def\csname matched@csv/5/cape_selective/lazy_andor_selective/seconds_saved_ci\endcsname{-0.273, 0.163}
\expandafter\def\csname matched@csv/5/cape_always/lazy_andor_always/success_gain\endcsname{0.000}
\expandafter\def\csname matched@csv/5/cape_always/lazy_andor_always/success_gain_ci\endcsname{0.000, 0.000}
\expandafter\def\csname matched@csv/5/cape_always/lazy_andor_always/calls_saved\endcsname{0.000}
\expandafter\def\csname matched@csv/5/cape_always/lazy_andor_always/calls_saved_ci\endcsname{0.000, 0.000}
\expandafter\def\csname matched@csv/5/cape_always/lazy_andor_always/queries_saved\endcsname{15.234}
\expandafter\def\csname matched@csv/5/cape_always/lazy_andor_always/queries_saved_ci\endcsname{0.000, 45.703}
\expandafter\def\csname matched@csv/5/cape_always/lazy_andor_always/seconds_saved\endcsname{-1.082}
\expandafter\def\csname matched@csv/5/cape_always/lazy_andor_always/seconds_saved_ci\endcsname{-2.013, -0.421}
\expandafter\def\csname matched@csv/5/cape_selective/cape_none/success_gain\endcsname{0.000}
\expandafter\def\csname matched@csv/5/cape_selective/cape_none/success_gain_ci\endcsname{0.000, 0.000}
\expandafter\def\csname matched@csv/5/cape_selective/cape_none/calls_saved\endcsname{0.000}
\expandafter\def\csname matched@csv/5/cape_selective/cape_none/calls_saved_ci\endcsname{0.000, 0.000}
\expandafter\def\csname matched@csv/5/cape_selective/cape_none/queries_saved\endcsname{0.000}
\expandafter\def\csname matched@csv/5/cape_selective/cape_none/queries_saved_ci\endcsname{0.000, 0.000}
\expandafter\def\csname matched@csv/5/cape_selective/cape_none/seconds_saved\endcsname{0.055}
\expandafter\def\csname matched@csv/5/cape_selective/cape_none/seconds_saved_ci\endcsname{-0.160, 0.278}
\expandafter\def\csname matched@csv/5/cape_selective/cape_always/success_gain\endcsname{0.000}
\expandafter\def\csname matched@csv/5/cape_selective/cape_always/success_gain_ci\endcsname{0.000, 0.000}
\expandafter\def\csname matched@csv/5/cape_selective/cape_always/calls_saved\endcsname{0.000}
\expandafter\def\csname matched@csv/5/cape_selective/cape_always/calls_saved_ci\endcsname{0.000, 0.000}
\expandafter\def\csname matched@csv/5/cape_selective/cape_always/queries_saved\endcsname{-13.000}
\expandafter\def\csname matched@csv/5/cape_selective/cape_always/queries_saved_ci\endcsname{-39.000, 0.000}
\expandafter\def\csname matched@csv/5/cape_selective/cape_always/seconds_saved\endcsname{1.145}
\expandafter\def\csname matched@csv/5/cape_selective/cape_always/seconds_saved_ci\endcsname{0.413, 1.966}
\expandafter\def\csname matched@csv/5/lazy_andor_selective/lazy_andor_none/success_gain\endcsname{0.000}
\expandafter\def\csname matched@csv/5/lazy_andor_selective/lazy_andor_none/success_gain_ci\endcsname{0.000, 0.000}
\expandafter\def\csname matched@csv/5/lazy_andor_selective/lazy_andor_none/calls_saved\endcsname{0.000}
\expandafter\def\csname matched@csv/5/lazy_andor_selective/lazy_andor_none/calls_saved_ci\endcsname{0.000, 0.000}
\expandafter\def\csname matched@csv/5/lazy_andor_selective/lazy_andor_none/queries_saved\endcsname{0.000}
\expandafter\def\csname matched@csv/5/lazy_andor_selective/lazy_andor_none/queries_saved_ci\endcsname{0.000, 0.000}
\expandafter\def\csname matched@csv/5/lazy_andor_selective/lazy_andor_none/seconds_saved\endcsname{-0.117}
\expandafter\def\csname matched@csv/5/lazy_andor_selective/lazy_andor_none/seconds_saved_ci\endcsname{-0.275, 0.039}
\expandafter\def\csname matched@csv/5/lazy_andor_selective/lazy_andor_always/success_gain\endcsname{0.000}
\expandafter\def\csname matched@csv/5/lazy_andor_selective/lazy_andor_always/success_gain_ci\endcsname{0.000, 0.000}
\expandafter\def\csname matched@csv/5/lazy_andor_selective/lazy_andor_always/calls_saved\endcsname{0.000}
\expandafter\def\csname matched@csv/5/lazy_andor_selective/lazy_andor_always/calls_saved_ci\endcsname{0.000, 0.000}
\expandafter\def\csname matched@csv/5/lazy_andor_selective/lazy_andor_always/queries_saved\endcsname{0.000}
\expandafter\def\csname matched@csv/5/lazy_andor_selective/lazy_andor_always/queries_saved_ci\endcsname{0.000, 0.000}
\expandafter\def\csname matched@csv/5/lazy_andor_selective/lazy_andor_always/seconds_saved\endcsname{0.117}
\expandafter\def\csname matched@csv/5/lazy_andor_selective/lazy_andor_always/seconds_saved_ci\endcsname{-0.134, 0.391}
\expandafter\def\csname matched@csv/5/scheduler_reduction_min\endcsname{-0.94}
\expandafter\def\csname matched@csv/5/scheduler_reduction_max\endcsname{0.60}
\expandafter\def\csname matched@json/3/cape_none/success_rate\endcsname{94.744}
\expandafter\def\csname matched@json/3/cape_none/calls\endcsname{1.477}
\expandafter\def\csname matched@json/3/cape_none/queries\endcsname{1505.717}
\expandafter\def\csname matched@json/3/cape_none/lifecycle_seconds\endcsname{12.007}
\expandafter\def\csname matched@json/3/cape_selective/success_rate\endcsname{94.744}
\expandafter\def\csname matched@json/3/cape_selective/calls\endcsname{1.476}
\expandafter\def\csname matched@json/3/cape_selective/queries\endcsname{1511.926}
\expandafter\def\csname matched@json/3/cape_selective/lifecycle_seconds\endcsname{11.961}
\expandafter\def\csname matched@json/3/cape_always/success_rate\endcsname{94.744}
\expandafter\def\csname matched@json/3/cape_always/calls\endcsname{1.476}
\expandafter\def\csname matched@json/3/cape_always/queries\endcsname{1504.997}
\expandafter\def\csname matched@json/3/cape_always/lifecycle_seconds\endcsname{12.012}
\expandafter\def\csname matched@json/3/lazy_andor_none/success_rate\endcsname{94.389}
\expandafter\def\csname matched@json/3/lazy_andor_none/calls\endcsname{1.481}
\expandafter\def\csname matched@json/3/lazy_andor_none/queries\endcsname{1980.328}
\expandafter\def\csname matched@json/3/lazy_andor_none/lifecycle_seconds\endcsname{16.130}
\expandafter\def\csname matched@json/3/lazy_andor_selective/success_rate\endcsname{94.389}
\expandafter\def\csname matched@json/3/lazy_andor_selective/calls\endcsname{1.478}
\expandafter\def\csname matched@json/3/lazy_andor_selective/queries\endcsname{1990.548}
\expandafter\def\csname matched@json/3/lazy_andor_selective/lifecycle_seconds\endcsname{16.161}
\expandafter\def\csname matched@json/3/lazy_andor_always/success_rate\endcsname{94.460}
\expandafter\def\csname matched@json/3/lazy_andor_always/calls\endcsname{1.483}
\expandafter\def\csname matched@json/3/lazy_andor_always/queries\endcsname{1975.267}
\expandafter\def\csname matched@json/3/lazy_andor_always/lifecycle_seconds\endcsname{16.235}
\expandafter\def\csname matched@json/3/cape_none/lazy_andor_none/success_gain\endcsname{0.355}
\expandafter\def\csname matched@json/3/cape_none/lazy_andor_none/success_gain_ci\endcsname{0.000, 0.781}
\expandafter\def\csname matched@json/3/cape_none/lazy_andor_none/calls_saved\endcsname{0.005}
\expandafter\def\csname matched@json/3/cape_none/lazy_andor_none/calls_saved_ci\endcsname{-0.006, 0.021}
\expandafter\def\csname matched@json/3/cape_none/lazy_andor_none/queries_saved\endcsname{474.612}
\expandafter\def\csname matched@json/3/cape_none/lazy_andor_none/queries_saved_ci\endcsname{280.981, 668.547}
\expandafter\def\csname matched@json/3/cape_none/lazy_andor_none/seconds_saved\endcsname{4.123}
\expandafter\def\csname matched@json/3/cape_none/lazy_andor_none/seconds_saved_ci\endcsname{2.785, 5.446}
\expandafter\def\csname matched@json/3/cape_selective/lazy_andor_selective/success_gain\endcsname{0.355}
\expandafter\def\csname matched@json/3/cape_selective/lazy_andor_selective/success_gain_ci\endcsname{0.000, 0.781}
\expandafter\def\csname matched@json/3/cape_selective/lazy_andor_selective/calls_saved\endcsname{0.002}
\expandafter\def\csname matched@json/3/cape_selective/lazy_andor_selective/calls_saved_ci\endcsname{-0.006, 0.013}
\expandafter\def\csname matched@json/3/cape_selective/lazy_andor_selective/queries_saved\endcsname{478.622}
\expandafter\def\csname matched@json/3/cape_selective/lazy_andor_selective/queries_saved_ci\endcsname{275.618, 676.500}
\expandafter\def\csname matched@json/3/cape_selective/lazy_andor_selective/seconds_saved\endcsname{4.200}
\expandafter\def\csname matched@json/3/cape_selective/lazy_andor_selective/seconds_saved_ci\endcsname{2.839, 5.570}
\expandafter\def\csname matched@json/3/cape_always/lazy_andor_always/success_gain\endcsname{0.284}
\expandafter\def\csname matched@json/3/cape_always/lazy_andor_always/success_gain_ci\endcsname{0.000, 0.568}
\expandafter\def\csname matched@json/3/cape_always/lazy_andor_always/calls_saved\endcsname{0.006}
\expandafter\def\csname matched@json/3/cape_always/lazy_andor_always/calls_saved_ci\endcsname{-0.004, 0.023}
\expandafter\def\csname matched@json/3/cape_always/lazy_andor_always/queries_saved\endcsname{470.270}
\expandafter\def\csname matched@json/3/cape_always/lazy_andor_always/queries_saved_ci\endcsname{273.665, 670.398}
\expandafter\def\csname matched@json/3/cape_always/lazy_andor_always/seconds_saved\endcsname{4.223}
\expandafter\def\csname matched@json/3/cape_always/lazy_andor_always/seconds_saved_ci\endcsname{2.860, 5.630}
\expandafter\def\csname matched@json/3/cape_selective/cape_none/success_gain\endcsname{0.000}
\expandafter\def\csname matched@json/3/cape_selective/cape_none/success_gain_ci\endcsname{0.000, 0.000}
\expandafter\def\csname matched@json/3/cape_selective/cape_none/calls_saved\endcsname{0.000}
\expandafter\def\csname matched@json/3/cape_selective/cape_none/calls_saved_ci\endcsname{0.000, 0.001}
\expandafter\def\csname matched@json/3/cape_selective/cape_none/queries_saved\endcsname{-6.210}
\expandafter\def\csname matched@json/3/cape_selective/cape_none/queries_saved_ci\endcsname{-23.719, 4.152}
\expandafter\def\csname matched@json/3/cape_selective/cape_none/seconds_saved\endcsname{0.046}
\expandafter\def\csname matched@json/3/cape_selective/cape_none/seconds_saved_ci\endcsname{-0.138, 0.268}
\expandafter\def\csname matched@json/3/cape_selective/cape_always/success_gain\endcsname{0.000}
\expandafter\def\csname matched@json/3/cape_selective/cape_always/success_gain_ci\endcsname{0.000, 0.000}
\expandafter\def\csname matched@json/3/cape_selective/cape_always/calls_saved\endcsname{0.000}
\expandafter\def\csname matched@json/3/cape_selective/cape_always/calls_saved_ci\endcsname{0.000, 0.000}
\expandafter\def\csname matched@json/3/cape_selective/cape_always/queries_saved\endcsname{-6.930}
\expandafter\def\csname matched@json/3/cape_selective/cape_always/queries_saved_ci\endcsname{-20.789, 0.000}
\expandafter\def\csname matched@json/3/cape_selective/cape_always/seconds_saved\endcsname{0.051}
\expandafter\def\csname matched@json/3/cape_selective/cape_always/seconds_saved_ci\endcsname{-0.097, 0.176}
\expandafter\def\csname matched@json/3/lazy_andor_selective/lazy_andor_none/success_gain\endcsname{0.000}
\expandafter\def\csname matched@json/3/lazy_andor_selective/lazy_andor_none/success_gain_ci\endcsname{0.000, 0.000}
\expandafter\def\csname matched@json/3/lazy_andor_selective/lazy_andor_none/calls_saved\endcsname{0.003}
\expandafter\def\csname matched@json/3/lazy_andor_selective/lazy_andor_none/calls_saved_ci\endcsname{0.000, 0.010}
\expandafter\def\csname matched@json/3/lazy_andor_selective/lazy_andor_none/queries_saved\endcsname{-10.220}
\expandafter\def\csname matched@json/3/lazy_andor_selective/lazy_andor_none/queries_saved_ci\endcsname{-43.820, 13.326}
\expandafter\def\csname matched@json/3/lazy_andor_selective/lazy_andor_none/seconds_saved\endcsname{-0.031}
\expandafter\def\csname matched@json/3/lazy_andor_selective/lazy_andor_none/seconds_saved_ci\endcsname{-0.203, 0.145}
\expandafter\def\csname matched@json/3/lazy_andor_selective/lazy_andor_always/success_gain\endcsname{-0.071}
\expandafter\def\csname matched@json/3/lazy_andor_selective/lazy_andor_always/success_gain_ci\endcsname{-0.213, 0.000}
\expandafter\def\csname matched@json/3/lazy_andor_selective/lazy_andor_always/calls_saved\endcsname{0.004}
\expandafter\def\csname matched@json/3/lazy_andor_selective/lazy_andor_always/calls_saved_ci\endcsname{0.000, 0.011}
\expandafter\def\csname matched@json/3/lazy_andor_selective/lazy_andor_always/queries_saved\endcsname{-15.281}
\expandafter\def\csname matched@json/3/lazy_andor_selective/lazy_andor_always/queries_saved_ci\endcsname{-53.156, 7.312}
\expandafter\def\csname matched@json/3/lazy_andor_selective/lazy_andor_always/seconds_saved\endcsname{0.074}
\expandafter\def\csname matched@json/3/lazy_andor_selective/lazy_andor_always/seconds_saved_ci\endcsname{-0.103, 0.239}
\expandafter\def\csname matched@json/3/scheduler_reduction_min\endcsname{22.51}
\expandafter\def\csname matched@json/3/scheduler_reduction_max\endcsname{26.04}
\expandafter\def\csname matched@json/5/cape_none/success_rate\endcsname{94.744}
\expandafter\def\csname matched@json/5/cape_none/calls\endcsname{1.475}
\expandafter\def\csname matched@json/5/cape_none/queries\endcsname{1493.963}
\expandafter\def\csname matched@json/5/cape_none/lifecycle_seconds\endcsname{11.975}
\expandafter\def\csname matched@json/5/cape_selective/success_rate\endcsname{94.744}
\expandafter\def\csname matched@json/5/cape_selective/calls\endcsname{1.475}
\expandafter\def\csname matched@json/5/cape_selective/queries\endcsname{1490.510}
\expandafter\def\csname matched@json/5/cape_selective/lifecycle_seconds\endcsname{11.929}
\expandafter\def\csname matched@json/5/cape_always/success_rate\endcsname{94.744}
\expandafter\def\csname matched@json/5/cape_always/calls\endcsname{1.475}
\expandafter\def\csname matched@json/5/cape_always/queries\endcsname{1473.893}
\expandafter\def\csname matched@json/5/cape_always/lifecycle_seconds\endcsname{12.005}
\expandafter\def\csname matched@json/5/lazy_andor_none/success_rate\endcsname{94.496}
\expandafter\def\csname matched@json/5/lazy_andor_none/calls\endcsname{1.483}
\expandafter\def\csname matched@json/5/lazy_andor_none/queries\endcsname{1941.740}
\expandafter\def\csname matched@json/5/lazy_andor_none/lifecycle_seconds\endcsname{15.784}
\expandafter\def\csname matched@json/5/lazy_andor_selective/success_rate\endcsname{94.531}
\expandafter\def\csname matched@json/5/lazy_andor_selective/calls\endcsname{1.483}
\expandafter\def\csname matched@json/5/lazy_andor_selective/queries\endcsname{1949.443}
\expandafter\def\csname matched@json/5/lazy_andor_selective/lifecycle_seconds\endcsname{15.728}
\expandafter\def\csname matched@json/5/lazy_andor_always/success_rate\endcsname{94.425}
\expandafter\def\csname matched@json/5/lazy_andor_always/calls\endcsname{1.480}
\expandafter\def\csname matched@json/5/lazy_andor_always/queries\endcsname{1932.552}
\expandafter\def\csname matched@json/5/lazy_andor_always/lifecycle_seconds\endcsname{15.699}
\expandafter\def\csname matched@json/5/cape_none/lazy_andor_none/success_gain\endcsname{0.249}
\expandafter\def\csname matched@json/5/cape_none/lazy_andor_none/success_gain_ci\endcsname{0.000, 0.568}
\expandafter\def\csname matched@json/5/cape_none/lazy_andor_none/calls_saved\endcsname{0.007}
\expandafter\def\csname matched@json/5/cape_none/lazy_andor_none/calls_saved_ci\endcsname{-0.004, 0.026}
\expandafter\def\csname matched@json/5/cape_none/lazy_andor_none/queries_saved\endcsname{447.777}
\expandafter\def\csname matched@json/5/cape_none/lazy_andor_none/queries_saved_ci\endcsname{297.969, 603.558}
\expandafter\def\csname matched@json/5/cape_none/lazy_andor_none/seconds_saved\endcsname{3.809}
\expandafter\def\csname matched@json/5/cape_none/lazy_andor_none/seconds_saved_ci\endcsname{2.348, 5.302}
\expandafter\def\csname matched@json/5/cape_selective/lazy_andor_selective/success_gain\endcsname{0.213}
\expandafter\def\csname matched@json/5/cape_selective/lazy_andor_selective/success_gain_ci\endcsname{0.000, 0.497}
\expandafter\def\csname matched@json/5/cape_selective/lazy_andor_selective/calls_saved\endcsname{0.008}
\expandafter\def\csname matched@json/5/cape_selective/lazy_andor_selective/calls_saved_ci\endcsname{-0.004, 0.028}
\expandafter\def\csname matched@json/5/cape_selective/lazy_andor_selective/queries_saved\endcsname{458.933}
\expandafter\def\csname matched@json/5/cape_selective/lazy_andor_selective/queries_saved_ci\endcsname{305.508, 613.542}
\expandafter\def\csname matched@json/5/cape_selective/lazy_andor_selective/seconds_saved\endcsname{3.799}
\expandafter\def\csname matched@json/5/cape_selective/lazy_andor_selective/seconds_saved_ci\endcsname{2.377, 5.228}
\expandafter\def\csname matched@json/5/cape_always/lazy_andor_always/success_gain\endcsname{0.320}
\expandafter\def\csname matched@json/5/cape_always/lazy_andor_always/success_gain_ci\endcsname{0.000, 0.710}
\expandafter\def\csname matched@json/5/cape_always/lazy_andor_always/calls_saved\endcsname{0.005}
\expandafter\def\csname matched@json/5/cape_always/lazy_andor_always/calls_saved_ci\endcsname{-0.004, 0.019}
\expandafter\def\csname matched@json/5/cape_always/lazy_andor_always/queries_saved\endcsname{458.660}
\expandafter\def\csname matched@json/5/cape_always/lazy_andor_always/queries_saved_ci\endcsname{305.312, 613.152}
\expandafter\def\csname matched@json/5/cape_always/lazy_andor_always/seconds_saved\endcsname{3.694}
\expandafter\def\csname matched@json/5/cape_always/lazy_andor_always/seconds_saved_ci\endcsname{2.332, 5.081}
\expandafter\def\csname matched@json/5/cape_selective/cape_none/success_gain\endcsname{0.000}
\expandafter\def\csname matched@json/5/cape_selective/cape_none/success_gain_ci\endcsname{0.000, 0.000}
\expandafter\def\csname matched@json/5/cape_selective/cape_none/calls_saved\endcsname{0.000}
\expandafter\def\csname matched@json/5/cape_selective/cape_none/calls_saved_ci\endcsname{0.000, 0.000}
\expandafter\def\csname matched@json/5/cape_selective/cape_none/queries_saved\endcsname{3.453}
\expandafter\def\csname matched@json/5/cape_selective/cape_none/queries_saved_ci\endcsname{0.000, 10.359}
\expandafter\def\csname matched@json/5/cape_selective/cape_none/seconds_saved\endcsname{0.046}
\expandafter\def\csname matched@json/5/cape_selective/cape_none/seconds_saved_ci\endcsname{-0.070, 0.164}
\expandafter\def\csname matched@json/5/cape_selective/cape_always/success_gain\endcsname{0.000}
\expandafter\def\csname matched@json/5/cape_selective/cape_always/success_gain_ci\endcsname{0.000, 0.000}
\expandafter\def\csname matched@json/5/cape_selective/cape_always/calls_saved\endcsname{0.000}
\expandafter\def\csname matched@json/5/cape_selective/cape_always/calls_saved_ci\endcsname{0.000, 0.000}
\expandafter\def\csname matched@json/5/cape_selective/cape_always/queries_saved\endcsname{-16.617}
\expandafter\def\csname matched@json/5/cape_selective/cape_always/queries_saved_ci\endcsname{-49.852, 0.000}
\expandafter\def\csname matched@json/5/cape_selective/cape_always/seconds_saved\endcsname{0.076}
\expandafter\def\csname matched@json/5/cape_selective/cape_always/seconds_saved_ci\endcsname{-0.066, 0.210}
\expandafter\def\csname matched@json/5/lazy_andor_selective/lazy_andor_none/success_gain\endcsname{0.036}
\expandafter\def\csname matched@json/5/lazy_andor_selective/lazy_andor_none/success_gain_ci\endcsname{0.000, 0.107}
\expandafter\def\csname matched@json/5/lazy_andor_selective/lazy_andor_none/calls_saved\endcsname{-0.001}
\expandafter\def\csname matched@json/5/lazy_andor_selective/lazy_andor_none/calls_saved_ci\endcsname{-0.002, 0.000}
\expandafter\def\csname matched@json/5/lazy_andor_selective/lazy_andor_none/queries_saved\endcsname{-7.703}
\expandafter\def\csname matched@json/5/lazy_andor_selective/lazy_andor_none/queries_saved_ci\endcsname{-16.406, 0.000}
\expandafter\def\csname matched@json/5/lazy_andor_selective/lazy_andor_none/seconds_saved\endcsname{0.056}
\expandafter\def\csname matched@json/5/lazy_andor_selective/lazy_andor_none/seconds_saved_ci\endcsname{-0.100, 0.207}
\expandafter\def\csname matched@json/5/lazy_andor_selective/lazy_andor_always/success_gain\endcsname{0.107}
\expandafter\def\csname matched@json/5/lazy_andor_selective/lazy_andor_always/success_gain_ci\endcsname{0.000, 0.320}
\expandafter\def\csname matched@json/5/lazy_andor_selective/lazy_andor_always/calls_saved\endcsname{-0.003}
\expandafter\def\csname matched@json/5/lazy_andor_selective/lazy_andor_always/calls_saved_ci\endcsname{-0.009, 0.000}
\expandafter\def\csname matched@json/5/lazy_andor_selective/lazy_andor_always/queries_saved\endcsname{-16.891}
\expandafter\def\csname matched@json/5/lazy_andor_selective/lazy_andor_always/queries_saved_ci\endcsname{-50.281, 0.000}
\expandafter\def\csname matched@json/5/lazy_andor_selective/lazy_andor_always/seconds_saved\endcsname{-0.029}
\expandafter\def\csname matched@json/5/lazy_andor_selective/lazy_andor_always/seconds_saved_ci\endcsname{-0.217, 0.151}
\expandafter\def\csname matched@json/5/scheduler_reduction_min\endcsname{19.52}
\expandafter\def\csname matched@json/5/scheduler_reduction_max\endcsname{24.18}
\expandafter\def\csname matched@profile/csv/3/cape_none/calibration\endcsname{9.612}
\expandafter\def\csname matched@profile/csv/3/cape_none/tight\endcsname{0.000}
\expandafter\def\csname matched@profile/csv/3/cape_selective/calibration\endcsname{8.814}
\expandafter\def\csname matched@profile/csv/3/cape_selective/tight\endcsname{0.621}
\expandafter\def\csname matched@profile/csv/3/cape_always/calibration\endcsname{7.658}
\expandafter\def\csname matched@profile/csv/3/cape_always/tight\endcsname{1.468}
\expandafter\def\csname matched@profile/csv/3/lazy_andor_none/calibration\endcsname{9.492}
\expandafter\def\csname matched@profile/csv/3/lazy_andor_none/tight\endcsname{0.000}
\expandafter\def\csname matched@profile/csv/3/lazy_andor_selective/calibration\endcsname{9.571}
\expandafter\def\csname matched@profile/csv/3/lazy_andor_selective/tight\endcsname{1.290}
\expandafter\def\csname matched@profile/csv/3/lazy_andor_always/calibration\endcsname{9.620}
\expandafter\def\csname matched@profile/csv/3/lazy_andor_always/tight\endcsname{2.266}
\expandafter\def\csname matched@profile/csv/5/cape_none/calibration\endcsname{9.155}
\expandafter\def\csname matched@profile/csv/5/cape_none/tight\endcsname{0.000}
\expandafter\def\csname matched@profile/csv/5/cape_selective/calibration\endcsname{9.259}
\expandafter\def\csname matched@profile/csv/5/cape_selective/tight\endcsname{0.019}
\expandafter\def\csname matched@profile/csv/5/cape_always/calibration\endcsname{9.148}
\expandafter\def\csname matched@profile/csv/5/cape_always/tight\endcsname{1.192}
\expandafter\def\csname matched@profile/csv/5/lazy_andor_none/calibration\endcsname{9.265}
\expandafter\def\csname matched@profile/csv/5/lazy_andor_none/tight\endcsname{0.000}
\expandafter\def\csname matched@profile/csv/5/lazy_andor_selective/calibration\endcsname{9.243}
\expandafter\def\csname matched@profile/csv/5/lazy_andor_selective/tight\endcsname{0.000}
\expandafter\def\csname matched@profile/csv/5/lazy_andor_always/calibration\endcsname{9.315}
\expandafter\def\csname matched@profile/csv/5/lazy_andor_always/tight\endcsname{0.265}
\expandafter\def\csname matched@profile/json/3/cape_none/calibration\endcsname{11.307}
\expandafter\def\csname matched@profile/json/3/cape_none/tight\endcsname{0.000}
\expandafter\def\csname matched@profile/json/3/cape_selective/calibration\endcsname{11.139}
\expandafter\def\csname matched@profile/json/3/cape_selective/tight\endcsname{0.070}
\expandafter\def\csname matched@profile/json/3/cape_always/calibration\endcsname{11.183}
\expandafter\def\csname matched@profile/json/3/cape_always/tight\endcsname{0.100}
\expandafter\def\csname matched@profile/json/3/lazy_andor_none/calibration\endcsname{15.561}
\expandafter\def\csname matched@profile/json/3/lazy_andor_none/tight\endcsname{0.000}
\expandafter\def\csname matched@profile/json/3/lazy_andor_selective/calibration\endcsname{15.541}
\expandafter\def\csname matched@profile/json/3/lazy_andor_selective/tight\endcsname{0.146}
\expandafter\def\csname matched@profile/json/3/lazy_andor_always/calibration\endcsname{15.581}
\expandafter\def\csname matched@profile/json/3/lazy_andor_always/tight\endcsname{0.154}
\expandafter\def\csname matched@profile/json/5/cape_none/calibration\endcsname{11.282}
\expandafter\def\csname matched@profile/json/5/cape_none/tight\endcsname{0.000}
\expandafter\def\csname matched@profile/json/5/cape_selective/calibration\endcsname{11.364}
\expandafter\def\csname matched@profile/json/5/cape_selective/tight\endcsname{0.004}
\expandafter\def\csname matched@profile/json/5/cape_always/calibration\endcsname{11.336}
\expandafter\def\csname matched@profile/json/5/cape_always/tight\endcsname{0.006}
\expandafter\def\csname matched@profile/json/5/lazy_andor_none/calibration\endcsname{14.989}
\expandafter\def\csname matched@profile/json/5/lazy_andor_none/tight\endcsname{0.000}
\expandafter\def\csname matched@profile/json/5/lazy_andor_selective/calibration\endcsname{15.188}
\expandafter\def\csname matched@profile/json/5/lazy_andor_selective/tight\endcsname{0.000}
\expandafter\def\csname matched@profile/json/5/lazy_andor_always/calibration\endcsname{15.025}
\expandafter\def\csname matched@profile/json/5/lazy_andor_always/tight\endcsname{0.000}

\newcommand{\MVal}[1]{\csname panel@#1\endcsname}
\expandafter\def\csname panel@models\endcsname{14}
\expandafter\def\csname panel@requests\endcsname{896}
\expandafter\def\csname panel@new_requests\endcsname{832}
\expandafter\def\csname panel@valid_proposals\endcsname{874}
\expandafter\def\csname panel@reference_agreement\endcsname{862}
\expandafter\def\csname panel@tokens\endcsname{804,853}
\expandafter\def\csname panel@new_tokens\endcsname{750,637}
\expandafter\def\csname panel@physical_records\endcsname{4,864}
\expandafter\def\csname panel@expanded_records\endcsname{57,344}
\expandafter\def\csname panel@candidate_records\endcsname{608}
\expandafter\def\csname panel@candidate_optimal\endcsname{606}
\expandafter\def\csname panel@candidate_certified\endcsname{595}
\expandafter\def\csname panel@candidate_reference_records\endcsname{528}
\expandafter\def\csname panel@candidate_reference_optimal\endcsname{526}
\expandafter\def\csname panel@native_models\endcsname{152}
\expandafter\def\csname panel@structures\endcsname{33}
\expandafter\def\csname panel@policy_pairs\endcsname{655}
\expandafter\def\csname panel@objectives\endcsname{304}
\expandafter\def\csname panel@resolved_objectives\endcsname{304}
\expandafter\def\csname panel@history_witnesses\endcsname{10,225}
\expandafter\def\csname panel@native_episodes\endcsname{94,160}
\expandafter\def\csname panel@native_calls\endcsname{218,266}
\expandafter\def\csname panel@transport_errors\endcsname{18}
\expandafter\def\csname panel@unobserved_requests\endcsname{2}
\expandafter\def\csname panel@csv/3/eager_optimized/success_positive_models\endcsname{0}
\expandafter\def\csname panel@csv/3/eager_optimized/success_gain/min\endcsname{0.00}
\expandafter\def\csname panel@csv/3/eager_optimized/success_gain/max\endcsname{0.00}
\expandafter\def\csname panel@csv/3/eager_optimized/calls_saved/min\endcsname{0.000}
\expandafter\def\csname panel@csv/3/eager_optimized/calls_saved/max\endcsname{0.000}
\expandafter\def\csname panel@csv/3/eager_optimized/seconds_saved/min\endcsname{0.401}
\expandafter\def\csname panel@csv/3/eager_optimized/seconds_saved/max\endcsname{0.421}
\expandafter\def\csname panel@csv/3/eager_optimized/queries_saved/min\endcsname{368.2}
\expandafter\def\csname panel@csv/3/eager_optimized/queries_saved/max\endcsname{470.4}
\expandafter\def\csname panel@csv/3/eager_optimized/time_reduction/min\endcsname{3.6}
\expandafter\def\csname panel@csv/3/eager_optimized/time_reduction/max\endcsname{4.5}
\expandafter\def\csname panel@csv/3/initial_safe_two_step/success_positive_models\endcsname{13}
\expandafter\def\csname panel@csv/3/initial_safe_two_step/success_gain/min\endcsname{0.00}
\expandafter\def\csname panel@csv/3/initial_safe_two_step/success_gain/max\endcsname{1.17}
\expandafter\def\csname panel@csv/3/initial_safe_two_step/calls_saved/min\endcsname{0.203}
\expandafter\def\csname panel@csv/3/initial_safe_two_step/calls_saved/max\endcsname{0.278}
\expandafter\def\csname panel@csv/3/initial_safe_two_step/seconds_saved/min\endcsname{0.349}
\expandafter\def\csname panel@csv/3/initial_safe_two_step/seconds_saved/max\endcsname{0.395}
\expandafter\def\csname panel@csv/3/initial_safe_two_step/queries_saved/min\endcsname{368.2}
\expandafter\def\csname panel@csv/3/initial_safe_two_step/queries_saved/max\endcsname{470.4}
\expandafter\def\csname panel@csv/3/initial_safe_two_step/time_reduction/min\endcsname{3.2}
\expandafter\def\csname panel@csv/3/initial_safe_two_step/time_reduction/max\endcsname{4.3}
\expandafter\def\csname panel@csv/3/native_program_only/success_positive_models\endcsname{14}
\expandafter\def\csname panel@csv/3/native_program_only/success_gain/min\endcsname{4.69}
\expandafter\def\csname panel@csv/3/native_program_only/success_gain/max\endcsname{8.40}
\expandafter\def\csname panel@csv/3/native_program_only/calls_saved/min\endcsname{-0.174}
\expandafter\def\csname panel@csv/3/native_program_only/calls_saved/max\endcsname{-0.133}
\expandafter\def\csname panel@csv/3/native_program_only/seconds_saved/min\endcsname{-7.136}
\expandafter\def\csname panel@csv/3/native_program_only/seconds_saved/max\endcsname{-5.081}
\expandafter\def\csname panel@csv/3/native_program_only/queries_saved/min\endcsname{-2356.5}
\expandafter\def\csname panel@csv/3/native_program_only/queries_saved/max\endcsname{-1703.5}
\expandafter\def\csname panel@csv/3/native_program_only/time_reduction/min\endcsname{-197.4}
\expandafter\def\csname panel@csv/3/native_program_only/time_reduction/max\endcsname{-139.3}
\expandafter\def\csname panel@csv/3/partial_native_full_rows/success_positive_models\endcsname{0}
\expandafter\def\csname panel@csv/3/partial_native_full_rows/success_gain/min\endcsname{0.00}
\expandafter\def\csname panel@csv/3/partial_native_full_rows/success_gain/max\endcsname{0.00}
\expandafter\def\csname panel@csv/3/partial_native_full_rows/calls_saved/min\endcsname{0.000}
\expandafter\def\csname panel@csv/3/partial_native_full_rows/calls_saved/max\endcsname{0.000}
\expandafter\def\csname panel@csv/3/partial_native_full_rows/seconds_saved/min\endcsname{0.440}
\expandafter\def\csname panel@csv/3/partial_native_full_rows/seconds_saved/max\endcsname{0.652}
\expandafter\def\csname panel@csv/3/partial_native_full_rows/queries_saved/min\endcsname{228.9}
\expandafter\def\csname panel@csv/3/partial_native_full_rows/queries_saved/max\endcsname{282.2}
\expandafter\def\csname panel@csv/3/partial_native_full_rows/time_reduction/min\endcsname{4.2}
\expandafter\def\csname panel@csv/3/partial_native_full_rows/time_reduction/max\endcsname{6.9}
\expandafter\def\csname panel@csv/3/partial_native_no_tightening/success_positive_models\endcsname{0}
\expandafter\def\csname panel@csv/3/partial_native_no_tightening/success_gain/min\endcsname{0.00}
\expandafter\def\csname panel@csv/3/partial_native_no_tightening/success_gain/max\endcsname{0.00}
\expandafter\def\csname panel@csv/3/partial_native_no_tightening/calls_saved/min\endcsname{0.000}
\expandafter\def\csname panel@csv/3/partial_native_no_tightening/calls_saved/max\endcsname{0.000}
\expandafter\def\csname panel@csv/3/partial_native_no_tightening/seconds_saved/min\endcsname{0.503}
\expandafter\def\csname panel@csv/3/partial_native_no_tightening/seconds_saved/max\endcsname{0.677}
\expandafter\def\csname panel@csv/3/partial_native_no_tightening/queries_saved/min\endcsname{314.3}
\expandafter\def\csname panel@csv/3/partial_native_no_tightening/queries_saved/max\endcsname{349.8}
\expandafter\def\csname panel@csv/3/partial_native_no_tightening/time_reduction/min\endcsname{4.8}
\expandafter\def\csname panel@csv/3/partial_native_no_tightening/time_reduction/max\endcsname{7.2}
\expandafter\def\csname panel@csv/5/eager_optimized/success_positive_models\endcsname{0}
\expandafter\def\csname panel@csv/5/eager_optimized/success_gain/min\endcsname{0.00}
\expandafter\def\csname panel@csv/5/eager_optimized/success_gain/max\endcsname{0.00}
\expandafter\def\csname panel@csv/5/eager_optimized/calls_saved/min\endcsname{0.000}
\expandafter\def\csname panel@csv/5/eager_optimized/calls_saved/max\endcsname{0.000}
\expandafter\def\csname panel@csv/5/eager_optimized/seconds_saved/min\endcsname{0.018}
\expandafter\def\csname panel@csv/5/eager_optimized/seconds_saved/max\endcsname{0.235}
\expandafter\def\csname panel@csv/5/eager_optimized/queries_saved/min\endcsname{148.3}
\expandafter\def\csname panel@csv/5/eager_optimized/queries_saved/max\endcsname{289.7}
\expandafter\def\csname panel@csv/5/eager_optimized/time_reduction/min\endcsname{0.2}
\expandafter\def\csname panel@csv/5/eager_optimized/time_reduction/max\endcsname{2.2}
\expandafter\def\csname panel@csv/5/initial_safe_two_step/success_positive_models\endcsname{0}
\expandafter\def\csname panel@csv/5/initial_safe_two_step/success_gain/min\endcsname{0.00}
\expandafter\def\csname panel@csv/5/initial_safe_two_step/success_gain/max\endcsname{0.00}
\expandafter\def\csname panel@csv/5/initial_safe_two_step/calls_saved/min\endcsname{0.254}
\expandafter\def\csname panel@csv/5/initial_safe_two_step/calls_saved/max\endcsname{0.490}
\expandafter\def\csname panel@csv/5/initial_safe_two_step/seconds_saved/min\endcsname{-0.109}
\expandafter\def\csname panel@csv/5/initial_safe_two_step/seconds_saved/max\endcsname{0.076}
\expandafter\def\csname panel@csv/5/initial_safe_two_step/queries_saved/min\endcsname{148.3}
\expandafter\def\csname panel@csv/5/initial_safe_two_step/queries_saved/max\endcsname{289.7}
\expandafter\def\csname panel@csv/5/initial_safe_two_step/time_reduction/min\endcsname{-1.2}
\expandafter\def\csname panel@csv/5/initial_safe_two_step/time_reduction/max\endcsname{0.7}
\expandafter\def\csname panel@csv/5/native_program_only/success_positive_models\endcsname{14}
\expandafter\def\csname panel@csv/5/native_program_only/success_gain/min\endcsname{2.54}
\expandafter\def\csname panel@csv/5/native_program_only/success_gain/max\endcsname{6.25}
\expandafter\def\csname panel@csv/5/native_program_only/calls_saved/min\endcsname{0.013}
\expandafter\def\csname panel@csv/5/native_program_only/calls_saved/max\endcsname{0.017}
\expandafter\def\csname panel@csv/5/native_program_only/seconds_saved/min\endcsname{-7.541}
\expandafter\def\csname panel@csv/5/native_program_only/seconds_saved/max\endcsname{-5.614}
\expandafter\def\csname panel@csv/5/native_program_only/queries_saved/min\endcsname{-2537.1}
\expandafter\def\csname panel@csv/5/native_program_only/queries_saved/max\endcsname{-1923.4}
\expandafter\def\csname panel@csv/5/native_program_only/time_reduction/min\endcsname{-208.4}
\expandafter\def\csname panel@csv/5/native_program_only/time_reduction/max\endcsname{-153.8}
\expandafter\def\csname panel@csv/5/partial_native_full_rows/success_positive_models\endcsname{0}
\expandafter\def\csname panel@csv/5/partial_native_full_rows/success_gain/min\endcsname{0.00}
\expandafter\def\csname panel@csv/5/partial_native_full_rows/success_gain/max\endcsname{0.00}
\expandafter\def\csname panel@csv/5/partial_native_full_rows/calls_saved/min\endcsname{0.000}
\expandafter\def\csname panel@csv/5/partial_native_full_rows/calls_saved/max\endcsname{0.000}
\expandafter\def\csname panel@csv/5/partial_native_full_rows/seconds_saved/min\endcsname{-0.042}
\expandafter\def\csname panel@csv/5/partial_native_full_rows/seconds_saved/max\endcsname{0.007}
\expandafter\def\csname panel@csv/5/partial_native_full_rows/queries_saved/min\endcsname{-0.2}
\expandafter\def\csname panel@csv/5/partial_native_full_rows/queries_saved/max\endcsname{-0.2}
\expandafter\def\csname panel@csv/5/partial_native_full_rows/time_reduction/min\endcsname{-0.4}
\expandafter\def\csname panel@csv/5/partial_native_full_rows/time_reduction/max\endcsname{0.1}
\expandafter\def\csname panel@csv/5/partial_native_no_tightening/success_positive_models\endcsname{0}
\expandafter\def\csname panel@csv/5/partial_native_no_tightening/success_gain/min\endcsname{0.00}
\expandafter\def\csname panel@csv/5/partial_native_no_tightening/success_gain/max\endcsname{0.00}
\expandafter\def\csname panel@csv/5/partial_native_no_tightening/calls_saved/min\endcsname{0.000}
\expandafter\def\csname panel@csv/5/partial_native_no_tightening/calls_saved/max\endcsname{0.000}
\expandafter\def\csname panel@csv/5/partial_native_no_tightening/seconds_saved/min\endcsname{-0.001}
\expandafter\def\csname panel@csv/5/partial_native_no_tightening/seconds_saved/max\endcsname{0.006}
\expandafter\def\csname panel@csv/5/partial_native_no_tightening/queries_saved/min\endcsname{0.0}
\expandafter\def\csname panel@csv/5/partial_native_no_tightening/queries_saved/max\endcsname{0.0}
\expandafter\def\csname panel@csv/5/partial_native_no_tightening/time_reduction/min\endcsname{-0.0}
\expandafter\def\csname panel@csv/5/partial_native_no_tightening/time_reduction/max\endcsname{0.1}
\expandafter\def\csname panel@json/3/eager_optimized/success_positive_models\endcsname{14}
\expandafter\def\csname panel@json/3/eager_optimized/success_gain/min\endcsname{0.14}
\expandafter\def\csname panel@json/3/eager_optimized/success_gain/max\endcsname{0.14}
\expandafter\def\csname panel@json/3/eager_optimized/calls_saved/min\endcsname{-0.002}
\expandafter\def\csname panel@json/3/eager_optimized/calls_saved/max\endcsname{-0.002}
\expandafter\def\csname panel@json/3/eager_optimized/seconds_saved/min\endcsname{5.309}
\expandafter\def\csname panel@json/3/eager_optimized/seconds_saved/max\endcsname{6.089}
\expandafter\def\csname panel@json/3/eager_optimized/queries_saved/min\endcsname{716.0}
\expandafter\def\csname panel@json/3/eager_optimized/queries_saved/max\endcsname{819.7}
\expandafter\def\csname panel@json/3/eager_optimized/time_reduction/min\endcsname{30.1}
\expandafter\def\csname panel@json/3/eager_optimized/time_reduction/max\endcsname{34.0}
\expandafter\def\csname panel@json/3/initial_safe_two_step/success_positive_models\endcsname{14}
\expandafter\def\csname panel@json/3/initial_safe_two_step/success_gain/min\endcsname{0.14}
\expandafter\def\csname panel@json/3/initial_safe_two_step/success_gain/max\endcsname{0.21}
\expandafter\def\csname panel@json/3/initial_safe_two_step/calls_saved/min\endcsname{0.009}
\expandafter\def\csname panel@json/3/initial_safe_two_step/calls_saved/max\endcsname{0.019}
\expandafter\def\csname panel@json/3/initial_safe_two_step/seconds_saved/min\endcsname{5.322}
\expandafter\def\csname panel@json/3/initial_safe_two_step/seconds_saved/max\endcsname{6.109}
\expandafter\def\csname panel@json/3/initial_safe_two_step/queries_saved/min\endcsname{713.1}
\expandafter\def\csname panel@json/3/initial_safe_two_step/queries_saved/max\endcsname{811.6}
\expandafter\def\csname panel@json/3/initial_safe_two_step/time_reduction/min\endcsname{30.2}
\expandafter\def\csname panel@json/3/initial_safe_two_step/time_reduction/max\endcsname{34.1}
\expandafter\def\csname panel@json/3/native_program_only/success_positive_models\endcsname{14}
\expandafter\def\csname panel@json/3/native_program_only/success_gain/min\endcsname{0.14}
\expandafter\def\csname panel@json/3/native_program_only/success_gain/max\endcsname{0.14}
\expandafter\def\csname panel@json/3/native_program_only/calls_saved/min\endcsname{-0.002}
\expandafter\def\csname panel@json/3/native_program_only/calls_saved/max\endcsname{-0.002}
\expandafter\def\csname panel@json/3/native_program_only/seconds_saved/min\endcsname{5.165}
\expandafter\def\csname panel@json/3/native_program_only/seconds_saved/max\endcsname{5.839}
\expandafter\def\csname panel@json/3/native_program_only/queries_saved/min\endcsname{530.8}
\expandafter\def\csname panel@json/3/native_program_only/queries_saved/max\endcsname{577.0}
\expandafter\def\csname panel@json/3/native_program_only/time_reduction/min\endcsname{29.5}
\expandafter\def\csname panel@json/3/native_program_only/time_reduction/max\endcsname{33.1}
\expandafter\def\csname panel@json/3/partial_native_full_rows/success_positive_models\endcsname{14}
\expandafter\def\csname panel@json/3/partial_native_full_rows/success_gain/min\endcsname{3.27}
\expandafter\def\csname panel@json/3/partial_native_full_rows/success_gain/max\endcsname{4.26}
\expandafter\def\csname panel@json/3/partial_native_full_rows/calls_saved/min\endcsname{-0.047}
\expandafter\def\csname panel@json/3/partial_native_full_rows/calls_saved/max\endcsname{-0.037}
\expandafter\def\csname panel@json/3/partial_native_full_rows/seconds_saved/min\endcsname{5.328}
\expandafter\def\csname panel@json/3/partial_native_full_rows/seconds_saved/max\endcsname{6.051}
\expandafter\def\csname panel@json/3/partial_native_full_rows/queries_saved/min\endcsname{718.0}
\expandafter\def\csname panel@json/3/partial_native_full_rows/queries_saved/max\endcsname{860.7}
\expandafter\def\csname panel@json/3/partial_native_full_rows/time_reduction/min\endcsname{30.2}
\expandafter\def\csname panel@json/3/partial_native_full_rows/time_reduction/max\endcsname{33.9}
\expandafter\def\csname panel@json/3/partial_native_no_tightening/success_positive_models\endcsname{0}
\expandafter\def\csname panel@json/3/partial_native_no_tightening/success_gain/min\endcsname{0.00}
\expandafter\def\csname panel@json/3/partial_native_no_tightening/success_gain/max\endcsname{0.00}
\expandafter\def\csname panel@json/3/partial_native_no_tightening/calls_saved/min\endcsname{0.000}
\expandafter\def\csname panel@json/3/partial_native_no_tightening/calls_saved/max\endcsname{0.000}
\expandafter\def\csname panel@json/3/partial_native_no_tightening/seconds_saved/min\endcsname{0.038}
\expandafter\def\csname panel@json/3/partial_native_no_tightening/seconds_saved/max\endcsname{0.061}
\expandafter\def\csname panel@json/3/partial_native_no_tightening/queries_saved/min\endcsname{0.9}
\expandafter\def\csname panel@json/3/partial_native_no_tightening/queries_saved/max\endcsname{8.6}
\expandafter\def\csname panel@json/3/partial_native_no_tightening/time_reduction/min\endcsname{0.3}
\expandafter\def\csname panel@json/3/partial_native_no_tightening/time_reduction/max\endcsname{0.5}
\expandafter\def\csname panel@json/5/eager_optimized/success_positive_models\endcsname{14}
\expandafter\def\csname panel@json/5/eager_optimized/success_gain/min\endcsname{0.14}
\expandafter\def\csname panel@json/5/eager_optimized/success_gain/max\endcsname{0.14}
\expandafter\def\csname panel@json/5/eager_optimized/calls_saved/min\endcsname{-0.002}
\expandafter\def\csname panel@json/5/eager_optimized/calls_saved/max\endcsname{-0.001}
\expandafter\def\csname panel@json/5/eager_optimized/seconds_saved/min\endcsname{5.236}
\expandafter\def\csname panel@json/5/eager_optimized/seconds_saved/max\endcsname{6.066}
\expandafter\def\csname panel@json/5/eager_optimized/queries_saved/min\endcsname{752.1}
\expandafter\def\csname panel@json/5/eager_optimized/queries_saved/max\endcsname{828.8}
\expandafter\def\csname panel@json/5/eager_optimized/time_reduction/min\endcsname{29.7}
\expandafter\def\csname panel@json/5/eager_optimized/time_reduction/max\endcsname{33.9}
\expandafter\def\csname panel@json/5/initial_safe_two_step/success_positive_models\endcsname{14}
\expandafter\def\csname panel@json/5/initial_safe_two_step/success_gain/min\endcsname{0.14}
\expandafter\def\csname panel@json/5/initial_safe_two_step/success_gain/max\endcsname{0.21}
\expandafter\def\csname panel@json/5/initial_safe_two_step/calls_saved/min\endcsname{0.010}
\expandafter\def\csname panel@json/5/initial_safe_two_step/calls_saved/max\endcsname{0.019}
\expandafter\def\csname panel@json/5/initial_safe_two_step/seconds_saved/min\endcsname{5.222}
\expandafter\def\csname panel@json/5/initial_safe_two_step/seconds_saved/max\endcsname{6.062}
\expandafter\def\csname panel@json/5/initial_safe_two_step/queries_saved/min\endcsname{750.8}
\expandafter\def\csname panel@json/5/initial_safe_two_step/queries_saved/max\endcsname{824.1}
\expandafter\def\csname panel@json/5/initial_safe_two_step/time_reduction/min\endcsname{29.6}
\expandafter\def\csname panel@json/5/initial_safe_two_step/time_reduction/max\endcsname{33.9}
\expandafter\def\csname panel@json/5/native_program_only/success_positive_models\endcsname{14}
\expandafter\def\csname panel@json/5/native_program_only/success_gain/min\endcsname{0.14}
\expandafter\def\csname panel@json/5/native_program_only/success_gain/max\endcsname{0.14}
\expandafter\def\csname panel@json/5/native_program_only/calls_saved/min\endcsname{-0.002}
\expandafter\def\csname panel@json/5/native_program_only/calls_saved/max\endcsname{-0.001}
\expandafter\def\csname panel@json/5/native_program_only/seconds_saved/min\endcsname{5.065}
\expandafter\def\csname panel@json/5/native_program_only/seconds_saved/max\endcsname{5.795}
\expandafter\def\csname panel@json/5/native_program_only/queries_saved/min\endcsname{551.0}
\expandafter\def\csname panel@json/5/native_program_only/queries_saved/max\endcsname{592.3}
\expandafter\def\csname panel@json/5/native_program_only/time_reduction/min\endcsname{29.0}
\expandafter\def\csname panel@json/5/native_program_only/time_reduction/max\endcsname{32.9}
\expandafter\def\csname panel@json/5/partial_native_full_rows/success_positive_models\endcsname{14}
\expandafter\def\csname panel@json/5/partial_native_full_rows/success_gain/min\endcsname{2.70}
\expandafter\def\csname panel@json/5/partial_native_full_rows/success_gain/max\endcsname{3.69}
\expandafter\def\csname panel@json/5/partial_native_full_rows/calls_saved/min\endcsname{-0.040}
\expandafter\def\csname panel@json/5/partial_native_full_rows/calls_saved/max\endcsname{-0.031}
\expandafter\def\csname panel@json/5/partial_native_full_rows/seconds_saved/min\endcsname{4.399}
\expandafter\def\csname panel@json/5/partial_native_full_rows/seconds_saved/max\endcsname{5.079}
\expandafter\def\csname panel@json/5/partial_native_full_rows/queries_saved/min\endcsname{478.5}
\expandafter\def\csname panel@json/5/partial_native_full_rows/queries_saved/max\endcsname{558.0}
\expandafter\def\csname panel@json/5/partial_native_full_rows/time_reduction/min\endcsname{26.2}
\expandafter\def\csname panel@json/5/partial_native_full_rows/time_reduction/max\endcsname{30.0}
\expandafter\def\csname panel@json/5/partial_native_no_tightening/success_positive_models\endcsname{0}
\expandafter\def\csname panel@json/5/partial_native_no_tightening/success_gain/min\endcsname{0.00}
\expandafter\def\csname panel@json/5/partial_native_no_tightening/success_gain/max\endcsname{0.00}
\expandafter\def\csname panel@json/5/partial_native_no_tightening/calls_saved/min\endcsname{0.000}
\expandafter\def\csname panel@json/5/partial_native_no_tightening/calls_saved/max\endcsname{0.000}
\expandafter\def\csname panel@json/5/partial_native_no_tightening/seconds_saved/min\endcsname{-0.043}
\expandafter\def\csname panel@json/5/partial_native_no_tightening/seconds_saved/max\endcsname{0.020}
\expandafter\def\csname panel@json/5/partial_native_no_tightening/queries_saved/min\endcsname{0.0}
\expandafter\def\csname panel@json/5/partial_native_no_tightening/queries_saved/max\endcsname{4.3}
\expandafter\def\csname panel@json/5/partial_native_no_tightening/time_reduction/min\endcsname{-0.4}
\expandafter\def\csname panel@json/5/partial_native_no_tightening/time_reduction/max\endcsname{0.2}
\expandafter\def\csname panel@csv/all/eager_optimized/time_reduction/min\endcsname{0.2}
\expandafter\def\csname panel@csv/all/eager_optimized/time_reduction/max\endcsname{4.5}
\expandafter\def\csname panel@json/all/eager_optimized/time_reduction/min\endcsname{29.7}
\expandafter\def\csname panel@json/all/eager_optimized/time_reduction/max\endcsname{34.0}

\newcommand{\SVal}[1]{\csname supplement@#1\endcsname}
\expandafter\def\csname supplement@e1/api_calls\endcsname{1442}
\expandafter\def\csname supplement@e1/common\endcsname{188}
\expandafter\def\csname supplement@e1/csv/3/cape/calls\endcsname{2.196}
\expandafter\def\csname supplement@e1/csv/3/cape/completion\endcsname{85.16}
\expandafter\def\csname supplement@e1/csv/3/retry/calls\endcsname{2.113}
\expandafter\def\csname supplement@e1/csv/3/retry/completion\endcsname{69.78}
\expandafter\def\csname supplement@e1/csv/3/trial/calls\endcsname{2.113}
\expandafter\def\csname supplement@e1/csv/3/trial/completion\endcsname{69.97}
\expandafter\def\csname supplement@e1/csv/5/cape/calls\endcsname{2.348}
\expandafter\def\csname supplement@e1/csv/5/cape/completion\endcsname{98.44}
\expandafter\def\csname supplement@e1/csv/5/retry/calls\endcsname{2.608}
\expandafter\def\csname supplement@e1/csv/5/retry/completion\endcsname{82.32}
\expandafter\def\csname supplement@e1/csv/5/trial/calls\endcsname{2.605}
\expandafter\def\csname supplement@e1/csv/5/trial/completion\endcsname{83.01}
\expandafter\def\csname supplement@e1/json/3/cape/calls\endcsname{1.476}
\expandafter\def\csname supplement@e1/json/3/cape/completion\endcsname{94.74}
\expandafter\def\csname supplement@e1/json/3/retry/calls\endcsname{1.957}
\expandafter\def\csname supplement@e1/json/3/retry/completion\endcsname{60.09}
\expandafter\def\csname supplement@e1/json/3/trial/calls\endcsname{1.957}
\expandafter\def\csname supplement@e1/json/3/trial/completion\endcsname{60.09}
\expandafter\def\csname supplement@e1/json/5/cape/calls\endcsname{1.475}
\expandafter\def\csname supplement@e1/json/5/cape/completion\endcsname{94.74}
\expandafter\def\csname supplement@e1/json/5/retry/calls\endcsname{2.682}
\expandafter\def\csname supplement@e1/json/5/retry/completion\endcsname{72.80}
\expandafter\def\csname supplement@e1/json/5/trial/calls\endcsname{2.731}
\expandafter\def\csname supplement@e1/json/5/trial/completion\endcsname{67.90}
\expandafter\def\csname supplement@e1/optimal\endcsname{251}
\expandafter\def\csname supplement@e1/retry/gain_max\endcsname{34.66}
\expandafter\def\csname supplement@e1/retry/gain_min\endcsname{15.38}
\expandafter\def\csname supplement@e1/tokens\endcsname{2,159,674}
\expandafter\def\csname supplement@e1/trial/gain_max\endcsname{34.66}
\expandafter\def\csname supplement@e1/trial/gain_min\endcsname{15.19}
\expandafter\def\csname supplement@e1/valid\endcsname{252}
\expandafter\def\csname supplement@e2/anchor/eager_optimized/gain\endcsname{0.00}
\expandafter\def\csname supplement@e2/anchor/eager_optimized/queries_saved\endcsname{78.0}
\expandafter\def\csname supplement@e2/anchor/lazy_andor_selective/gain\endcsname{0.00}
\expandafter\def\csname supplement@e2/anchor/lazy_andor_selective/queries_saved\endcsname{0.0}
\expandafter\def\csname supplement@e2/certificates\endcsname{150}
\expandafter\def\csname supplement@e2/dates-128/eager_optimized/gain\endcsname{21.88}
\expandafter\def\csname supplement@e2/dates-128/eager_optimized/queries_saved\endcsname{1085.2}
\expandafter\def\csname supplement@e2/dates-128/lazy_andor_selective/gain\endcsname{14.06}
\expandafter\def\csname supplement@e2/dates-128/lazy_andor_selective/queries_saved\endcsname{172.0}
\expandafter\def\csname supplement@e2/dates-256/eager_optimized/gain\endcsname{5.57}
\expandafter\def\csname supplement@e2/dates-256/eager_optimized/queries_saved\endcsname{627.8}
\expandafter\def\csname supplement@e2/dates-256/lazy_andor_selective/gain\endcsname{4.00}
\expandafter\def\csname supplement@e2/dates-256/lazy_andor_selective/queries_saved\endcsname{81.4}
\expandafter\def\csname supplement@e2/horizon-8/eager_optimized/gain\endcsname{0.00}
\expandafter\def\csname supplement@e2/horizon-8/eager_optimized/queries_saved\endcsname{78.0}
\expandafter\def\csname supplement@e2/horizon-8/lazy_andor_selective/gain\endcsname{0.00}
\expandafter\def\csname supplement@e2/horizon-8/lazy_andor_selective/queries_saved\endcsname{0.0}
\expandafter\def\csname supplement@e2/merged-feedback/eager_optimized/gain\endcsname{0.00}
\expandafter\def\csname supplement@e2/merged-feedback/eager_optimized/queries_saved\endcsname{78.0}
\expandafter\def\csname supplement@e2/merged-feedback/lazy_andor_selective/gain\endcsname{0.00}
\expandafter\def\csname supplement@e2/merged-feedback/lazy_andor_selective/queries_saved\endcsname{0.0}
\expandafter\def\csname supplement@e2/resolved\endcsname{36}

\newcommand{\AVal}[1]{\csname allowance@#1\endcsname}
\expandafter\def\csname allowance@anchor/10/eager_optimized/completion\endcsname{85.16}
\expandafter\def\csname allowance@anchor/10/eager_optimized/gain\endcsname{10.16}
\expandafter\def\csname allowance@anchor/10/initial\endcsname{64.84}
\expandafter\def\csname allowance@anchor/10/lazy_andor_selective/completion\endcsname{92.97}
\expandafter\def\csname allowance@anchor/10/lazy_andor_selective/gain\endcsname{2.34}
\expandafter\def\csname allowance@anchor/10/partial_native/completion\endcsname{95.31}
\expandafter\def\csname allowance@anchor/15/eager_optimized/completion\endcsname{88.28}
\expandafter\def\csname allowance@anchor/15/eager_optimized/gain\endcsname{11.72}
\expandafter\def\csname allowance@anchor/15/initial\endcsname{64.84}
\expandafter\def\csname allowance@anchor/15/lazy_andor_selective/completion\endcsname{95.31}
\expandafter\def\csname allowance@anchor/15/lazy_andor_selective/gain\endcsname{4.69}
\expandafter\def\csname allowance@anchor/15/partial_native/completion\endcsname{100.00}
\expandafter\def\csname allowance@anchor/20/eager_optimized/completion\endcsname{95.31}
\expandafter\def\csname allowance@anchor/20/eager_optimized/gain\endcsname{4.69}
\expandafter\def\csname allowance@anchor/20/initial\endcsname{64.84}
\expandafter\def\csname allowance@anchor/20/lazy_andor_selective/completion\endcsname{94.53}
\expandafter\def\csname allowance@anchor/20/lazy_andor_selective/gain\endcsname{5.47}
\expandafter\def\csname allowance@anchor/20/partial_native/completion\endcsname{100.00}
\expandafter\def\csname allowance@anchor/30/eager_optimized/completion\endcsname{100.00}
\expandafter\def\csname allowance@anchor/30/eager_optimized/gain\endcsname{0.00}
\expandafter\def\csname allowance@anchor/30/initial\endcsname{64.84}
\expandafter\def\csname allowance@anchor/30/lazy_andor_selective/completion\endcsname{98.44}
\expandafter\def\csname allowance@anchor/30/lazy_andor_selective/gain\endcsname{1.56}
\expandafter\def\csname allowance@anchor/30/partial_native/completion\endcsname{100.00}
\expandafter\def\csname allowance@anchor/5/eager_optimized/completion\endcsname{83.20}
\expandafter\def\csname allowance@anchor/5/eager_optimized/gain\endcsname{-0.39}
\expandafter\def\csname allowance@anchor/5/initial\endcsname{64.84}
\expandafter\def\csname allowance@anchor/5/lazy_andor_selective/completion\endcsname{86.72}
\expandafter\def\csname allowance@anchor/5/lazy_andor_selective/gain\endcsname{-3.91}
\expandafter\def\csname allowance@anchor/5/partial_native/completion\endcsname{82.81}
\expandafter\def\csname allowance@attempts\endcsname{1107}
\expandafter\def\csname allowance@certificates\endcsname{366}
\expandafter\def\csname allowance@dates-128/10/eager_optimized/completion\endcsname{65.62}
\expandafter\def\csname allowance@dates-128/10/eager_optimized/gain\endcsname{0.00}
\expandafter\def\csname allowance@dates-128/10/initial\endcsname{53.91}
\expandafter\def\csname allowance@dates-128/10/lazy_andor_selective/completion\endcsname{64.45}
\expandafter\def\csname allowance@dates-128/10/lazy_andor_selective/gain\endcsname{1.17}
\expandafter\def\csname allowance@dates-128/10/partial_native/completion\endcsname{65.62}
\expandafter\def\csname allowance@dates-128/120/eager_optimized/completion\endcsname{95.31}
\expandafter\def\csname allowance@dates-128/120/eager_optimized/gain\endcsname{2.34}
\expandafter\def\csname allowance@dates-128/120/initial\endcsname{53.91}
\expandafter\def\csname allowance@dates-128/120/lazy_andor_selective/completion\endcsname{92.97}
\expandafter\def\csname allowance@dates-128/120/lazy_andor_selective/gain\endcsname{4.69}
\expandafter\def\csname allowance@dates-128/120/partial_native/completion\endcsname{97.66}
\expandafter\def\csname allowance@dates-128/15/eager_optimized/completion\endcsname{66.41}
\expandafter\def\csname allowance@dates-128/15/eager_optimized/gain\endcsname{8.59}
\expandafter\def\csname allowance@dates-128/15/initial\endcsname{53.91}
\expandafter\def\csname allowance@dates-128/15/lazy_andor_selective/completion\endcsname{74.22}
\expandafter\def\csname allowance@dates-128/15/lazy_andor_selective/gain\endcsname{0.78}
\expandafter\def\csname allowance@dates-128/15/partial_native/completion\endcsname{75.00}
\expandafter\def\csname allowance@dates-128/20/eager_optimized/completion\endcsname{66.41}
\expandafter\def\csname allowance@dates-128/20/eager_optimized/gain\endcsname{15.23}
\expandafter\def\csname allowance@dates-128/20/initial\endcsname{53.91}
\expandafter\def\csname allowance@dates-128/20/lazy_andor_selective/completion\endcsname{74.22}
\expandafter\def\csname allowance@dates-128/20/lazy_andor_selective/gain\endcsname{7.42}
\expandafter\def\csname allowance@dates-128/20/partial_native/completion\endcsname{81.64}
\expandafter\def\csname allowance@dates-128/30/eager_optimized/completion\endcsname{66.41}
\expandafter\def\csname allowance@dates-128/30/eager_optimized/gain\endcsname{21.09}
\expandafter\def\csname allowance@dates-128/30/initial\endcsname{53.91}
\expandafter\def\csname allowance@dates-128/30/lazy_andor_selective/completion\endcsname{74.22}
\expandafter\def\csname allowance@dates-128/30/lazy_andor_selective/gain\endcsname{13.28}
\expandafter\def\csname allowance@dates-128/30/partial_native/completion\endcsname{87.50}
\expandafter\def\csname allowance@dates-128/45/eager_optimized/completion\endcsname{75.10}
\expandafter\def\csname allowance@dates-128/45/eager_optimized/gain\endcsname{18.65}
\expandafter\def\csname allowance@dates-128/45/initial\endcsname{53.91}
\expandafter\def\csname allowance@dates-128/45/lazy_andor_selective/completion\endcsname{74.22}
\expandafter\def\csname allowance@dates-128/45/lazy_andor_selective/gain\endcsname{19.53}
\expandafter\def\csname allowance@dates-128/45/partial_native/completion\endcsname{93.75}
\expandafter\def\csname allowance@dates-128/5/eager_optimized/completion\endcsname{56.64}
\expandafter\def\csname allowance@dates-128/5/eager_optimized/gain\endcsname{2.73}
\expandafter\def\csname allowance@dates-128/5/initial\endcsname{53.91}
\expandafter\def\csname allowance@dates-128/5/lazy_andor_selective/completion\endcsname{58.20}
\expandafter\def\csname allowance@dates-128/5/lazy_andor_selective/gain\endcsname{1.17}
\expandafter\def\csname allowance@dates-128/5/partial_native/completion\endcsname{59.38}
\expandafter\def\csname allowance@dates-128/60/eager_optimized/completion\endcsname{82.62}
\expandafter\def\csname allowance@dates-128/60/eager_optimized/gain\endcsname{14.26}
\expandafter\def\csname allowance@dates-128/60/initial\endcsname{53.91}
\expandafter\def\csname allowance@dates-128/60/lazy_andor_selective/completion\endcsname{84.57}
\expandafter\def\csname allowance@dates-128/60/lazy_andor_selective/gain\endcsname{12.30}
\expandafter\def\csname allowance@dates-128/60/partial_native/completion\endcsname{96.88}
\expandafter\def\csname allowance@dates-128/90/eager_optimized/completion\endcsname{95.31}
\expandafter\def\csname allowance@dates-128/90/eager_optimized/gain\endcsname{2.34}
\expandafter\def\csname allowance@dates-128/90/initial\endcsname{53.91}
\expandafter\def\csname allowance@dates-128/90/lazy_andor_selective/completion\endcsname{92.97}
\expandafter\def\csname allowance@dates-128/90/lazy_andor_selective/gain\endcsname{4.69}
\expandafter\def\csname allowance@dates-128/90/partial_native/completion\endcsname{97.66}
\expandafter\def\csname allowance@dates-256/10/eager_optimized/completion\endcsname{50.78}
\expandafter\def\csname allowance@dates-256/10/eager_optimized/gain\endcsname{0.39}
\expandafter\def\csname allowance@dates-256/10/initial\endcsname{48.44}
\expandafter\def\csname allowance@dates-256/10/lazy_andor_selective/completion\endcsname{50.98}
\expandafter\def\csname allowance@dates-256/10/lazy_andor_selective/gain\endcsname{0.20}
\expandafter\def\csname allowance@dates-256/10/partial_native/completion\endcsname{51.17}
\expandafter\def\csname allowance@dates-256/120/eager_optimized/completion\endcsname{57.03}
\expandafter\def\csname allowance@dates-256/120/eager_optimized/gain\endcsname{12.50}
\expandafter\def\csname allowance@dates-256/120/initial\endcsname{48.44}
\expandafter\def\csname allowance@dates-256/120/lazy_andor_selective/completion\endcsname{63.67}
\expandafter\def\csname allowance@dates-256/120/lazy_andor_selective/gain\endcsname{5.86}
\expandafter\def\csname allowance@dates-256/120/partial_native/completion\endcsname{69.53}
\expandafter\def\csname allowance@dates-256/20/eager_optimized/completion\endcsname{56.74}
\expandafter\def\csname allowance@dates-256/20/eager_optimized/gain\endcsname{-0.10}
\expandafter\def\csname allowance@dates-256/20/initial\endcsname{48.44}
\expandafter\def\csname allowance@dates-256/20/lazy_andor_selective/completion\endcsname{54.69}
\expandafter\def\csname allowance@dates-256/20/lazy_andor_selective/gain\endcsname{1.95}
\expandafter\def\csname allowance@dates-256/20/partial_native/completion\endcsname{56.64}
\expandafter\def\csname allowance@dates-256/240/eager_optimized/completion\endcsname{67.77}
\expandafter\def\csname allowance@dates-256/240/eager_optimized/gain\endcsname{5.27}
\expandafter\def\csname allowance@dates-256/240/initial\endcsname{48.44}
\expandafter\def\csname allowance@dates-256/240/lazy_andor_selective/completion\endcsname{63.67}
\expandafter\def\csname allowance@dates-256/240/lazy_andor_selective/gain\endcsname{9.38}
\expandafter\def\csname allowance@dates-256/240/partial_native/completion\endcsname{73.05}
\expandafter\def\csname allowance@dates-256/30/eager_optimized/completion\endcsname{57.03}
\expandafter\def\csname allowance@dates-256/30/eager_optimized/gain\endcsname{5.18}
\expandafter\def\csname allowance@dates-256/30/initial\endcsname{48.44}
\expandafter\def\csname allowance@dates-256/30/lazy_andor_selective/completion\endcsname{58.59}
\expandafter\def\csname allowance@dates-256/30/lazy_andor_selective/gain\endcsname{3.61}
\expandafter\def\csname allowance@dates-256/30/partial_native/completion\endcsname{62.21}
\expandafter\def\csname allowance@dates-256/60/eager_optimized/completion\endcsname{57.03}
\expandafter\def\csname allowance@dates-256/60/eager_optimized/gain\endcsname{9.96}
\expandafter\def\csname allowance@dates-256/60/initial\endcsname{48.44}
\expandafter\def\csname allowance@dates-256/60/lazy_andor_selective/completion\endcsname{63.67}
\expandafter\def\csname allowance@dates-256/60/lazy_andor_selective/gain\endcsname{3.32}
\expandafter\def\csname allowance@dates-256/60/partial_native/completion\endcsname{66.99}
\expandafter\def\csname allowance@disturbed\endcsname{105}
\expandafter\def\csname allowance@disturbed/eager_optimized\endcsname{43}
\expandafter\def\csname allowance@disturbed/lazy_andor_selective\endcsname{25}
\expandafter\def\csname allowance@disturbed/partial_native\endcsname{37}
\expandafter\def\csname allowance@disturbed_remaining\endcsname{0}
\expandafter\def\csname allowance@groups\endcsname{9}
\expandafter\def\csname allowance@records\endcsname{960}
\expandafter\def\csname allowance@round/1/disturbed\endcsname{32}
\expandafter\def\csname allowance@round/1/measurements\endcsname{105}
\expandafter\def\csname allowance@round/2/disturbed\endcsname{10}
\expandafter\def\csname allowance@round/2/measurements\endcsname{32}
\expandafter\def\csname allowance@round/3/disturbed\endcsname{0}
\expandafter\def\csname allowance@round/3/measurements\endcsname{10}
\expandafter\def\csname allowance@unresolved_certificates\endcsname{0}

\definecolor{cvprblue}{rgb}{0.21,0.49,0.74}
\usepackage[breaklinks,colorlinks,allcolors=cvprblue]{hyperref}
\newcommand{\PaperTitle}{CAPE: Coordinating Calibration and Search for Tool-Use Policies}
\title{\PaperTitle}
\newcommand{\AuthorNames}{Jiahao Wang, Xiaoxuan Chen, Aijun Yang, Xiaohua Wang}
\newcommand{\AuthorAffiliations}{State Key Laboratory of Electrical Insulation and Power Equipment, Xi’an Jiaotong University}
\author{%
  \AuthorNames\\
  {\normalsize\shortstack[c]{\AuthorAffiliations}}\\
}

\hypersetup{pdftitle={\PaperTitle},pdfauthor={\AuthorNames}}
\begin{document}
\maketitle
\begin{figure*}[t]
\centering
\includegraphics[width=\linewidth]{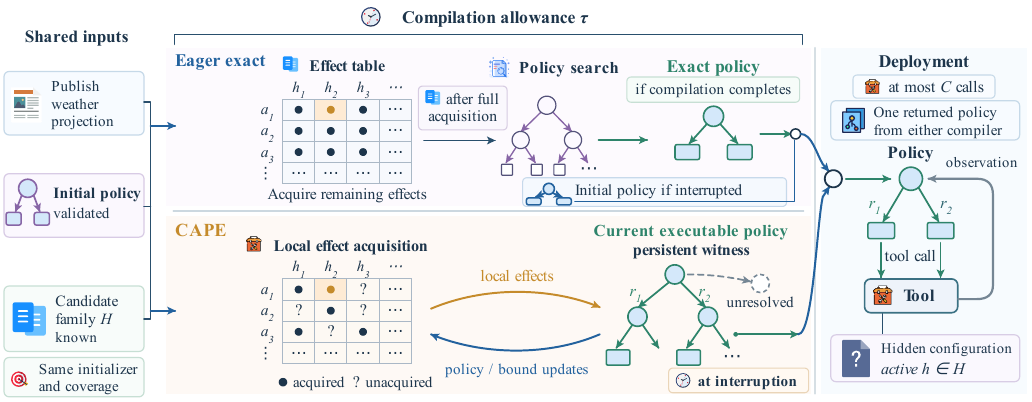}
\caption{\textbf{Calibration and search share a compilation allowance.}
Eager exact completes acquisition before search; CAPE updates executable
policies during acquisition. Both retain the validated initial policy;
deployment has a separate $C$-call limit.}
\label{fig:overview}
\end{figure*}

\begin{abstract}
A tool call can fail even when the operation behind it is correct,
because the receiving tool parses delimiters, dates or record layouts
under a configuration the agent cannot inspect.
Compiling a tool-use policy for such a tool requires both calibration,
which evaluates candidate calls on candidate configurations, and search
over the evaluated calls.
The two share one compilation budget: calibrating everything first
postpones any improved policy until the end, while re-solving every
partially known model delays the next useful query.
A call becomes usable as soon as its effects are known on the
configurations that reach it.
We present CAPE, a training-free compiler built on this observation.
It maintains an executable policy at all times, directs calibration to
the calls that can improve this policy, and tightens bounds only where
doing so ends calibration early.
Returned policies preserve every configuration the initial policy
completes, and CAPE certifies optimality when its executable and
optimistic values meet.
On CSV and JSON workflows built from 64 requests, CAPE reaches the
independently computed optimum in \JVal{candidate/optimal_records} of
\JVal{candidate/native_records} compilations while taking
\JVal{overall/eager_optimized/time_reduction}\% less compilation time
than eager exact search.
On larger CSV families it returns policies with higher completion at
similar compilation time, and it raises completion by
\SVal{e1/retry/gain_min}--\SVal{e1/retry/gain_max} percentage points over
an LLM that revises its serialization from error feedback.
\end{abstract}

\section{Introduction}
\label{sec:intro}

Tool-using agents act by sending calls to external systems, and a call
can fail even when the operation behind it is correct.
Consider an agent that uploads a weather table to a data-ingestion tool.
A parser that expects semicolons rejects a comma-delimited file, and one
that expects day-first dates reads \texttt{05/04} as April~5 rather than
May~4.
The parser's options are set where the tool is deployed, and the agent
sees only the parsed result or an error.
Completing the task under every such configuration therefore takes a
\emph{policy}, a plan that fixes which call to send after each response
within the few calls that one execution is allowed.
When the candidate configurations can be reproduced in a test
environment, this policy can be compiled before deployment.

Such a compiler combines two activities that prior work has studied
separately: acting on tool feedback, as in ReAct and
CodeAct~\cite{react,codeact}, and searching when evaluations are
expensive, as in LazySP, LRA* and GLS~\cite{lazysp,lra,gls}.
For tool policies, \emph{calibration} evaluates a candidate call on a
candidate configuration and records its effect, and search assembles the
recorded effects into the policy that completes the most configurations
with the fewest calls.

Calibration and search draw on one compilation budget, and neither
measures progress by itself.
Evaluating every call on every configuration before searching spends
calibration on configurations the policy never reaches
(Fig.~\ref{fig:overview}, top), while re-solving the partially known
model after each query delays the next useful effect.
Tool policies tighten this coupling: an unevaluated call has an unknown
outcome and also an unknown effect on the policy's branches, because
configurations that an optimistic model keeps apart may return the same
response and must then share one continuation.
The compiler has to decide what to learn next, what to compute next, and
what it can return now.

A policy needs effects only on its own execution paths.
In Fig.~\ref{fig:example}, a rejected upload rules out one of four
configurations, and a single probe then distinguishes the remaining
three and directs each to a compatible program.
The probe needs effects on those three alone: a call is ready to use
once its effects are known on the configurations that reach it.
Known local effects thus yield an executable policy and a lower bound on
the achievable value, optimistic assumptions about the unknown effects
yield an upper bound, and each new effect can improve the policy in
place or narrow the gap.

We present CAPE, a training-free compiler that allocates calibration and
search by their contribution to this executable policy.
For each set of configurations consistent with the responses so far,
CAPE keeps the best policy that the acquired effects support and an
optimistic bound on what the unevaluated calls can still add.
Calibration goes to the calls where the two differ: first where a
configuration completed by the initial policy is not yet covered, and
then by potential gain per missing effect.
While a call is being evaluated, CAPE tightens its bound with a two-call
lookahead and stops as soon as the call is ready to use or can no longer
beat the current policy.
Unlike lazy search that follows the optimistic solution alone, this
acquisition frontier also follows the executable policy.
Every returned policy completes all configurations that the initial
policy completes, and when the two values meet at the root CAPE
certifies optimality over the full action catalog.

On CSV and JSON workflows built from 64 requests, CAPE reaches the
independently computed optimum in \JVal{candidate/optimal_records} of
\JVal{candidate/native_records} compilations and takes
\JVal{overall/eager_optimized/time_reduction}\% less compilation time
than eager exact search.
When the CSV family grows to 128 configurations, CAPE completes
\SVal{e2/dates-128/lazy_andor_selective/gain} and
\SVal{e2/dates-128/eager_optimized/gain} percentage points more
configurations than a matched AND/OR scheduler and eager exact search at
similar compilation time.
In execution, the compiled policies raise completion by
\SVal{e1/retry/gain_min}--\SVal{e1/retry/gain_max} percentage points
over an LLM that revises its serialization from error feedback under the
same call limit.

We contribute an analysis of how calibration and search compete for the
compilation budget (Sec.~\ref{sec:analysis}), the CAPE compiler with
coverage and optimality guarantees (Sec.~\ref{sec:cape}), and matched
scheduling, scaling and online-retry studies that locate the gains
(Sec.~\ref{sec:experiments}).

\section{Related Work}
\label{sec:related}

\paragraph{Tool use and agent execution.}
ReAct and CodeAct choose each action from execution
feedback~\cite{react,codeact}, ReWOO plans before observing tool
evidence~\cite{rewoo}, LLMCompiler executes a dependency graph of calls
in parallel~\cite{llmcompiler}, and ADaPT decomposes a task further when
its executor fails~\cite{adapt_llm}.
Tool documentation and exploratory interaction improve what an agent
knows about a tool before it acts~\cite{tooldocs,play2prompt,toolobserver}.
CAPE addresses the step that follows these decisions: the operation is
chosen, and what remains unknown is the representation that the tool
accepts.

\paragraph{Search with expensive evaluations.}
LazySP evaluates edges only on optimistic shortest paths~\cite{lazysp}, LRA*
and GLS relate lookahead to the numbers of evaluations and graph
operations~\cite{lra,gls}, and PDDLStream allocates effort between
search and sampling~\cite{pddlstream}.
CAPE inherits two principles from this line: optimistic candidates
direct evaluation, and evaluation and search are charged to one budget.
What changes is the object being evaluated.
An edge evaluation returns a cost on a fixed graph, whereas an effect
returns a response that determines how the configurations split and
thereby the shape of the policy below it.
CAPE's frontier therefore follows the executable policy in addition to
the optimistic candidate.

\paragraph{Partial policies and guarantees.}
LAO* expands a partial solution graph and propagates value updates to
its ancestors~\cite{laostar}, and BRTDP directs search with lower and
upper values~\cite{brtdp}.
Decision-region determination and decision-theoretic troubleshooting
select the observations and tests that a decision
needs~\cite{drd,troubleshooting}.
SPIBB bounds the loss of expected return relative to a baseline
policy~\cite{spibb}, and shielding enforces guarantees under partial
observability~\cite{partial_shield,safe_pomdp}.
CAPE differs in three respects.
Its lower values are attained by stored policies built from acquired
effects, its upper values relax the unknown effects themselves, and its
requirement is configuration-wise: every configuration completed by the
initial policy stays completed, whereas baseline-relative methods
preserve an expectation.

\section{Analysis}
\label{sec:analysis}

\subsection{Problem formulation}
\label{sec:formulation}

\begin{figure}[t]
\centering
\includegraphics[width=.94\linewidth]{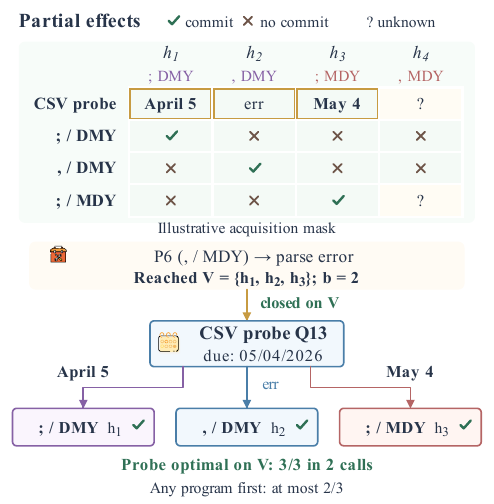}
\caption{\textbf{A locally closed probe is optimal on the reached scope.}
With two calls left, Q13 completes all three configurations in $V$;
any program-first continuation completes at most two.
The table shows the three programs that commit on $V$; dates abbreviate
the full response; DMY/MDY denote date order.
The acquisition mask is illustrative.}
\label{fig:example}
\end{figure}

We formalize the ingestion example of Sec.~\ref{sec:intro}.
An agent holds a table and a proposed operation, here a projection of
the Seattle weather records, and has to publish the operation's result
through a tool whose parsing configuration it cannot inspect.
Figure~\ref{fig:example} uses four configurations: comma or semicolon
delimiters, crossed with day-first or month-first dates.

\paragraph{Actions and effects.}
The candidate calls, which we call \emph{actions}, are of two kinds.
A \emph{program} submits the table in one serialization together with
the operation.
A \emph{probe} submits a small diagnostic fixture and returns how the
tool parsed it, without publishing anything.
The CSV probe Q13 in Fig.~\ref{fig:example} submits a semicolon-separated
fixture whose first row contains the date \texttt{05/04/2026}:
semicolon parsers report April~5 or May~4 according to their date
convention, and comma parsers return a parse error (the \supp\ gives the
full fixture).

Let $H$ be the finite family of candidate configurations and
$A=P\cup Q$ the catalog of programs $P$ and probes $Q$.
For a fixed request, calling action $a$ under configuration $h$ has the
deterministic \emph{effect} $f(a,h)=(o(a,h),g(a,h))$: a response and a
commit bit.
The bit is one when a program publishes the requested result, which ends
the execution; probes never commit.
A call that does not commit leaves the task state unchanged.
A \emph{calibration query} evaluates one pair $(a,h)$ in a test
environment that reproduces $h$, with the same response and commit
semantics as deployment.
Acquired effects accumulate in a table $\calD\subseteq A\times H$.

\paragraph{Policies and objective.}
A policy chooses each action from the responses received so far and
uses at most $C$ calls on any execution.
Because effects are deterministic and failed calls change nothing, a
policy's knowledge after a sequence of responses is the set
$V\subseteq H$ of configurations consistent with them, together with the
number $b$ of remaining calls.
Write $S(\pi;V,b)$ for the configurations that $\pi$ completes from
$V$ with $b$ calls remaining, and $M(\pi;V,b)$ for its total calls over
$V$, including calls on executions that fail.
The initial policy $\pi_0$ defines the required set $R_0=S(\pi_0;H,C)$,
and the compiler maximizes
\begin{equation}
 \begin{aligned}
 J(\pi)&=\bigl(|S(\pi;H,C)|,-M(\pi;H,C)\bigr),\\
 &\hspace{12pt}R_0\subseteq S(\pi;H,C),
 \end{aligned}
 \label{eq:objective}
\end{equation}
in lexicographic order with equal weight on every configuration: it
first completes as many configurations as possible and then uses as few
calls as possible, without losing a configuration that $\pi_0$
completes.
The compilation allowance $\tau$ limits the work of constructing $\pi$,
and $C$ limits the calls of each execution.

\paragraph{A running example.}
In Fig.~\ref{fig:example}, the comma/month-first program P6 commits on
$h_4$ and returns the same parse error on the other three
configurations, which leaves $V=\{h_1,h_2,h_3\}$ with two calls.
Probe Q13 separates $V$ into three responses, each followed by its
matching program, so every configuration in $V$ is completed within the
two remaining calls.
A program-first continuation completes at most two
(Sec.~\ref{sec:local}), so Q13 is optimal at this state, and its branch
is executable
although the effects of Q13 and of the semicolon/month-first program on
$h_4$ are unknown.

\subsection{The acquisition--computation tradeoff}
\label{sec:tradeoff}

Calibration supplies effects, and search determines whether those
effects support a better policy.
If $c_i$ is the measured cost of the $i$-th calibration query,
compilation time decomposes as
\begin{equation}
 T_{\mathrm{comp}}
 = T_{\mathrm{fixed}}+T_{\mathrm{search}}+\sum_i c_i,
 \label{eq:cost}
\end{equation}
where $T_{\mathrm{fixed}}$ covers input construction, final
serialization and cleanup, and $T_{\mathrm{search}}$ covers policy
construction, bound computation and the updates that follow new effects.
The two variable terms pull against each other: search that selects
queries more carefully shortens the sum while lengthening
$T_{\mathrm{search}}$, as in lazy graph search~\cite{lra,gls}.

The exchange is costly for tool policies.
An earlier solver of ours solved the acquired part of the model exactly
before each further query.
On a bird-strike projection it spent approximately 22 seconds replanning
after about 600 calibration queries beyond initial validation, and on
its development workload it reached 93.60\% CSV completion at $C=5$
where eager exact search reached 98.44\%.

Bound computation shows the other side.
We profile one two-call bound operator applied in two ways: only to the
action currently being acquired (Selected), or at every eligible state
(Always).
At $C=3$ on CSV, tightening everywhere raises bound computation from
\MFVal{profile/csv/3/cape_selective/tight} to
\MFVal{profile/csv/3/cape_always/tight} seconds and lowers calibration
from \MFVal{profile/csv/3/cape_selective/calibration} to
\MFVal{profile/csv/3/cape_always/calibration} seconds, a net saving.
At $C=5$ it raises bound computation from
\MFVal{profile/csv/5/cape_selective/tight} to
\MFVal{profile/csv/5/cape_always/tight} seconds while calibration barely
changes (\supp, Sec.~E.1).
A refinement pays for itself only when the queries it avoids cost more
than the search it adds.

\subsection{Implications for policy construction}
\label{sec:implications}

Three consequences shape the compiler.
A branch needs effects only on the configurations that reach it, so it
becomes executable before the row of its action is complete.
A policy built from acquired effects follows the same paths whatever is
acquired later, so it can be stored and improved in place.
Optimistic candidates, which give every unknown configuration its own
hypothetical response, tell the compiler where to look, not what to
return.
Further calibration or computation is therefore useful when it improves
the stored policy or shows that an unresolved action cannot beat it.

\section{CAPE}
\label{sec:cape}

\begin{figure*}[t]
\centering
\includegraphics[width=\linewidth]{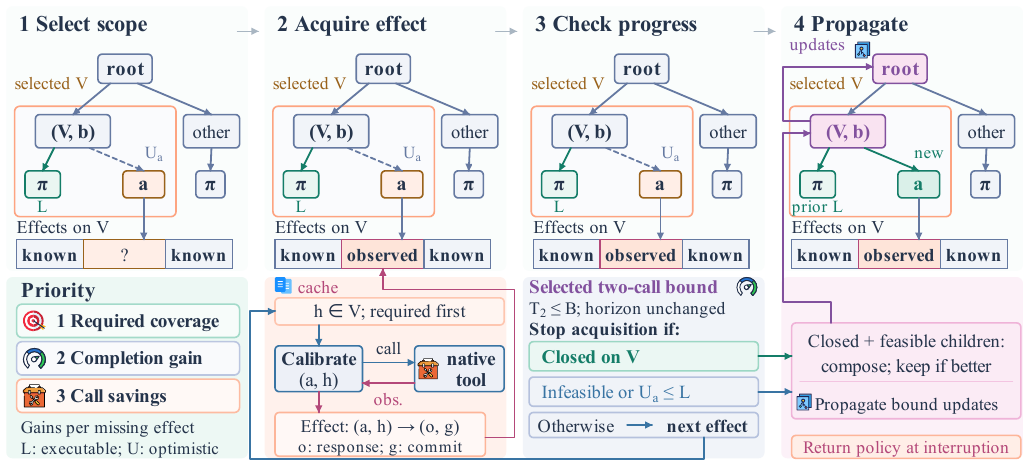}
\caption{\textbf{A local observation updates a persistent policy graph.}
Acquisition stops on closure, infeasibility or domination.
Composition additionally requires feasible child witnesses;
witness and bound changes propagate to the affected parents.}
\label{fig:policy-update}
\end{figure*}

CAPE alternates calibration on a chosen local scope with updates to a
persistent policy graph (Fig.~\ref{fig:policy-update}; pseudocode is
in the \supp).
A \emph{state} $(V,b)$ of this graph is a response-consistent
configuration set with its remaining calls.
Each state carries its obligation $R_0\cap V$, a lower value $L$
attained by a stored executable policy, and an upper value $U$ that no
policy can exceed.
Values are pairs of completed configurations and negated calls, compared
lexicographically as in Eq.~\eqref{eq:objective}.
We instantiate each step on the state $(V,2)$ of Fig.~\ref{fig:example}.

\subsection{Local policies and incremental composition}
\label{sec:local}

An action can enter an executable policy at $(V,b)$ once its effects are
acquired for every configuration in $V$; we call it
\emph{closed on $V$}.
A closed action completes the configurations $G_a(V)$ on which it
commits and partitions the remaining ones by their observed responses
into disjoint cells $V_o$, each a child state $(V_o,b-1)$.
If the children hold policies with success sets $S_o$ and call totals
$M_o$, composing them under $a$ gives
\begin{equation}
 S_a=G_a(V)\cup\bigcup_o S_o,\qquad
 M_a=|V|+\sum_o M_o.
 \label{eq:backup}
\end{equation}
Every configuration in $V$ pays for the first call, and the
configurations of each cell pay for the calls of its child.
The composition is feasible when $S_a$ covers $R_0\cap V$.
Stopping has value $(0,0)$ and is feasible when the obligation is empty.

On $(V,2)$, Q13 is closed and commits nowhere.
Its response cells $\{h_1\}$, $\{h_2\}$ and $\{h_3\}$ are each completed
by the matching program, so Eq.~\eqref{eq:backup} gives $S_a=V$ and
$M_a=3+1+1+1=6$.
Starting with the semicolon/day-first program instead completes $h_1$
and leaves the failure cell $\{h_2,h_3\}$, where one call completes one
configuration: two completions with $M_a=3+2=5$.
The probe composition has the higher value, $(3,-6)$ against $(2,-5)$.

CAPE stores the best feasible composition of each state as its
\emph{witness}: a policy with its success set and call total.
Here the witness is the program-first policy until Q13 closes and the
probe policy afterwards.
When the witness of a child improves, Eq.~\eqref{eq:backup} updates the
parents that reach this child through a closed action, and no other
action is re-examined.
Validating $\pi_0$ seeds the root witness.
Because a witness uses acquired effects only, later acquisitions leave
its executions and its value unchanged.

\subsection{Optimistic bounds and acquisition frontiers}
\label{sec:frontier}

Upper values tell the compiler which unfinished actions can still
improve a witness.
Let $E_a(V)=\{h\in V:(a,h)\notin\calD\}$ be the configurations on which
action $a$ is unknown.
A program $p$ can at best commit on all of them, so its optimistic
success set is $\overline G_p(V)=G_p(V)\cup E_p(V)$.
The configurations that some program might complete, and the most that a
single program might complete, are
\begin{equation}
 \overline G(V)=\bigcup_{p\in P}\overline G_p(V),\qquad
 \overline d(V)=\max_{p\in P}|\overline G_p(V)|.
 \label{eq:support}
\end{equation}
For nonempty $\overline G(V)$ and $b\geq2$, a cheap upper value is
\begin{equation}
 B(V,b)=\left(|\overline G(V)|,\,
 -\bigl[|V|+|\overline G(V)|-\overline d(V)\bigr]\right).
 \label{eq:cheap}
\end{equation}
No policy completes more than $|\overline G(V)|$ configurations, and one
that completes all of them calls once on every configuration of $V$ and
at least once more on the $|\overline G|-\overline d$ completions that
its first program cannot cover.
A required configuration outside $\overline G(V)$ makes the state
infeasible.
In the running example with the program effects on $V$ acquired, three
programs commit on one configuration each, so $\overline G(V)=V$,
$\overline d(V)=1$ and $B(V,2)=(3,-5)$.

The same optimism bounds an individual unfinished action.
An unfinished program is assumed to commit on its unknown
configurations, and an unfinished probe to return a distinct response on
each of them, the finest partition that any completion of its row could
produce; acquired failures keep their actual response cells.
Let $\calF_a^+$ denote the resulting optimistic cells, with child upper
values $U_W=(k_W,c_W)$ for $W\in\calF_a^+$.
The upper value of the action is
\begin{equation}
 U_a(V,b)=
 \left(|\overline G_a(V)|+\sum_W k_W,\,
 -|V|+\sum_W c_W\right).
 \label{eq:action}
\end{equation}
We write $U_a(V,b;B')$ when a bound function $B'$ supplies the child
values.
An infeasible required child excludes the action.
Closed actions use the current upper values of their child states, and
unfinished actions use the cheap bound $B$ unless they are being
tightened (Sec.~\ref{sec:tightening}).
A state takes its best feasible action bound, includes Stop when
allowed, and is capped by Eq.~\eqref{eq:cheap}.
Upper values stay valid as effects arrive, so CAPE keeps the minimum of
the old and the new one.

Calibration is chosen from an \emph{acquisition frontier}: the states
that the current executable policy reaches from the root, together with
the states along one best optimistic alternative, whose closed actions
are expanded lazily.
On this frontier, unfinished actions are ranked in three tiers
(Fig.~\ref{fig:policy-update}).
\emph{Required coverage} comes first: states that lack a feasible
witness for their obligation, ranked by required configurations per
missing effect.
\emph{Completion gain} is next: actions whose upper value completes more
configurations than the witness, ranked by this gain per missing effect.
\emph{Call savings} come last: actions that can only save calls, ranked
by the saving per missing effect.
Within the selected action, required configurations are queried first.
Before Q13 closes on $(V,2)$, its upper value permits one more completion
than the program-first witness, which places it in the second tier.

\subsection{Selective tightening and computation allocation}
\label{sec:tightening}

The cheap bound ignores whether the remaining calls can exploit the
observed response cells.
For a state with two calls left, the optimistic problem is small enough
to solve exactly: CAPE enumerates every first action and, after each of
its optimistic cells, the best single program,
\begin{equation}
 T_2(V)=\max_{a\in A} U_a(V,2;B),
 \label{eq:t2}
\end{equation}
including Stop when feasible.
$T_2(V)$ is the optimum of the relaxed two-call problem and satisfies
$T_2(V)\leq B(V,2)$.
In the running example, a program-first continuation completes at most
two configurations, and a probe needs a second call on each
configuration it completes, so $T_2(V)=(3,-6)$.
This equals the witness value whatever the other probe effects are, so
no further acquisition can improve this state.

Where to apply this operator is the allocation question of
Sec.~\ref{sec:tradeoff}, since tightening every state can cost more
than the calibration it avoids.
CAPE applies $T_2$ to the two-call continuations of the action it is
currently acquiring, where a tighter value changes the next decision.
After each new effect, acquisition of this action stops as soon as it is
closed, infeasible, or bounded by the available witness.
All other unfinished actions keep the cheap bound.

% Table 1 is placed here so that it lands at the top of the first
% page of the Experiments section.
\begin{table*}[t]
\centering\small\setlength{\tabcolsep}{4pt}
\caption{\textbf{CAPE combines high completion with lower compilation cost.}
DeepSeek mechanism study under the 30-second allowance.
Success is completion (\%), counting invalid proposals as failures;
calls are per configuration; time is compilation time in seconds.
Bold marks unique column optima and CAPE's tied optima; the supplement
adds both ablations and all intervals.}
\label{tab:main}
\begin{tabular*}{\linewidth}{@{\extracolsep{\fill}}llrrrrrr@{}}
\toprule
& & \multicolumn{3}{c}{CSV} & \multicolumn{3}{c}{JSON} \\
\cmidrule(lr){3-5}\cmidrule(l){6-8}
$C$ & Method & Success & Calls & Time & Success & Calls & Time \\
\midrule
3 & \textbf{CAPE} & \textbf{85.16} & 2.196 & 10.774 & \textbf{94.74} & 1.476 & \textbf{11.919} \\
 & Eager exact & 85.16 & 2.196 & 11.173 & 94.60 & 1.474 & 18.021 \\
 & Program-only & 76.76 & \textbf{2.022} & \textbf{3.631} & 94.60 & 1.474 & 17.792 \\
 & Protected two-step & 83.98 & 2.474 & 11.121 & 94.60 & 1.493 & 18.008 \\
 & Generic lazy & 83.74 & 2.198 & 17.089 & 92.90 & 1.495 & 16.200 \\
 & Program-first & 85.16 & 2.196 & 11.402 & 94.60 & 1.474 & 17.878 \\
\midrule
5 & \textbf{CAPE} & \textbf{98.44} & \textbf{2.348} & 11.201 & \textbf{94.74} & 1.475 & \textbf{11.903} \\
 & Eager exact & 98.44 & 2.348 & 11.417 & 94.60 & 1.474 & 18.079 \\
 & Program-only & 92.19 & 2.365 & \textbf{3.624} & 94.60 & 1.474 & 17.744 \\
 & Protected two-step & 98.44 & 2.838 & 11.192 & 94.60 & 1.493 & 18.005 \\
 & Generic lazy & 84.18 & 2.534 & 13.700 & 92.90 & 1.546 & 16.008 \\
 & Program-first & 98.44 & 2.348 & 11.572 & 94.60 & 1.474 & 17.919 \\
\bottomrule
\end{tabular*}

\end{table*}

\subsection{Interruption, coverage and certification}
\label{sec:guarantees}

\begin{proposition}
\label{prop:guarantee}
Under the effect model of Sec.~\ref{sec:formulation}, every accepted root
witness preserves $R_0$ within $C$ calls, and accepted objective values
are nondecreasing.
Witness values persist under further consistent acquisitions.
Every maintained root upper value bounds the full-catalog optimum.
Equality of the root lower and upper values certifies optimality.
\end{proposition}

The proof (\supp) rests on two facts.
A composition of closed actions executes exactly as computed, because
its cells are the observed ones, and optimism is a relaxation, because
completing the missing effects can only remove assumed commits or merge
assumed cells.
The proposition holds for every acquisition order; the order determines
how much calibration and search precede a given policy or certificate
(Sec.~\ref{sec:mechanisms}).

On interruption, CAPE returns the latest root witness; if the initial
validation itself is interrupted, it returns $\pi_0$ without a
certificate.
Once the effect table is complete, a compact exact solver for the same
objective uses the remaining allowance.

\section{Experiments}
\label{sec:experiments}

\subsection{Setup}
\label{sec:setup}

\paragraph{Requests and tools.}
Eight Vega source files~\cite{vega_datasets} provide 64 requests, one
per source for each of eight operations such as projection, grouped
count and filtering, with at most 256 records per source.
Two tools ingest the data: a CSV tool built on the DuckDB reader with 64
configurations, and a JSON tool built on pandas with
88~\cite{duckcsv,pandas_json}.
The configurations vary delimiters, quoting, missing-value markers,
numeric and date encodings, and JSON orientation.
Programs are the distinct serializations of the table under these
encodings, and probes are fixed diagnostic fixtures.
A program call commits only when its result equals the reference output,
which is computed in advance from the requested operation on the typed
data, independently of the proposal; otherwise the call is rolled back.

\paragraph{Proposals and studies.}
An LLM proposes one operation per request.
The mechanism study uses DeepSeek-V4-Pro, for which 63 of 64 proposals
are schema-valid and match the typed reference; the remaining one counts
as a failure for every method.
It compares the methods below on both tools at $C\in\{3,5\}$ with two
fresh-process repetitions.
The matched, scaling and execution studies below use the same method
without modification; the matched study runs on the distinct inputs of a
fourteen-model proposal panel.

% Placed here so that Figure 4 lands at the top of the next page.
\begin{figure*}[t]
\centering
\includegraphics[width=\linewidth]{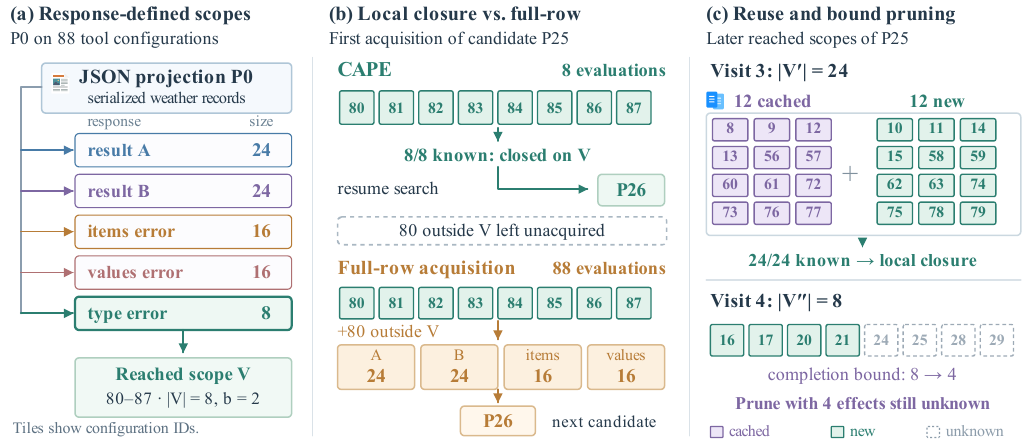}
\caption{\textbf{A recorded calibration case} (JSON weather projection, $C=3$).
(a) P0 responses partition 88 configurations.
(b) CAPE closes the eight-configuration scope; full-row acquisition also
evaluates the 80 outside it.
(c) Later visits reuse 12 effects to close a 24-configuration scope and
prune an eight-configuration scope after four evaluations.}
\label{fig:footprint}
\end{figure*}

\paragraph{Comparisons.}
All compilers start from the same initial policy, which tries the
programs in catalog order, and inherit its required configurations.
\emph{Eager exact} acquires all remaining effects and then runs a
compact constrained search.
\emph{Generic lazy} search validates optimistic models in batches of 32.
\emph{Program-first} acquires program rows before any probe effect.
\emph{Protected two-step} uses rolling diagnostic lookahead and keeps
the initial policy whenever an alternative loses coverage or objective
value.
\emph{Program-only} removes the probes from the catalog.
The matched study crosses two acquisition schedules, CAPE's frontier and
an AND/OR search that follows the best optimistic solution path, with
three bound modes: none, selected-action and always, where Always
applies the same $T_2$ operator at every eligible bound request.
Both schedules share witnesses, validation, caches, bound operators and
the exact fallback, so they differ only in the order of acquisition.

\paragraph{Measurement and verification.}
Each timed compilation starts in a fresh process with empty caches and a
soft 30-second allowance.
\emph{Compilation time} is the full lifecycle of one compilation: setup,
validation of the initial policy, acquisition, search, serialization and
cleanup.
\emph{Completion} is the percentage of configurations completed and
counts invalid proposals as failures; calls and compilation costs are
averaged over schema-valid proposals.
Sources have equal weight, and paired 95\% intervals come from 5,000
bootstrap resamples of the eight sources.
All results are verified independently of the compilers, by dynamic
programming on the complete effect table and by fresh executions.
A runtime certificate is the compiler's own statement that its root
values met.
The \supp\ gives every protocol.

\subsection{Policy quality under the compilation allowance}
\label{sec:quality}

CAPE returns optimal or near-optimal policies.
It attains the full-catalog optimum in
\JVal{candidate/optimal_records}/\JVal{candidate/native_records}
compilations and certifies \JVal{candidate/certified_records} of them
at runtime, and every audited policy preserves the configurations
completed by the initial policy.
In the two remaining compilations, on the JSON weather projection at
$C=3$, CAPE reaches optimal completion with more calls.

CAPE matches eager exact search in policy quality at lower compilation
cost (Table~\ref{tab:main}).
Completion and calls are equal on CSV, and CAPE completes
\JVal{json/3/eager_optimized/success_gain} points more on JSON.
Mean compilation time falls to
\JVal{overall/partial_native/lifecycle_seconds} seconds from
\JVal{overall/eager_optimized/lifecycle_seconds} for eager exact and
\JVal{overall/lazy_quotient/lifecycle_seconds} for generic lazy search.
Against eager exact, the paired savings are
\JVal{csv/3/eager_optimized/seconds_saved} seconds on CSV at $C=3$ and
\JVal{json/3/eager_optimized/seconds_saved}/%
\JVal{json/5/eager_optimized/seconds_saved} seconds on JSON at $C=3/5$,
while the CSV interval at $C=5$,
$[\JVal{csv/5/eager_optimized/seconds_interval}]$, includes zero.

Probes matter most under a tight call limit.
At $C=3$, policies with probes raise CSV completion from
\JVal{csv/3/native_program_only/success_rate}\% for program-only search
to \JVal{csv/3/partial_native/success_rate}\%.
Against protected two-step, CAPE gains
\JVal{csv/3/initial_safe_two_step/success_gain} percentage points and
saves \JVal{csv/3/initial_safe_two_step/calls_saved} calls per
configuration.
At $C=5$ both complete \JVal{csv/5/partial_native/success_rate}\% of the
CSV configurations, and CAPE uses \JVal{csv/5/partial_native/calls}
calls where protected two-step uses
\JVal{csv/5/initial_safe_two_step/calls}.

\subsection{Acquisition and computation tradeoffs}
\label{sec:mechanisms}

\paragraph{Local effect scopes.}
Calibrating only the reached scope is the largest single saving.
The full-row ablation acquires the remaining cells of an action after
its reached scope closes.
On JSON it issues
\JVal{json/3/partial_native_full_rows/queries_saved} and
\JVal{json/5/partial_native_full_rows/queries_saved} more calibration
queries at $C=3/5$ and completes
\JVal{json/3/partial_native_full_rows/success_gain} and
\JVal{json/5/partial_native_full_rows/success_gain} fewer percentage
points, because queries outside the reached scope consume the allowance
before another continuation becomes executable.
Figure~\ref{fig:footprint} traces one candidate program, P25, in the
recorded JSON weather projection.
The responses of the initial program P0 leave an eight-configuration
scope, on which CAPE closes P25 with eight queries where full-row
acquisition issues 88, and later visits reuse cached effects and stop
early on tightened bounds.

\paragraph{Acquisition order.}
Following the executable policy as well as the optimistic one saves
compilation time.
With selective bounds in both schedules, CAPE is
\MFVal{csv/3/cape_selective/lazy_andor_selective/seconds_saved} seconds
faster than the matched AND/OR scheduler on CSV at $C=3$ and
\MFVal{json/3/cape_selective/lazy_andor_selective/seconds_saved}/%
\MFVal{json/5/cape_selective/lazy_andor_selective/seconds_saved}
seconds faster on JSON at $C=3/5$, and all three paired intervals
exclude zero.
On CSV at $C=5$ the two schedules return equal execution quality and the
time interval contains zero; the JSON savings hold under all three bound
modes.
Without tightening, the CSV differences shrink and CAPE is slower at
$C=5$, so the benefit of the order depends on the bound information
available to it.

\paragraph{When tighter computation pays.}
Tightening pays when it removes enough calibration.
At CSV $C=3$, always tightening avoids 433.6 further queries and is
0.741 seconds faster than selective tightening at equal execution
quality.
At $C=5$ it avoids few queries, and selective tightening is
\MFVal{csv/5/cape_selective/cape_always/seconds_saved} seconds faster.
Both paired intervals exclude zero, and the bound costs of
Sec.~\ref{sec:tradeoff} account for the reversal.
On JSON the two modes differ little.

\paragraph{Sources of improvement.}
Policy quality depends on what the action catalog can attain and on
what the compiler finds within $\tau$.
On CSV, probes raise the attainable completion on a subset of requests,
as in Fig.~\ref{fig:example}.
On JSON the optima with and without probes coincide for every audited
request (\supp, Sec.~D.1), so the timed comparison measures how
efficiently each compiler recovers the same attainable completion.
At $C=3$, CAPE reaches \JVal{json/3/partial_native/success_rate}\%
completion in \JVal{json/3/partial_native/lifecycle_seconds} seconds,
whereas program-only search reaches
\JVal{json/3/native_program_only/success_rate}\% in
\JVal{json/3/native_program_only/lifecycle_seconds} seconds
(Table~\ref{tab:main}), a difference that the matched scheduler and the
full-row ablation attribute to acquisition order and local closure.

\subsection{Structural and scaling behavior}
\label{sec:structure}

The compilers separate in quality once the family outgrows the
allowance.
We take the projection request of each source and enlarge the CSV family
from 64 to 128 and 256 configurations by adding date encodings, keeping
the original effects (Table~\ref{tab:scaling}).
At 64 configurations all three methods complete every configuration.
At 128, CAPE completes \SVal{e2/dates-128/eager_optimized/gain}
percentage points more configurations than eager exact and
\SVal{e2/dates-128/lazy_andor_selective/gain} more than matched AND/OR
while all three use approximately 27 seconds.
At 256, the gains are \SVal{e2/dates-256/eager_optimized/gain} and
\SVal{e2/dates-256/lazy_andor_selective/gain} points.
All four paired completion intervals exclude zero, and from 5 to 120
seconds of allowance at 128 configurations CAPE is never below either
control.
While the model is unresolved, the compiler that keeps improving an
executable policy returns the better one.

\begin{table}[t]
\centering\small\setlength{\tabcolsep}{3pt}
\caption{\textbf{CAPE improves completion on larger families.}
Eight CSV projection requests at $C=5$ under the 30-second allowance.
Columns follow Table~\ref{tab:main}.}
\label{tab:scaling}
\begin{tabular}{rlrrr}
\toprule
$|H|$ & Method & Success $\uparrow$ & Calls $\downarrow$ & Time (s) \\
\midrule
64 & Eager exact & 100.00 & 2.875 & 14.17 \\
 & Matched AND/OR & 100.00 & 2.875 & 14.87 \\
 & CAPE & \textbf{100.00} & 2.875 & 15.14 \\
\midrule
128 & Eager exact & 66.41 & 3.309 & 26.64 \\
 & Matched AND/OR & 74.22 & 3.379 & 26.74 \\
 & CAPE & \textbf{88.28} & 3.309 & 26.66 \\
\midrule
256 & Eager exact & 57.03 & 3.506 & 26.85 \\
 & Matched AND/OR & 58.59 & 3.619 & 26.90 \\
 & CAPE & \textbf{62.60} & 3.479 & 26.90 \\
\bottomrule
\end{tabular}

\end{table}

\begin{table}[t]
\centering\small\setlength{\tabcolsep}{3pt}
\caption{\textbf{Compiled policies complete more configurations.}
Completion (\%) over 256 conditions; calls average over valid
proposals, including failed paths.}
\label{tab:execution-retry}
\begin{tabular}{llrr}
\toprule
Interface/$C$ & Method & Success $\uparrow$ & Calls $\downarrow$ \\
\midrule
CSV/3 & Catalog retry & 69.97 & 2.113 \\
 & LLM retry & 69.78 & 2.113 \\
 & CAPE & \textbf{85.16} & 2.196 \\
\midrule
CSV/5 & Catalog retry & 83.01 & 2.605 \\
 & LLM retry & 82.32 & 2.608 \\
 & CAPE & \textbf{98.44} & 2.348 \\
\midrule
JSON/3 & Catalog retry & 60.09 & 1.957 \\
 & LLM retry & 60.09 & 1.957 \\
 & CAPE & \textbf{94.74} & 1.476 \\
\midrule
JSON/5 & Catalog retry & 67.90 & 2.731 \\
 & LLM retry & 72.80 & 2.682 \\
 & CAPE & \textbf{94.74} & 1.475 \\
\bottomrule
\end{tabular}

\end{table}

The \supp\ adds this sweep (Sec.~G.4), $C=8$, merged probe feedback and
the proposal panel.

\subsection{Execution against feedback-driven retry}
\label{sec:practical}

Table~\ref{tab:execution-retry} compares the compiled policies in
execution with two online controllers on all 64 requests, both tools and
$C=3/5$.
Catalog retry executes the initial policy.
LLM retry uses DeepSeek-V4-Pro with the operation fixed: it executes the
first program of the initial policy and, after each failure, selects an
untried serialization from the request, the schema, a data example and
the error history.

CAPE completes \SVal{e1/retry/gain_min} to \SVal{e1/retry/gain_max}
percentage points more configurations than LLM retry, with all four
paired 95\% intervals above zero, and it preserves every configuration
that the initial policy completes.
On JSON at $C=5$ it also needs fewer calls,
\SVal{e1/json/5/cape/calls} per configuration against
\SVal{e1/json/5/retry/calls}.
Compilation selects probes and response-dependent continuations before
deployment, with every candidate configuration in view.
Counting compilation, CAPE's total time is 6.2--10.2 seconds above that
of LLM retry over 1, 8 and 32 executions at a fixed endpoint
(\supp, Sec.~G.2).

\section{Conclusion}
\label{sec:conclusion}

Compiling a tool-use policy is a question of where a limited budget
goes.
CAPE makes an executable policy the common object of calibration and
search: effects are acquired where they can extend or improve this
policy, bounds are tightened where they can end acquisition, and the
policy that exists at any moment can be returned with its coverage
guarantee and, when its value meets the optimistic bound, an optimality
certificate.
In our experiments, CAPE reaches the optimum in
\JVal{candidate/optimal_records} of \JVal{candidate/native_records}
compilations with \JVal{overall/eager_optimized/time_reduction}\% less
compilation time than eager exact search, returns better policies than
eager and matched lazy search once the configuration family outgrows the
allowance, and completes more configurations in execution than
feedback-driven LLM retry.

\clearpage
{\small
\setlength{\bibsep}{0pt}

}

\onecolumn
\raggedbottom
\appendix
\pdfbookmark[0]{Supplementary Material}{supplement}
\section*{Supplementary Material}
\section{Correctness of local policies and partial-effect bounds}
\label{app:proofs}

Fix the request, configuration family $H$, action catalog $A=P\cup Q$,
initial policy $\pi_0$, execution budget $C$, and protected set
$R_0=S(\pi_0;H,C)$.
The effect of an action is deterministic and consists of a response and a
commit bit. A successful program terminates; a probe does not commit;
a failed call leaves the state relevant to future calls unchanged.
Calibration and execution use the same semantics, and acquired effects
are immutable.
Thus a surviving configuration set $V$ and remaining budget $b$ suffice
to describe a policy state.

\subsection{Executable composition}

An action is closed on $V$ when all of its effects on $V$ are acquired.
Its immediate successes $G_a(V)$ and failed-response cells $V_o$ partition
$V$.
Let every child witness satisfy its obligation $R_0\cap V_o$ in at most
$b-1$ calls.
Every required configuration either succeeds immediately or belongs to
one child that completes it.
The parent therefore preserves $R_0\cap V$ within $b$ calls.
Disjoint cells give
\[
 S_a=G_a(V)\cup\bigcup_o S_o,\qquad
 M_a=|V|+\sum_o M_o.
\]
This counts calls on failed paths as well as successful paths.
At zero remaining calls, Stop is feasible precisely when the obligation
is empty. It has success set $\varnothing$ and call mass zero.

\begin{lemma}[Persistence of an executable witness]
A witness built entirely from effects in $\calD$ has the same execution
paths, success set and call mass in every consistent extension
$\calD'\supseteq\calD$.
\end{lemma}
\begin{proof}
For any reached configuration, the first action's acquired effect is the
same in both tables.
Its commit bit either terminates both executions or its response selects
the same child.
Induction on the remaining call budget gives identical continuations.
Effects acquired outside the witness's reached scopes cannot change
these paths.
\end{proof}

The policy graph is acyclic in remaining calls.
Closed-action backups compose feasible witnesses even when a child's
best value has not yet been resolved.
Only strict improvements replace stored witnesses.
Persistence then ensures that accepted root values never decrease.
The obligation remains the original $R_0$; other successful
configurations may change between policies with a better lexicographic
value.

\subsection{Admissible upper values}

For each program, $\overline G_p(V)=G_p(V)\cup E_p(V)$ contains all
configurations it could complete in a consistent full model.
Their union $\overline G(V)$ contains the support of every policy.
If a required configuration is outside this union, the state is infeasible.

Let $\overline d(V)=\max_p|\overline G_p(V)|$.
A policy completing all $|\overline G(V)|$ possible successes must start
on all of $V$ and needs a second call on at least
$|\overline G(V)|-\overline d(V)$ successful configurations.
Its call mass is therefore at least
$|V|+|\overline G(V)|-\overline d(V)$.
A policy completing fewer configurations is already lexicographically
below the support bound, independent of its call cost.
This proves admissibility of
\[
 B(V,b)=\left(|\overline G(V)|,
 -[|V|+|\overline G(V)|-\overline d(V)]\right),\qquad b\geq2,
\]
for nonempty support.
Empty support with no obligation permits Stop.
At one remaining call, enumerating optimistic terminal programs that
cover the obligation gives the exact relaxed terminal value.

\begin{lemma}[Optimistic effect relaxation]
Treating unknown program effects as immediate successes and unknown probe
effects as distinct response cells gives an upper bound on every
consistent completion of the effect table.
\end{lemma}
\begin{proof}
Take a feasible policy in any consistent complete model.
Whenever an unknown program effect is encountered, replacing its
continuation by immediate success preserves its success and can reduce
its calls.
Acquired failures retain their actual cells.
For unknown probe effects, assigning a separate response to each
configuration refines the partition produced by any real completion.
The refined observer can choose the same continuation for every cell
that would be merged, reproducing the original policy.
The simulation preserves every required success.
Optimizing over this enlarged set of choices and composing valid child
bounds therefore gives an upper value.
\end{proof}

In the relaxed two-call problem, the first action is followed by a terminal
program in each failed-response cell, or Stop where feasible.
Enumerating the first action and best terminal choices computes the
relaxation exactly.
Its value $T_2(V)$ is admissible and no greater than the support bound
$B(V,2)$.
Substituting it for a child upper value preserves admissibility at the
ancestors.

\subsection{Propagation, reductions and caches}

Every old upper value remains valid for the smaller set of consistent
models after an acquisition.
Taking the minimum of old and new valid bounds is therefore sound.
A previously pruned action cannot become better than the persistent
witness that already dominated its upper value.
If lower and upper values meet at a state, every consistent later
model has that same optimal value.

The cache for a scope-dependent calculation uses a generation derived
from acquisitions within that scope.
If $r_h$ counts acquisitions on configuration $h$, then
$\sum_{h\in V}r_h$ is unchanged exactly when no new effect in $V$ arrives,
because each count is nondecreasing.
This generation is sufficient for a fixed scope and obligation.
Program support and closed terminal choices depend only on program
effects, so probe acquisitions preserve those cache entries.

Acquired actions with the same success set and the same failed-response
partition can share a representative: continuations are attached to the
representative's observed label for each cell.
Success and call mass are preserved.
A predictable noncommitting action has a single failure cell and no
successes. Deleting it and using its continuation preserves completion
while saving a call on each reached configuration.
The unchanged-state failure semantics justify this reduction.

\subsection{Root guarantee and termination}

Initial validation establishes $R_0$ and the executable initial policy.
The composition argument and witness persistence establish feasible,
nondecreasing root improvements.
The relaxation argument and bound propagation establish admissible
root upper values.
If the root values agree, the executable value is at least every
full-catalog policy value and is therefore optimal.

Acquisition adds at most $|A||H|$ distinct effects.
When the table is complete, the same finite-horizon recurrence becomes
exact, and the compact complete-information solver processes it with
the remaining allowance.
An interrupted run returns its best available executable witness.
Interruption before initial validation finishes returns $\pi_0$ with
certification unset.
The finite policy-state space can be exponential in $|H|$.
These guarantees concern feasibility and value bounds; the measured
cost differences arise from how much acquisition and graph computation
each schedule performs.

\section{Action catalogs, observations and implementation}
\label{app:implementation}

\subsection{Programs and diagnostic probes}

CSV configurations combine six binary reader choices: comma/semicolon
delimiter, single/double quote, backslash/doubled-quote escaping,
\texttt{NULL}/\texttt{NA} missing marker, comma/point decimal separator,
and month-first/day-first dates.
Their product contains 64 configurations.
The proposed structured operation is translated into a query;
program actions use distinct serialized batches obtained from the
encoding choices.
Identical payload bytes are deduplicated.
The catalog contains 68 fixed probes, including delimiter, quoting,
escaping, decimal, date and missing-value fixtures, plus compound
fixtures combining those features.

JSON configurations use records, split, index, columns, values and table
orientations.
The first five orientations have inferred or preserved dtypes; table
uses its schema, giving 11 modes.
Combining them with date conversion on/off, precise floating-point
parsing on/off, and millisecond/nanosecond units gives 88 configurations.
Program candidates vary orientation, date representation and
native/string numeric representation, with byte-identical payloads
deduplicated.
The probe catalog contains 36 fixtures.

\begin{table}[H]
\centering
\caption{Catalog sizes across the 152 retained native input models.
Program counts vary with payload deduplication for the actual source data.}
\label{tab:catalog}
\begin{tabular}{lrrrr}
\toprule
Interface & Input models & Configurations & Distinct program counts & Probes\\
\midrule
CSV & 76 & 64 & 2, 5, 10, 12, 24 & 68\\
JSON & 76 & 88 & 12, 36 & 36\\
\bottomrule
\end{tabular}
\end{table}

\subsection{Responses and commit semantics}

A probe returns parsed column names, data types and rows.
A successful program evaluation returns a fingerprint of its canonical
result. The fingerprint contains the SHA-256 digest and byte length.
CSV parse/query failures return
\texttt{\{error: invalid\_csv\_or\_query\}}.
JSON failures return the exception class and up to 500 characters of
its message.
Policy branches use canonical response keys.
The policy receives the response and commit status; the tool controls
its parser configuration.

The requested structured program and typed source data define the
reference output.
Dates are normalized to UTC ISO representations.
Missing and nonfinite scalar values map to JSON null, and integral
floating-point values map to integers.
Dictionary keys are sorted during canonical serialization; list and row
order follow the operation.
There is no floating-point tolerance in the fingerprint comparison.
Projection and filtering preserve original row order, grouped counts
sort keys by canonical JSON representation, and sorted-row operations
break ties by original row index.

The tool starts a transaction for a program action.
A matching reference fingerprint commits the result and receipt;
a mismatch rolls back.
Diagnostic previews do not publish artifacts.
Fresh replay reconstructs the reference and checks stored content as
well as receipts, observations and call counts.
Reference construction is included in evaluator setup.

\subsection{Binding the running example to recorded effects}

The main-paper example selects configurations 0, 1, 32 and 33 from the
Seattle weather projection record, renamed $h_1,h_2,h_3,h_4$.
They share single-quote handling, doubled escaping, the \texttt{NA}
missing marker and point decimals.
Only delimiter and date interpretation vary.
Program P6 uses the comma/month-first representation: it commits on
$h_4$ and returns \texttt{invalid\_csv\_or\_query} on the other three.
In this four-configuration example, its failure therefore reaches
$V=\{h_1,h_2,h_3\}$ with two calls left from the initial budget of three.

Q13 is the recorded compound CSV fixture with fields
\texttt{amount: DECIMAL(18,2)}, \texttt{due: DATE} and
\texttt{memo: VARCHAR}. Its complete payload is:
\begin{quote}\small
\verb|7.25;05/04/2026;NA|\\
\verb|2.5;07/06/2026;'priority, note; "yes" and ''ok'''|
\end{quote}
The figure abbreviates a successful preview by the first row's
\texttt{due} value. The two date responses and the parse error are
three distinct native observations.

\begin{table}[H]
\centering
\caption{Recorded effects for the optimal local-probe example.
Dates omit the common UTC time suffix. The complete-model audit supplies
these effects; the figure shows an illustrative partial acquisition mask.}
\begin{tabular}{cllll}
\toprule
Configuration & Delimiter & Date parser & Program P6 & Probe Q13\\
\midrule
$h_1$ & semicolon & day-first & parse error & 2026-04-05\\
$h_2$ & comma & day-first & parse error & parse error\\
$h_3$ & semicolon & month-first & parse error & 2026-05-04\\
$h_4$ & comma & month-first & commit & parse error\\
\bottomrule
\end{tabular}
\end{table}

The three leaf programs have catalog indices 0, 1 and 5 and complete
$h_1$, $h_2$ and $h_3$, respectively.
Each of the full catalog's ten programs commits on at most one
configuration in $V$, and all of its failures in $V$ share one response.
A program-first policy therefore has a single failure branch containing
at least two configurations and can complete at most one of them with
its remaining call. Its maximum completion is two out of three.
Q13 separates all three configurations and completes each with its
matching terminal program, using two calls per configuration.
Any policy completing all three in two calls must begin with a probe,
so its total call count is at least six. Q13 attains this lower bound.
It is therefore optimal for the lexicographic objective at $(V,2)$.

Exhaustive enumeration over all ten programs and 68 probes gives
$(3,-6)$ for both the optimum and the Q13 continuation, versus
$(2,-5)$ for the best program-first continuation, even when its remaining
call may use any catalog action.
The catalog-order initial policy uses programs 0, 1 and 2 at $C=3$;
its required configurations within this example are $h_1$ and $h_2$.
Both compared continuations preserve that obligation.
Q13, Q36 and Q52 tie for the optimum; Q13 is selected for its compact
three-field fixture.
The proof concerns the reached state $(V,2)$. The full experimental
runs retain all 64 configurations and the complete catalog.

\subsection{Acquisition order and implementation details}

Algorithm~\ref{alg:cape} gives the compilation loop of the main text.

\begin{center}
\begin{minipage}{.54\linewidth}
\begin{algorithm}[H]
\caption{CAPE policy compilation}
\label{alg:cape}
\small
\begin{algorithmic}[1]
\REQUIRE Initial policy $\pi_0$, configurations $H$, actions $A$, budgets $C,\tau$
\STATE Validate $\pi_0$; set $R_0$ and seed executable witnesses
\STATE Initialize root lower and upper values
\WHILE{resources remain and root values differ}
  \IF{the effect table is complete}
    \STATE Run exact search with the remaining allowance
    \STATE \textbf{break}
  \ENDIF
  \STATE Propagate pending witness and bound updates
  \STATE Expand a closed action on an optimistic alternative
  \STATE Form frontier with the current executable policy
  \STATE Select an unfinished action by obligation and gain
  \IF{an unfinished action can improve its state}
    \REPEAT
      \STATE Acquire a missing effect, required configurations first
      \STATE Update its bound; tighten two-call continuations
    \UNTIL{closed, infeasible, dominated, or interrupted}
  \ENDIF
  \STATE Compose closed actions and update their parents
\ENDWHILE
\RETURN Best executable policy; certify if root values agree
\end{algorithmic}
\end{algorithm}
\end{minipage}
\end{center}

An acquisition score first distinguishes states without a feasible
required witness.
Such states are ranked by required configurations per missing cell.
For a state with a witness, positive completion gain precedes call-only
improvement; completion gain and call savings are divided by the missing
cell count.
Smaller missing scopes break remaining ties, followed by program actions
and lower catalog indices.
Within the selected scope, required configurations are acquired first,
then catalog configuration order applies.

Native parents and upper-bound parents are stored separately.
Closed alternatives are activated lazily, and pending updates are
deduplicated.
Closed one- and two-call calculations seed witnesses without resolving
every larger subproblem.
The selected acquisition uses the tighter continuation calculation;
unselected unfinished actions use support bounds.
New effects invalidate the affected scope or program-generation caches.
At model completion, the shared compact exact solver handles the
remaining recurrence.

The full-row ablation continues querying the rest of an action's row
after it closes on its selected scope.
The no-tightening ablation retains the same frontier and witnesses while
using support bounds.
The matched AND/OR control follows the best optimistic solution path.
At a closed action it descends into a child with the largest unresolved
value gap, prioritizing missing required witnesses.
Its none, selected and always variants use the same bound operators as
CAPE.
Always additionally invalidates affected continuation bounds after probe
acquisitions, because a new probe response can change another action's
tight continuation.

\section{Workload and measurement protocol}
\label{app:protocol}

\subsection{Sources and request construction}

The eight source files are drawn from the Vega datasets cited in the
main paper.
Source identities and operation rules were fixed before retrieval.
Preparation retains the first at most 256 records, for 1,850 records
from 17,681 original records.
Table~\ref{tab:sources} gives the source sizes.

\begin{table}[H]
\centering
\caption{Source files and fixed retained prefixes. Fields exclude the internal row index.}
\label{tab:sources}
\begin{tabular}{lrrr}
\toprule
Source & Original rows & Retained rows & Fields \\
\midrule
Airports & 3,376 & 256 & 7 \\
Bird strikes & 10,000 & 256 & 14 \\
CO$_2$ concentration & 741 & 256 & 3 \\
Disasters & 803 & 256 & 3 \\
Development indicators & 682 & 256 & 6 \\
Political contributions & 58 & 58 & 25 \\
Seattle weather & 1,461 & 256 & 6 \\
Stock prices & 560 & 256 & 3 \\
\bottomrule
\end{tabular}

\end{table}

CSV preparation infers scalar types over source columns before selecting
the prefix.
Empty fields become missing values.
Exact signed 64-bit integers are retained; numeric conversion must
preserve integer values. Leading-zero strings and whitespace remain
strings.
JSON is parsed strictly.
The common preparer expands nested paths and lists of at most 16 elements;
remaining containers are canonical JSON text.
An internal original-row index is added.

Each source yields eight requests.
Projection selects all source columns in fixed order.
Row count returns the batch size; missing-value counts cover every
source column.
Grouped count uses the first string column, falling back to the first
source column, includes missing groups, and sorts keys canonically.
Numeric summaries return nonmissing count, minimum, maximum and sum
for every numeric field, with zero sum for an empty column.
Sorting returns the first 20 rows ordered by the selected field and then
original row order, with missing keys last.
Numeric filtering selects row indices above the field's lower median;
union filtering adds a category predicate for the most frequent
nonmissing value.
Field selection follows schema order, preferring varying numeric fields
and dates for sorting.
Canonical ordering breaks category frequency ties.

\subsection{Proposal generation}

The mechanism study uses DeepSeek-V4-Pro at temperature zero with an
8,192-token completion allowance.
Prompts contain the request, typed schema, record count and first five
records.
Each request has one completed proposal.
Of 64 outputs, 63 are valid and match the typed reference.
The invalid development-indicator filtering response combines unsupported
predicate and sorting fields; it contributes a pipeline failure to every
method.
These requests report 54,216 input and completion tokens and include two
rate-limit retries.
The fourteen-deployment panel is documented in Sec.~\ref{app:panel}.

\subsection{Measurements and statistical units}

All three studies use both interfaces, execution budgets $C=3/5$,
empty caches and a soft 30-second compilation allowance.
Primary jobs have two fresh-process repetitions.
The method queue is randomized, with eight single-threaded workers
assigned to distinct physical cores.
The runtime uses Python 3.12.13, pandas 3.0.2 and DuckDB 1.5.5.
Python hash seed is 2070; native and BLAS query threads are limited
to one.
DuckDB connections disable automatic thread pinning and use one query
thread from creation.
Thread affinities are checked before and after timing.

Lifecycle time includes evaluator setup, reference construction,
validation of the initial policy, acquisition, search, policy serialization and
cleanup.
Interpreter startup and proposal generation are recorded outside this
lifecycle.
Atomic operations and cleanup may overrun the soft deadline, and actual
durations are retained.
CPU time and effect-query counts accompany wall time.
Complete-table audits and fresh publication replay follow compilation.

Completion averages over all proposals, including invalid proposals as
zero.
Native costs and independent-optimality comparisons condition on
schema-valid proposals.
Runtime-certificate percentages include invalid proposals in their
denominator.
Each source contributes its mean over operations and repetitions, giving
equal source weight.
Paired intervals use 5,000 bootstrap resamples of the eight source means,
with seed 38117.
Shared native inputs and timing repetitions remain inside their source
clusters.

\subsection{Independent checks}

Audits acquire complete native effect tables, reexecute probes and solve
the constrained objective using independent dynamic programming.
Initial and final policies are replayed with fresh publication state.
The replay checks attempted requests, observations, success indicators,
call counts, receipts and committed content.
Each saved CAPE witness is also checked against the effects already
available at that point, its original required set and the independent
optimum between its lower and upper values.

The mechanism study resolves all 252 native objectives, checks 8,970
acquired-cell witnesses, and replays 586 distinct policy pairs.
The panel resolves 304 objectives across 152 native models.
The matched study verifies 32,182 historical witnesses and 414 distinct
policy pairs, covering 61,440 native episodes and 133,066 calls.
Instrumented cost profiles have separate timing and quality records.

\section{Complete mechanism-study results}
\label{app:mechanism-results}

Tables~\ref{tab:complete-csv3}--\ref{tab:complete-json5} retain all eight
methods.
Success is pipeline completion, calls are per configuration, queries
are calibration evaluations, and time is compilation lifecycle.
Cert.\ is the runtime-certificate rate over all proposals; Opt.\ is the
rate of attaining the independently computed full-catalog optimum among
executable proposals.
All means balance the eight sources.

\begin{figure}[H]
\centering
\includegraphics[width=.86\linewidth]{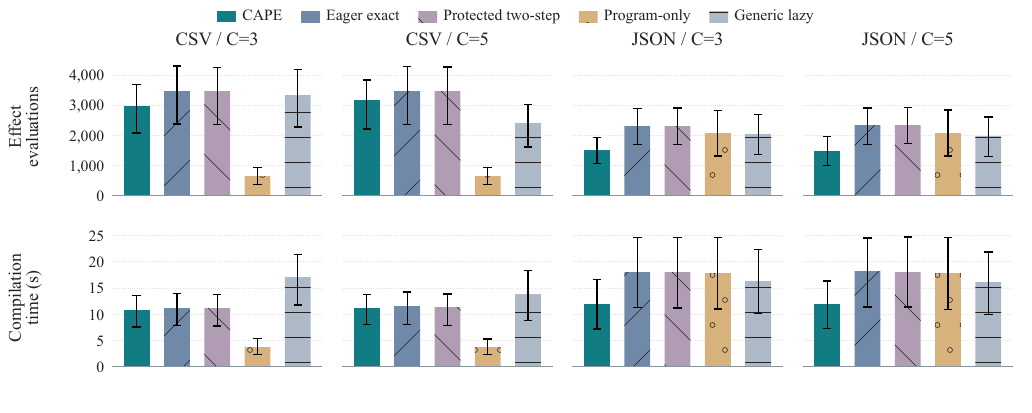}
\caption{\textbf{Local acquisition saves work relative to eager search.}
DeepSeek mechanism study: source-balanced means and 95\% source-cluster
intervals, with shared scales within rows. Time includes setup,
calibration, search and cleanup.}
\label{fig:cost}
\end{figure}

\begin{table}[H]
\centering\small
\caption{CSV, $C=3$.}
\label{tab:complete-csv3}
\fitinput{\begin{tabular}{lrrrrrr}
\toprule
Method & Success (\%) & Calls & Queries & Time (s) & Cert. (\%) & Opt. (\%) \\
\midrule
CAPE & 85.16 & 2.196 & 2980.5 & 10.774 & 98.44 & 100.00 \\
Eager exact & 85.16 & 2.196 & 3450.9 & 11.173 & 98.44 & 100.00 \\
Program-only & 76.76 & 2.022 & 624.0 & 3.631 & 34.38 & 35.04 \\
Protected two-step & 83.98 & 2.474 & 3450.9 & 11.121 & 34.38 & 35.04 \\
Generic lazy & 83.74 & 2.198 & 3332.5 & 17.089 & 93.75 & 95.31 \\
Program-first expansion & 85.16 & 2.196 & 3307.3 & 11.402 & 98.44 & 100.00 \\
Full-row acquisition & 85.16 & 2.196 & 3243.6 & 11.311 & 98.44 & 100.00 \\
No tight bounds & 85.16 & 2.196 & 3330.3 & 11.374 & 98.44 & 100.00 \\
\bottomrule
\end{tabular}
}
\end{table}
\begin{table}[H]
\centering\small
\caption{CSV, $C=5$.}
\label{tab:complete-csv5}
\fitinput{\begin{tabular}{lrrrrrr}
\toprule
Method & Success (\%) & Calls & Queries & Time (s) & Cert. (\%) & Opt. (\%) \\
\midrule
CAPE & 98.44 & 2.348 & 3161.1 & 11.201 & 98.44 & 100.00 \\
Eager exact & 98.44 & 2.348 & 3450.9 & 11.417 & 98.44 & 100.00 \\
Program-only & 92.19 & 2.365 & 624.0 & 3.624 & 34.38 & 75.00 \\
Protected two-step & 98.44 & 2.838 & 3450.9 & 11.192 & 34.38 & 70.31 \\
Generic lazy & 84.18 & 2.534 & 2382.0 & 13.700 & 73.44 & 75.00 \\
Program-first expansion & 98.44 & 2.348 & 3191.6 & 11.572 & 98.44 & 100.00 \\
Full-row acquisition & 98.44 & 2.348 & 3161.0 & 11.177 & 98.44 & 100.00 \\
No tight bounds & 98.44 & 2.348 & 3161.1 & 11.217 & 98.44 & 100.00 \\
\bottomrule
\end{tabular}
}
\end{table}
\begin{table}[H]
\centering\small
\caption{JSON, $C=3$.}
\label{tab:complete-json3}
\fitinput{\begin{tabular}{lrrrrrr}
\toprule
Method & Success (\%) & Calls & Queries & Time (s) & Cert. (\%) & Opt. (\%) \\
\midrule
CAPE & 94.74 & 1.476 & 1506.2 & 11.919 & 95.31 & 98.44 \\
Eager exact & 94.60 & 1.474 & 2309.0 & 18.021 & 84.38 & 98.44 \\
Program-only & 94.60 & 1.474 & 2075.7 & 17.792 & 77.34 & 98.44 \\
Protected two-step & 94.60 & 1.493 & 2307.2 & 18.008 & 77.34 & 91.96 \\
Generic lazy & 92.90 & 1.495 & 2029.5 & 16.200 & 93.75 & 95.31 \\
Program-first expansion & 94.60 & 1.474 & 2170.0 & 17.878 & 85.94 & 98.44 \\
Full-row acquisition & 90.48 & 1.430 & 2354.3 & 18.003 & 85.16 & 86.72 \\
No tight bounds & 94.74 & 1.476 & 1516.4 & 11.921 & 95.31 & 98.44 \\
\bottomrule
\end{tabular}
}
\end{table}
\begin{table}[H]
\centering\small
\caption{JSON, $C=5$.}
\label{tab:complete-json5}
\fitinput{\begin{tabular}{lrrrrrr}
\toprule
Method & Success (\%) & Calls & Queries & Time (s) & Cert. (\%) & Opt. (\%) \\
\midrule
CAPE & 94.74 & 1.475 & 1485.9 & 11.903 & 96.88 & 100.00 \\
Eager exact & 94.60 & 1.474 & 2317.8 & 18.079 & 84.38 & 98.44 \\
Program-only & 94.60 & 1.474 & 2077.9 & 17.744 & 76.56 & 98.44 \\
Protected two-step & 94.60 & 1.493 & 2314.1 & 18.005 & 77.34 & 91.96 \\
Generic lazy & 92.90 & 1.546 & 1954.9 & 16.008 & 93.75 & 95.31 \\
Program-first expansion & 94.60 & 1.474 & 2174.5 & 17.919 & 83.59 & 98.44 \\
Full-row acquisition & 91.05 & 1.435 & 2049.3 & 17.036 & 86.72 & 89.84 \\
No tight bounds & 94.74 & 1.475 & 1486.2 & 11.862 & 96.88 & 100.00 \\
\bottomrule
\end{tabular}
}
\end{table}
\FloatBarrier

\begin{table}[H]
\centering\small
\caption{Paired contrasts for the mechanism study.
Positive differences favor CAPE; calls are interpreted together with
completion.
Intervals are for the seconds saved and resample sources.}
\label{tab:mechanism-contrasts}
\fitinput{\begin{tabular}{llrrrr}
\toprule
Interface, $C$ & Control & Gain (pp) & Calls saved & Seconds saved & 95\% interval \\
\midrule
CSV, 3 & Eager exact & 0.00 & 0.000 & 0.399 & [0.041, 0.768] \\
CSV, 3 & Program-only & 8.40 & -0.174 & -7.143 & [-9.004, -4.786] \\
CSV, 3 & Protected two-step & 1.17 & 0.278 & 0.348 & [-0.054, 0.713] \\
CSV, 3 & Generic lazy & 1.42 & 0.001 & 6.316 & [3.972, 8.385] \\
CSV, 3 & Program-first expansion & 0.00 & 0.000 & 0.628 & [-0.076, 1.255] \\
CSV, 3 & Full-row acquisition & 0.00 & 0.000 & 0.538 & [-0.071, 1.050] \\
CSV, 3 & No tight bounds & 0.00 & 0.000 & 0.601 & [0.113, 1.076] \\
JSON, 3 & Eager exact & 0.14 & -0.002 & 6.102 & [4.074, 8.107] \\
JSON, 3 & Program-only & 0.14 & -0.002 & 5.873 & [3.659, 8.080] \\
JSON, 3 & Protected two-step & 0.14 & 0.017 & 6.088 & [4.061, 8.108] \\
JSON, 3 & Generic lazy & 1.85 & 0.019 & 4.281 & [2.827, 5.757] \\
JSON, 3 & Program-first expansion & 0.14 & -0.002 & 5.959 & [3.876, 8.132] \\
JSON, 3 & Full-row acquisition & 4.26 & -0.046 & 6.084 & [4.122, 8.051] \\
JSON, 3 & No tight bounds & 0.00 & 0.000 & 0.002 & [-0.022, 0.026] \\
CSV, 5 & Eager exact & 0.00 & 0.000 & 0.216 & [-0.125, 0.693] \\
CSV, 5 & Program-only & 6.25 & 0.017 & -7.577 & [-9.510, -5.041] \\
CSV, 5 & Protected two-step & 0.00 & 0.490 & -0.010 & [-0.558, 0.562] \\
CSV, 5 & Generic lazy & 14.26 & 0.187 & 2.498 & [0.342, 4.938] \\
CSV, 5 & Program-first expansion & 0.00 & 0.000 & 0.371 & [0.146, 0.729] \\
CSV, 5 & Full-row acquisition & 0.00 & 0.000 & -0.024 & [-0.124, 0.080] \\
CSV, 5 & No tight bounds & 0.00 & 0.000 & 0.015 & [-0.033, 0.065] \\
JSON, 5 & Eager exact & 0.14 & -0.001 & 6.176 & [4.131, 8.157] \\
JSON, 5 & Program-only & 0.14 & -0.001 & 5.842 & [3.735, 8.026] \\
JSON, 5 & Protected two-step & 0.14 & 0.017 & 6.103 & [4.136, 8.141] \\
JSON, 5 & Generic lazy & 1.85 & 0.071 & 4.105 & [2.774, 5.458] \\
JSON, 5 & Program-first expansion & 0.14 & -0.001 & 6.016 & [3.899, 8.168] \\
JSON, 5 & Full-row acquisition & 3.69 & -0.040 & 5.134 & [2.801, 7.523] \\
JSON, 5 & No tight bounds & 0.00 & 0.000 & -0.041 & [-0.100, 0.016] \\
\bottomrule
\end{tabular}
}
\end{table}
\FloatBarrier

\subsection{Diagnostic headroom and local acquisition}

\begin{figure}[H]
\centering
\includegraphics[width=.82\linewidth]{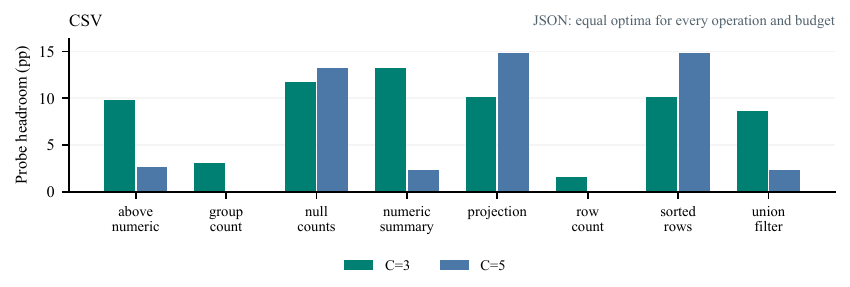}
\caption{\textbf{Diagnostics expand achievable completion on CSV.}
Full-catalog optimal completion minus independently computed program-only
optimal completion, using complete effects.
This comparison separates action-space headroom from interrupted search.}
\label{fig:headroom}
\end{figure}

The complete-model diagnostic comparison optimizes the same request,
required set and call budget twice, with and without probes.
Figure~\ref{fig:headroom} shows completion headroom.
Its full-catalog optimum includes every program and probe effect.
The program-only optimum retains program observations and retry policies.
CSV contains instances where diagnostic actions increase achievable
completion; the JSON completion optima coincide.

\begin{figure}[H]
\centering
\includegraphics[width=.82\linewidth]{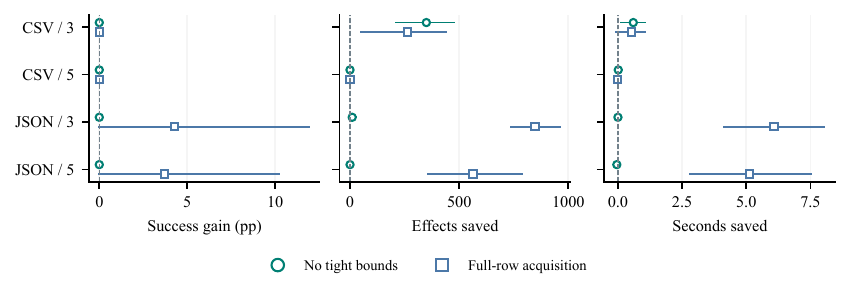}
\caption{\textbf{Scope closure and bound tightening affect different work.}
The mechanism-study ablations replace local closure by full-row
acquisition or selected tightening by cheap support bounds.
Paired intervals resample the source clusters.}
\label{fig:ablations}
\end{figure}

In the timed CAPE runs, 380 compilations acquire effects on strict
descendant scopes.
Of 41,626 acquisition chunks, 23,180 are local and contain 308,858
effect queries.
The full-row ablation measures the cost of extending these chunks to
unused configuration cells.
Figure~\ref{fig:ablations} gives the source-cluster comparisons.

\FloatBarrier

\subsection{Policy progress and proof completion}

\begin{figure}[H]
\centering
\includegraphics[width=.86\linewidth]{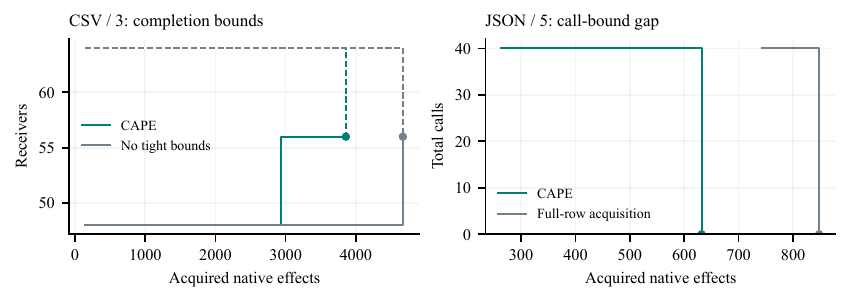}
\caption{\textbf{Policy improvement can precede proof completion.}
Retained development trajectories show acquired effects and lower/upper
policy values.
These traces illustrate the selected mechanism's progress; the primary
comparisons use the separate frozen measurement study.}
\label{fig:trajectories}
\end{figure}

Figure~\ref{fig:trajectories} retains the development trajectories for
CSV projection and JSON numeric summary.
On the CSV request, the feasible policy completes 56 configurations
before the completion bounds meet.
On the JSON request, completion bounds meet before the call bounds;
subsequent acquisition closes the call gap.
The trajectories show the distinction between improving the executable
policy and proving its final value.

\FloatBarrier

\section{Matched acquisition schedules and bound computation}
\label{app:matched}

The matched study uses 304 exact task/program/interface/budget groups.
Each group receives six methods and two fresh-process primary
repetitions, yielding 3,648 measurements.
One additional instrumented run per method and group yields 1,824
profiles.
The experimental machine is an AMD EPYC 7742.
Source weighting, bootstrap units and lifecycle accounting follow
Sec.~\ref{app:protocol}.

CAPE/selective attains 606/608 physical full-catalog optima, while
matched AND/OR/selective attains 595/608.
The two CAPE misses have optimal completion and excess calls on JSON
weather projection at $C=3$.
The tables below use DeepSeek-V4-Pro proposal bindings.
They retain the six configurations and their signed paired contrasts.

\begin{figure}[H]
\centering
\includegraphics[width=.86\linewidth]{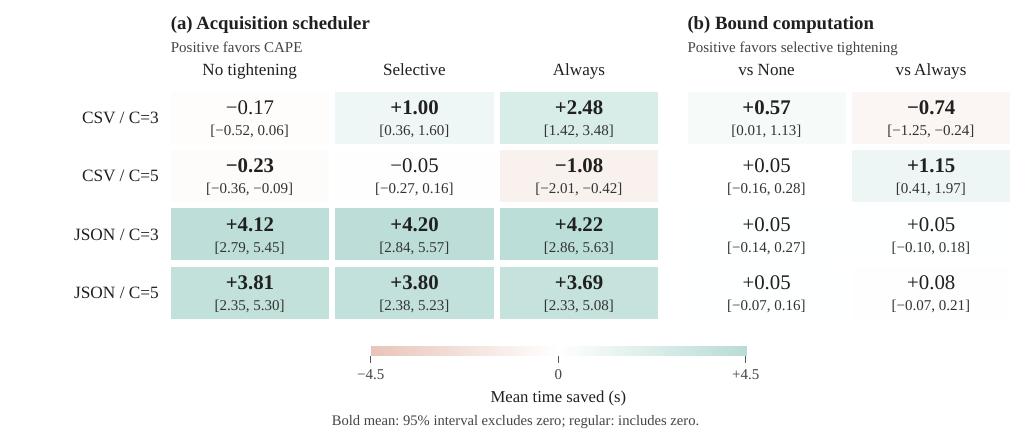}
\caption{\textbf{Paired effects separate acquisition and tightening.}
(a) CAPE versus matched AND/OR with the same bound mode;
(b) selective versus no/always tightening.
Cells give mean seconds saved and 95\% source-cluster intervals.
Positive values favor CAPE in (a) and selective computation in (b).}
\label{fig:matched}
\end{figure}

\begin{table}[H]
\centering\small
\caption{Matched primary comparisons on CSV, $C=3$.}
\fitinput{\begin{tabular}{llrrrrrr}
\toprule
Scheduler & Bounds & Success (\%) & Calls & Queries & Time (s) & Cert. (\%) & Opt. (\%) \\
\midrule
CAPE & No tightening & 85.16 & 2.196 & 3330.3 & 11.489 & 98.44 & 100.00 \\
CAPE & Selective & 85.16 & 2.196 & 2979.9 & 10.918 & 97.66 & 100.00 \\
CAPE & Always & 85.16 & 2.196 & 2546.4 & 10.177 & 98.44 & 100.00 \\
AND/OR & No tightening & 85.16 & 2.196 & 3278.6 & 11.315 & 98.44 & 100.00 \\
AND/OR & Selective & 85.16 & 2.196 & 3250.5 & 11.916 & 98.44 & 100.00 \\
AND/OR & Always & 85.16 & 2.196 & 3249.7 & 12.654 & 98.44 & 100.00 \\
\bottomrule
\end{tabular}
}
\end{table}
\begin{table}[H]
\centering\small
\caption{Matched primary comparisons on CSV, $C=5$.}
\fitinput{\begin{tabular}{llrrrrrr}
\toprule
Scheduler & Bounds & Success (\%) & Calls & Queries & Time (s) & Cert. (\%) & Opt. (\%) \\
\midrule
CAPE & No tightening & 98.44 & 2.348 & 3161.1 & 11.374 & 98.44 & 100.00 \\
CAPE & Selective & 98.44 & 2.348 & 3161.1 & 11.319 & 98.44 & 100.00 \\
CAPE & Always & 98.44 & 2.348 & 3148.1 & 12.464 & 93.75 & 100.00 \\
AND/OR & No tightening & 98.44 & 2.348 & 3163.4 & 11.148 & 98.44 & 100.00 \\
AND/OR & Selective & 98.44 & 2.348 & 3163.4 & 11.265 & 98.44 & 100.00 \\
AND/OR & Always & 98.44 & 2.348 & 3163.4 & 11.381 & 98.44 & 100.00 \\
\bottomrule
\end{tabular}
}
\end{table}
\begin{table}[H]
\centering\small
\caption{Matched primary comparisons on JSON, $C=3$.}
\fitinput{\begin{tabular}{llrrrrrr}
\toprule
Scheduler & Bounds & Success (\%) & Calls & Queries & Time (s) & Cert. (\%) & Opt. (\%) \\
\midrule
CAPE & No tightening & 94.74 & 1.477 & 1505.7 & 12.007 & 95.31 & 97.66 \\
CAPE & Selective & 94.74 & 1.476 & 1511.9 & 11.961 & 96.88 & 98.44 \\
CAPE & Always & 94.74 & 1.476 & 1505.0 & 12.012 & 95.31 & 98.44 \\
AND/OR & No tightening & 94.39 & 1.481 & 1980.3 & 16.130 & 92.97 & 94.53 \\
AND/OR & Selective & 94.39 & 1.478 & 1990.5 & 16.161 & 92.97 & 94.53 \\
AND/OR & Always & 94.46 & 1.483 & 1975.3 & 16.235 & 93.75 & 95.31 \\
\bottomrule
\end{tabular}
}
\end{table}
\begin{table}[H]
\centering\small
\caption{Matched primary comparisons on JSON, $C=5$.}
\fitinput{\begin{tabular}{llrrrrrr}
\toprule
Scheduler & Bounds & Success (\%) & Calls & Queries & Time (s) & Cert. (\%) & Opt. (\%) \\
\midrule
CAPE & No tightening & 94.74 & 1.475 & 1494.0 & 11.975 & 97.66 & 100.00 \\
CAPE & Selective & 94.74 & 1.475 & 1490.5 & 11.929 & 96.88 & 100.00 \\
CAPE & Always & 94.74 & 1.475 & 1473.9 & 12.005 & 96.09 & 100.00 \\
AND/OR & No tightening & 94.50 & 1.483 & 1941.7 & 15.784 & 93.75 & 95.31 \\
AND/OR & Selective & 94.53 & 1.483 & 1949.4 & 15.728 & 93.75 & 95.31 \\
AND/OR & Always & 94.42 & 1.480 & 1932.6 & 15.699 & 93.75 & 95.31 \\
\bottomrule
\end{tabular}
}
\end{table}
\FloatBarrier

\begin{table}[H]
\centering\small
\caption{CSV paired contrasts.
C and A denote CAPE and matched AND/OR; N, S and T denote no,
selected and always tightening.
Positive differences favor the named candidate.}
\fitinput{\begin{tabular}{rllrrrr}
\toprule
$C$ & Candidate & Control & Gain (pp) & Queries saved & Seconds saved & 95\% interval \\
\midrule
3 & C-N & A-N & 0.000 & -51.7 & -0.173 & [-0.522, 0.061] \\
3 & C-S & A-S & 0.000 & 270.6 & 0.998 & [0.358, 1.596] \\
3 & C-T & A-T & 0.000 & 703.4 & 2.477 & [1.420, 3.479] \\
3 & C-S & C-N & 0.000 & 350.4 & 0.571 & [0.009, 1.130] \\
3 & C-S & C-T & 0.000 & -433.6 & -0.741 & [-1.255, -0.242] \\
3 & A-S & A-N & 0.000 & 28.1 & -0.600 & [-0.869, -0.291] \\
3 & A-S & A-T & 0.000 & -0.8 & 0.738 & [0.512, 0.925] \\
5 & C-N & A-N & 0.000 & 2.2 & -0.226 & [-0.357, -0.095] \\
5 & C-S & A-S & 0.000 & 2.2 & -0.054 & [-0.273, 0.163] \\
5 & C-T & A-T & 0.000 & 15.2 & -1.082 & [-2.013, -0.421] \\
5 & C-S & C-N & 0.000 & 0.0 & 0.055 & [-0.160, 0.278] \\
5 & C-S & C-T & 0.000 & -13.0 & 1.145 & [0.413, 1.966] \\
5 & A-S & A-N & 0.000 & 0.0 & -0.117 & [-0.275, 0.039] \\
5 & A-S & A-T & 0.000 & 0.0 & 0.117 & [-0.134, 0.391] \\
\bottomrule
\end{tabular}
}
\end{table}
\begin{table}[H]
\centering\small
\caption{JSON paired contrasts.
Completion gains are percentage points; time is lifecycle seconds.
The two selective-scheduler completion-gain intervals include zero.}
\fitinput{\begin{tabular}{rllrrrr}
\toprule
$C$ & Candidate & Control & Gain (pp) & Queries saved & Seconds saved & 95\% interval \\
\midrule
3 & C-N & A-N & 0.355 & 474.6 & 4.123 & [2.785, 5.446] \\
3 & C-S & A-S & 0.355 & 478.6 & 4.200 & [2.839, 5.570] \\
3 & C-T & A-T & 0.284 & 470.3 & 4.223 & [2.860, 5.630] \\
3 & C-S & C-N & 0.000 & -6.2 & 0.046 & [-0.138, 0.268] \\
3 & C-S & C-T & 0.000 & -6.9 & 0.051 & [-0.097, 0.176] \\
3 & A-S & A-N & 0.000 & -10.2 & -0.031 & [-0.203, 0.145] \\
3 & A-S & A-T & -0.071 & -15.3 & 0.074 & [-0.103, 0.239] \\
5 & C-N & A-N & 0.249 & 447.8 & 3.809 & [2.348, 5.302] \\
5 & C-S & A-S & 0.213 & 458.9 & 3.799 & [2.377, 5.228] \\
5 & C-T & A-T & 0.320 & 458.7 & 3.694 & [2.332, 5.081] \\
5 & C-S & C-N & 0.000 & 3.5 & 0.046 & [-0.070, 0.164] \\
5 & C-S & C-T & 0.000 & -16.6 & 0.076 & [-0.066, 0.210] \\
5 & A-S & A-N & 0.036 & -7.7 & 0.056 & [-0.100, 0.207] \\
5 & A-S & A-T & 0.107 & -16.9 & -0.029 & [-0.217, 0.151] \\
\bottomrule
\end{tabular}
}
\end{table}
\FloatBarrier

\subsection{Independent cost profiles}

The profiler uses exclusive nested wall-clock accounting.
Time spent in descendant bound calls inside $T_2$ is charged to tight
bounds.
Other support/action bounds and graph search have separate buckets.
Compiler-other includes initial validation and bookkeeping;
lifecycle-other includes final encoding, cleanup and measurement
overhead.
Instrumentation can change deadline termination, so these runs retain
their own quality records.
Of 1,824 profiles, 1,805 match both primary repetitions in success
identities and call mass.

\begin{figure}[H]
\centering
\includegraphics[width=.86\linewidth]{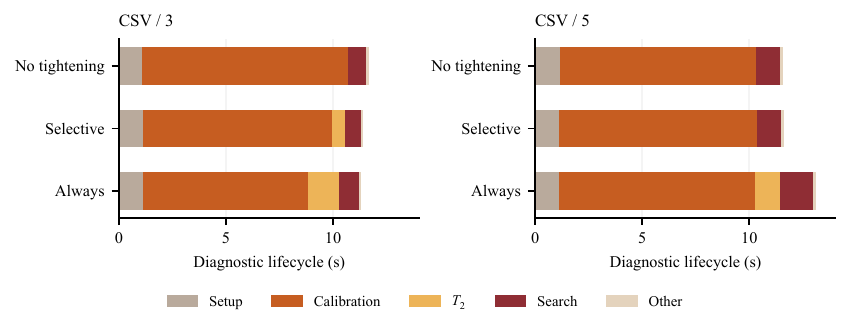}
\caption{\textbf{Tightening exchanges computation for calibration.}
Separate CSV diagnostic runs with DeepSeek-V4-Pro bindings.
At $C=3$, always tightening reduces calibration work enough to offset
its added computation.
At $C=5$, additional bound time accompanies similar calibration work.}
\label{fig:profiles}
\end{figure}
\begin{table}[H]
\centering\small
\caption{Exclusive diagnostic costs in seconds.
Search groups graph work, other bounds and complete-model solving.
Agree compares each profile's value with both primary repetitions;
$\Delta$ queries is the difference from the primary mean.
Method abbreviations follow the contrast tables.}
\fitinput{\begin{tabular}{llrrrrrrr}
\toprule
Setting & Method & Setup & Calibration & $T_2$ & Search & Other & Agree (\%) & $\Delta$ queries \\
\midrule
CSV/3 & C-N & 1.103 & 9.612 & 0.000 & 0.844 & 0.110 & 100.00 & 0.0 \\
CSV/3 & C-S & 1.122 & 8.814 & 0.621 & 0.747 & 0.114 & 100.00 & 0.5 \\
CSV/3 & C-T & 1.149 & 7.658 & 1.468 & 0.925 & 0.108 & 100.00 & -1.8 \\
CSV/3 & A-N & 1.114 & 9.492 & 0.000 & 0.696 & 0.108 & 100.00 & 0.0 \\
CSV/3 & A-S & 1.139 & 9.571 & 1.290 & 0.617 & 0.118 & 100.00 & 0.0 \\
CSV/3 & A-T & 1.157 & 9.620 & 2.266 & 0.653 & 0.114 & 96.88 & -18.9 \\
CSV/5 & C-N & 1.152 & 9.155 & 0.000 & 1.109 & 0.134 & 100.00 & 0.0 \\
CSV/5 & C-S & 1.101 & 9.259 & 0.019 & 1.115 & 0.134 & 100.00 & 0.0 \\
CSV/5 & C-T & 1.101 & 9.148 & 1.192 & 1.539 & 0.133 & 100.00 & -39.3 \\
CSV/5 & A-N & 1.121 & 9.265 & 0.000 & 1.025 & 0.133 & 100.00 & 0.0 \\
CSV/5 & A-S & 1.110 & 9.243 & 0.000 & 1.043 & 0.137 & 100.00 & 0.0 \\
CSV/5 & A-T & 1.111 & 9.315 & 0.265 & 1.040 & 0.136 & 96.88 & -8.0 \\
JSON/3 & C-N & 0.431 & 11.307 & 0.000 & 0.349 & 0.093 & 98.44 & 3.0 \\
JSON/3 & C-S & 0.454 & 11.139 & 0.070 & 0.276 & 0.094 & 100.00 & -11.6 \\
JSON/3 & C-T & 0.435 & 11.183 & 0.100 & 0.277 & 0.094 & 98.44 & -20.4 \\
JSON/3 & A-N & 0.446 & 15.561 & 0.000 & 0.312 & 0.096 & 93.75 & 10.8 \\
JSON/3 & A-S & 0.424 & 15.541 & 0.146 & 0.212 & 0.097 & 95.31 & -10.1 \\
JSON/3 & A-T & 0.418 & 15.581 & 0.154 & 0.214 & 0.097 & 96.88 & 3.0 \\
JSON/5 & C-N & 0.414 & 11.282 & 0.000 & 0.326 & 0.096 & 100.00 & -8.9 \\
JSON/5 & C-S & 0.421 & 11.364 & 0.004 & 0.326 & 0.097 & 100.00 & -5.4 \\
JSON/5 & C-T & 0.428 & 11.336 & 0.006 & 0.342 & 0.096 & 100.00 & 8.0 \\
JSON/5 & A-N & 0.455 & 14.989 & 0.000 & 0.295 & 0.098 & 98.44 & -29.0 \\
JSON/5 & A-S & 0.420 & 15.188 & 0.000 & 0.303 & 0.098 & 98.44 & -6.4 \\
JSON/5 & A-T & 0.427 & 15.025 & 0.000 & 0.309 & 0.098 & 96.88 & 15.2 \\
\bottomrule
\end{tabular}
}
\end{table}
\FloatBarrier

\section{Proposal models and complete panel results}
\label{app:panel}

The panel contains four closed deployments and ten open-weight
deployments.
Every deployment receives the same 64 requests.
The prompt specifies the operation schema, typed fields, record count
and first five records, with an 8,192-token completion allowance.
GPT deployments use low reasoning and default sampling;
Qwen3.5 uses nonthinking mode and temperature zero;
Kimi-K2.5 uses its nonreasoning option.
Other deployments use temperature zero.
The VL models receive the same text-only requests.
Table~\ref{tab:panel-settings} records the reasoning settings.

\begin{table}[H]
\centering\small
\caption{Proposal deployments and reasoning settings.
Default indicates an omitted reasoning option.}
\label{tab:panel-settings}
\begin{tabular}{ll}
\toprule
Model & Reasoning setting \\
\midrule
GPT-6 Astra & low \\
GPT-5.6 Terra & low \\
GPT-5.6 Luna & low \\
Grok 4.6 & default \\
DeepSeek-V4-Pro & default \\
DeepSeek-V4-Flash & default \\
Kimi-K2.5 & none \\
Qwen3.5-122B-A10B & nonthinking \\
Llama-3.3-70B-Instruct & default \\
Qwen3.5-27B & nonthinking \\
Qwen3.5-9B & nonthinking \\
Llama-3.1-8B-Instruct & default \\
Qwen2.5-VL-7B-Instruct & default \\
Qwen3-VL-4B-Instruct & default \\
\bottomrule
\end{tabular}

\end{table}
\begin{table}[H]
\centering\small
\caption{Proposal outcomes and reported token usage.
Each deployment has 64 requests.
Reference agreement evaluates the proposed operation on the typed data.}
\label{tab:panel-proposals}
\fitinput{\begin{tabular}{llrrr}
\toprule
Model & Access & Valid / 64 & Reference / 64 & Tokens \\
\midrule
GPT-6 Astra & Closed & 64 & 64 & 52,099 \\
GPT-5.6 Terra & Closed & 64 & 64 & 52,291 \\
GPT-5.6 Luna & Closed & 64 & 64 & 53,843 \\
Grok 4.6 & Closed & 64 & 64 & 87,453 \\
DeepSeek-V4-Pro & Open & 63 & 63 & 54,216 \\
DeepSeek-V4-Flash & Open & 64 & 64 & 53,673 \\
Kimi-K2.5 & Open & 64 & 64 & 52,926 \\
Qwen3.5-122B-A10B & Open & 64 & 64 & 57,984 \\
Llama-3.3-70B-Instruct & Open & 62 & 62 & 54,697 \\
Qwen3.5-27B & Open & 64 & 64 & 57,461 \\
Qwen3.5-9B & Open & 64 & 64 & 57,700 \\
Llama-3.1-8B-Instruct & Open & 47 & 41 & 57,619 \\
Qwen2.5-VL-7B-Instruct & Open & 62 & 59 & 56,554 \\
Qwen3-VL-4B-Instruct & Open & 64 & 61 & 56,337 \\
\bottomrule
\end{tabular}
}
\end{table}
\FloatBarrier

The panel retains the 64 DeepSeek-V4-Pro responses and adds 832 observed
completed responses.
There are 874 schema-valid proposals and 862 typed-reference agreements.
Each completed response is used once, including invalid outcomes.
The collection records 18 transport errors and two interrupted requests
with no observed response.
New completed requests report 750,637 tokens; including retained Pro
responses gives 804,853.

Exact-input deduplication matches task bytes, batch bytes, schema,
requested operation, proposed operation, backend and call allowance.
The panel contains 4,864 fresh physical measurements and 57,344
provider-bound records, including failed proposals.
These bindings attach different proposal-source labels to the same
measured computational problem where inputs coincide.
The source remains the statistical resampling unit.

The native audit covers 152 input models, 33 normalized action
structures and 655 distinct policy pairs.
It resolves all 304 independent objectives and checks 10,225
acquired-cell witnesses, 94,160 native episodes and 218,266 calls.
CAPE attains 606/608 physical optima, and 526/528 on reference-agreeing
proposals.
A schema-valid incorrect operation can have zero achievable completion;
independent optimality is evaluated for the proposed operation and
requested-output guard.

\begin{figure}[H]
\centering
\includegraphics[width=.87\linewidth]{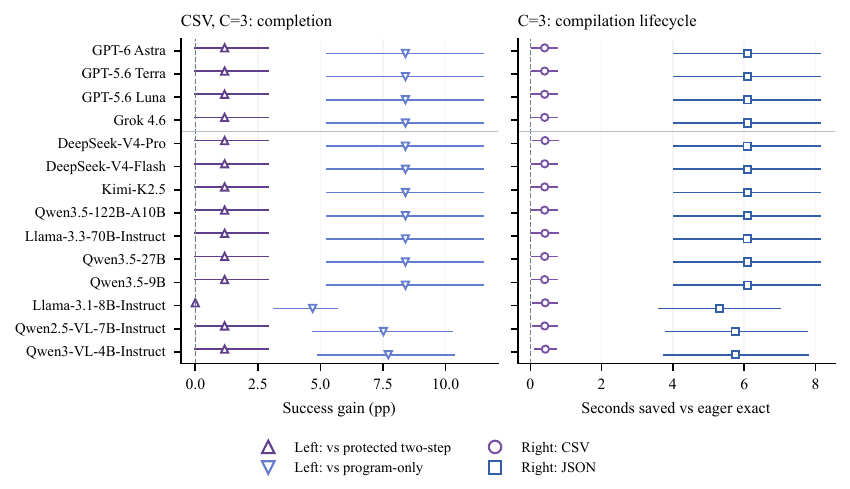}
\caption{\textbf{Proposal sources change completion and compiler inputs.}
Provider-conditioned comparisons at $C=3$.
Positive values favor CAPE.
Completion includes all proposals; native costs condition on executable
ones. Intervals resample sources, with shared physical measurements
retained inside their source clusters.}
\label{fig:panel-three}
\end{figure}
\begin{figure}[H]
\centering
\includegraphics[width=.87\linewidth]{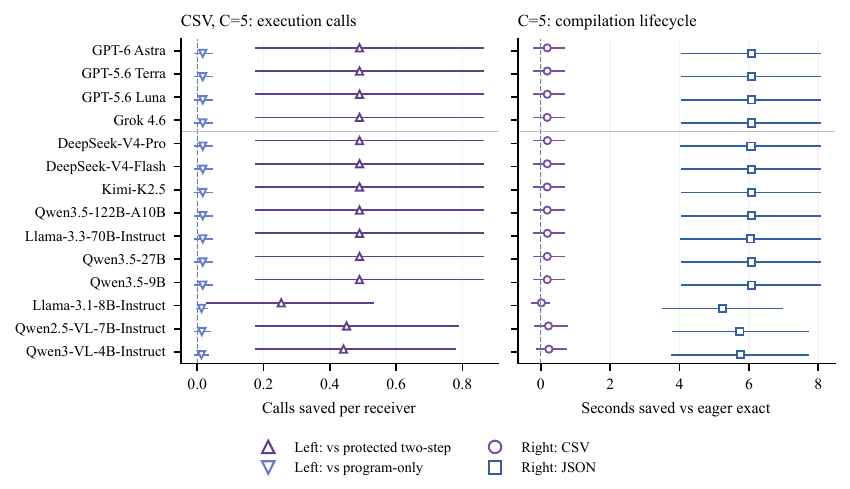}
\caption{\textbf{CAPE reduces CSV execution calls.}
Provider-conditioned comparisons at $C=5$, where CAPE and protected
two-step have equal completion.
The complete tables give the completion and native costs of every
method.}
\label{fig:panel-five}
\end{figure}
\FloatBarrier

\subsection{Complete results by distinct input group}

Nine deployments produce the same executable input for every request:
GPT-6 Astra, GPT-5.6 Terra, GPT-5.6 Luna, Grok 4.6,
DeepSeek-V4-Flash, Kimi-K2.5, Qwen3.5-122B-A10B,
Qwen3.5-27B and Qwen3.5-9B.
Their complete native results occupy one table.
Each of the five remaining deployments has a separate table.
Success and certificate rates include all proposals; calls, queries,
lifecycle and independently attained optima condition on executable
proposals.
The original per-deployment numerical sources are retained.

\begin{table}[H]
\centering\small\setlength{\tabcolsep}{3pt}
\caption{Shared native measurements for the nine deployments with identical inputs.}
\fitinput{\begin{tabular}{llrrrrrr}
\toprule
Interface, $C$ & Method & Success (\%) & Calls & Queries & Time (s) & Cert. (\%) & Opt. (\%) \\
\midrule
CSV, 3 & CAPE & 86.72 & 2.188 & 2948.4 & 10.655 & 100.00 & 100.00 \\
CSV, 3 & Eager exact & 86.72 & 2.188 & 3412.0 & 11.056 & 100.00 & 100.00 \\
CSV, 3 & Program-only & 78.32 & 2.016 & 624.0 & 3.611 & 35.94 & 35.94 \\
CSV, 3 & Protected two-step & 85.55 & 2.461 & 3412.0 & 11.004 & 35.94 & 35.94 \\
CSV, 3 & Generic lazy & 85.21 & 2.188 & 3293.3 & 16.868 & 94.53 & 94.53 \\
CSV, 3 & Program-first expansion & 86.72 & 2.188 & 3268.5 & 11.268 & 100.00 & 100.00 \\
CSV, 3 & Full-row acquisition & 86.72 & 2.188 & 3204.5 & 11.131 & 100.00 & 100.00 \\
CSV, 3 & No tight bounds & 86.72 & 2.188 & 3291.0 & 11.191 & 100.00 & 100.00 \\
CSV, 5 & CAPE & 100.00 & 2.338 & 3124.6 & 11.060 & 100.00 & 100.00 \\
CSV, 5 & Eager exact & 100.00 & 2.338 & 3412.0 & 11.246 & 100.00 & 100.00 \\
CSV, 5 & Program-only & 93.75 & 2.354 & 624.0 & 3.618 & 35.94 & 75.00 \\
CSV, 5 & Protected two-step & 100.00 & 2.828 & 3412.0 & 11.051 & 35.94 & 70.31 \\
CSV, 5 & Generic lazy & 85.74 & 2.524 & 2363.8 & 13.618 & 75.00 & 75.00 \\
CSV, 5 & Program-first expansion & 100.00 & 2.338 & 3155.6 & 11.361 & 100.00 & 100.00 \\
CSV, 5 & Full-row acquisition & 100.00 & 2.338 & 3124.4 & 11.019 & 100.00 & 100.00 \\
CSV, 5 & No tight bounds & 100.00 & 2.338 & 3124.6 & 11.062 & 100.00 & 100.00 \\
JSON, 3 & CAPE & 96.31 & 1.475 & 1502.5 & 11.800 & 96.88 & 98.44 \\
JSON, 3 & Eager exact & 96.16 & 1.473 & 2319.9 & 17.889 & 85.16 & 98.44 \\
JSON, 3 & Program-only & 96.16 & 1.473 & 2079.5 & 17.639 & 78.91 & 98.44 \\
JSON, 3 & Protected two-step & 96.09 & 1.491 & 2311.7 & 17.910 & 78.91 & 91.41 \\
JSON, 3 & Generic lazy & 94.46 & 1.494 & 2026.2 & 16.010 & 95.31 & 95.31 \\
JSON, 3 & Program-first expansion & 96.16 & 1.473 & 2165.1 & 17.764 & 85.94 & 98.44 \\
JSON, 3 & Full-row acquisition & 92.05 & 1.428 & 2363.2 & 17.851 & 88.28 & 88.28 \\
JSON, 3 & No tight bounds & 96.31 & 1.475 & 1510.9 & 11.845 & 96.88 & 98.44 \\
JSON, 5 & CAPE & 96.31 & 1.474 & 1489.0 & 11.837 & 99.22 & 100.00 \\
JSON, 5 & Eager exact & 96.16 & 1.473 & 2316.4 & 17.904 & 85.16 & 98.44 \\
JSON, 5 & Program-only & 96.16 & 1.473 & 2081.3 & 17.633 & 78.12 & 98.44 \\
JSON, 5 & Protected two-step & 96.09 & 1.491 & 2311.7 & 17.899 & 79.69 & 91.41 \\
JSON, 5 & Generic lazy & 94.46 & 1.545 & 1952.8 & 15.787 & 95.31 & 95.31 \\
JSON, 5 & Program-first expansion & 96.16 & 1.473 & 2174.6 & 17.769 & 86.72 & 98.44 \\
JSON, 5 & Full-row acquisition & 92.61 & 1.434 & 2044.8 & 16.916 & 88.28 & 89.84 \\
JSON, 5 & No tight bounds & 96.31 & 1.474 & 1493.0 & 11.803 & 100.00 & 100.00 \\
\bottomrule
\end{tabular}
}
\end{table}
\clearpage

\begin{table}[H]
\centering\small\setlength{\tabcolsep}{3pt}
\caption{DeepSeek-V4-Pro: complete fresh panel measurements.}
\fitinput{\begin{tabular}{llrrrrrr}
\toprule
Interface, $C$ & Method & Success (\%) & Calls & Queries & Time (s) & Cert. (\%) & Opt. (\%) \\
\midrule
CSV, 3 & CAPE & 85.16 & 2.196 & 2980.5 & 10.749 & 98.44 & 100.00 \\
CSV, 3 & Eager exact & 85.16 & 2.196 & 3450.9 & 11.159 & 98.44 & 100.00 \\
CSV, 3 & Program-only & 76.76 & 2.022 & 624.0 & 3.614 & 34.38 & 35.04 \\
CSV, 3 & Protected two-step & 83.98 & 2.474 & 3450.9 & 11.110 & 34.38 & 35.04 \\
CSV, 3 & Generic lazy & 83.64 & 2.197 & 3332.7 & 17.040 & 92.97 & 94.53 \\
CSV, 3 & Program-first expansion & 85.16 & 2.196 & 3307.3 & 11.380 & 98.44 & 100.00 \\
CSV, 3 & Full-row acquisition & 85.16 & 2.196 & 3243.6 & 11.238 & 98.44 & 100.00 \\
CSV, 3 & No tight bounds & 85.16 & 2.196 & 3330.3 & 11.302 & 98.44 & 100.00 \\
CSV, 5 & CAPE & 98.44 & 2.348 & 3161.1 & 11.160 & 98.44 & 100.00 \\
CSV, 5 & Eager exact & 98.44 & 2.348 & 3450.9 & 11.350 & 98.44 & 100.00 \\
CSV, 5 & Program-only & 92.19 & 2.365 & 624.0 & 3.619 & 34.38 & 75.00 \\
CSV, 5 & Protected two-step & 98.44 & 2.838 & 3450.9 & 11.155 & 34.38 & 70.31 \\
CSV, 5 & Generic lazy & 84.18 & 2.534 & 2384.9 & 13.702 & 73.44 & 75.00 \\
CSV, 5 & Program-first expansion & 98.44 & 2.348 & 3191.6 & 11.463 & 98.44 & 100.00 \\
CSV, 5 & Full-row acquisition & 98.44 & 2.348 & 3161.0 & 11.122 & 98.44 & 100.00 \\
CSV, 5 & No tight bounds & 98.44 & 2.348 & 3161.1 & 11.164 & 98.44 & 100.00 \\
JSON, 3 & CAPE & 94.74 & 1.476 & 1507.5 & 11.815 & 95.31 & 98.44 \\
JSON, 3 & Eager exact & 94.60 & 1.474 & 2326.9 & 17.899 & 83.59 & 98.44 \\
JSON, 3 & Program-only & 94.60 & 1.474 & 2079.5 & 17.642 & 77.34 & 98.44 \\
JSON, 3 & Protected two-step & 94.53 & 1.492 & 2318.8 & 17.922 & 77.34 & 91.18 \\
JSON, 3 & Generic lazy & 92.90 & 1.495 & 2031.7 & 16.026 & 93.75 & 95.31 \\
JSON, 3 & Program-first expansion & 94.60 & 1.474 & 2169.4 & 17.769 & 84.38 & 98.44 \\
JSON, 3 & Full-row acquisition & 90.48 & 1.429 & 2367.0 & 17.859 & 86.72 & 88.28 \\
JSON, 3 & No tight bounds & 94.74 & 1.476 & 1515.9 & 11.863 & 95.31 & 98.44 \\
JSON, 5 & CAPE & 94.74 & 1.475 & 1494.8 & 11.854 & 97.66 & 100.00 \\
JSON, 5 & Eager exact & 94.60 & 1.474 & 2323.5 & 17.911 & 83.59 & 98.44 \\
JSON, 5 & Program-only & 94.60 & 1.474 & 2081.3 & 17.636 & 76.56 & 98.44 \\
JSON, 5 & Protected two-step & 94.53 & 1.492 & 2318.8 & 17.910 & 78.12 & 91.18 \\
JSON, 5 & Generic lazy & 92.90 & 1.546 & 1958.4 & 15.805 & 93.75 & 95.31 \\
JSON, 5 & Program-first expansion & 94.60 & 1.474 & 2179.3 & 17.779 & 85.16 & 98.44 \\
JSON, 5 & Full-row acquisition & 91.05 & 1.435 & 2052.7 & 16.929 & 86.72 & 89.84 \\
JSON, 5 & No tight bounds & 94.74 & 1.475 & 1498.8 & 11.811 & 98.44 & 100.00 \\
\bottomrule
\end{tabular}
}
\end{table}
\clearpage

\begin{table}[H]
\centering\small\setlength{\tabcolsep}{3pt}
\caption{Llama-3.3-70B-Instruct: complete panel measurements.}
\fitinput{\begin{tabular}{llrrrrrr}
\toprule
Interface, $C$ & Method & Success (\%) & Calls & Queries & Time (s) & Cert. (\%) & Opt. (\%) \\
\midrule
CSV, 3 & CAPE & 83.59 & 2.196 & 2980.5 & 10.750 & 96.88 & 100.00 \\
CSV, 3 & Eager exact & 83.59 & 2.196 & 3450.9 & 11.158 & 96.88 & 100.00 \\
CSV, 3 & Program-only & 75.20 & 2.022 & 624.0 & 3.614 & 32.81 & 35.04 \\
CSV, 3 & Protected two-step & 82.42 & 2.474 & 3450.9 & 11.111 & 32.81 & 35.04 \\
CSV, 3 & Generic lazy & 82.08 & 2.197 & 3332.7 & 17.040 & 91.41 & 94.53 \\
CSV, 3 & Program-first expansion & 83.59 & 2.196 & 3307.3 & 11.382 & 96.88 & 100.00 \\
CSV, 3 & Full-row acquisition & 83.59 & 2.196 & 3243.6 & 11.238 & 96.88 & 100.00 \\
CSV, 3 & No tight bounds & 83.59 & 2.196 & 3330.3 & 11.304 & 96.88 & 100.00 \\
CSV, 5 & CAPE & 96.88 & 2.348 & 3161.1 & 11.160 & 96.88 & 100.00 \\
CSV, 5 & Eager exact & 96.88 & 2.348 & 3450.9 & 11.350 & 96.88 & 100.00 \\
CSV, 5 & Program-only & 90.62 & 2.365 & 624.0 & 3.620 & 32.81 & 75.00 \\
CSV, 5 & Protected two-step & 96.88 & 2.838 & 3450.9 & 11.154 & 32.81 & 70.31 \\
CSV, 5 & Generic lazy & 82.62 & 2.534 & 2384.9 & 13.702 & 71.88 & 75.00 \\
CSV, 5 & Program-first expansion & 96.88 & 2.348 & 3191.6 & 11.463 & 96.88 & 100.00 \\
CSV, 5 & Full-row acquisition & 96.88 & 2.348 & 3161.0 & 11.123 & 96.88 & 100.00 \\
CSV, 5 & No tight bounds & 96.88 & 2.348 & 3161.1 & 11.165 & 96.88 & 100.00 \\
JSON, 3 & CAPE & 93.18 & 1.476 & 1507.2 & 11.812 & 93.75 & 98.44 \\
JSON, 3 & Eager exact & 93.04 & 1.474 & 2326.9 & 17.896 & 82.03 & 98.44 \\
JSON, 3 & Program-only & 93.04 & 1.474 & 2079.5 & 17.641 & 75.78 & 98.44 \\
JSON, 3 & Protected two-step & 92.97 & 1.492 & 2318.8 & 17.919 & 75.78 & 91.18 \\
JSON, 3 & Generic lazy & 91.34 & 1.495 & 2031.7 & 16.024 & 92.19 & 95.31 \\
JSON, 3 & Program-first expansion & 93.04 & 1.474 & 2169.4 & 17.769 & 82.81 & 98.44 \\
JSON, 3 & Full-row acquisition & 88.92 & 1.429 & 2367.0 & 17.856 & 85.16 & 88.28 \\
JSON, 3 & No tight bounds & 93.18 & 1.476 & 1515.6 & 11.860 & 93.75 & 98.44 \\
JSON, 5 & CAPE & 93.18 & 1.475 & 1494.7 & 11.855 & 96.09 & 100.00 \\
JSON, 5 & Eager exact & 93.04 & 1.474 & 2323.5 & 17.907 & 82.03 & 98.44 \\
JSON, 5 & Program-only & 93.04 & 1.474 & 2081.3 & 17.635 & 75.00 & 98.44 \\
JSON, 5 & Protected two-step & 92.97 & 1.492 & 2318.8 & 17.908 & 76.56 & 91.18 \\
JSON, 5 & Generic lazy & 91.34 & 1.546 & 1958.4 & 15.802 & 92.19 & 95.31 \\
JSON, 5 & Program-first expansion & 93.04 & 1.474 & 2179.3 & 17.776 & 83.59 & 98.44 \\
JSON, 5 & Full-row acquisition & 89.49 & 1.435 & 2052.7 & 16.930 & 85.16 & 89.84 \\
JSON, 5 & No tight bounds & 93.18 & 1.475 & 1498.8 & 11.812 & 96.88 & 100.00 \\
\bottomrule
\end{tabular}
}
\end{table}
\clearpage

\begin{table}[H]
\centering\small\setlength{\tabcolsep}{3pt}
\caption{Llama-3.1-8B-Instruct: complete panel measurements.}
\fitinput{\begin{tabular}{llrrrrrr}
\toprule
Interface, $C$ & Method & Success (\%) & Calls & Queries & Time (s) & Cert. (\%) & Opt. (\%) \\
\midrule
CSV, 3 & CAPE & 57.81 & 1.798 & 2327.5 & 8.728 & 73.44 & 100.00 \\
CSV, 3 & Eager exact & 57.81 & 1.798 & 2695.7 & 9.144 & 73.44 & 100.00 \\
CSV, 3 & Program-only & 53.12 & 1.666 & 624.0 & 3.647 & 39.06 & 52.40 \\
CSV, 3 & Protected two-step & 57.81 & 2.001 & 2695.7 & 9.123 & 39.06 & 52.40 \\
CSV, 3 & Generic lazy & 56.84 & 1.801 & 2581.0 & 13.167 & 70.31 & 96.43 \\
CSV, 3 & Program-first expansion & 57.81 & 1.798 & 2680.4 & 9.596 & 73.44 & 100.00 \\
CSV, 3 & Full-row acquisition & 57.81 & 1.798 & 2609.7 & 9.380 & 73.44 & 100.00 \\
CSV, 3 & No tight bounds & 57.81 & 1.798 & 2645.5 & 9.405 & 73.44 & 100.00 \\
CSV, 5 & CAPE & 64.06 & 1.894 & 2547.4 & 9.265 & 73.44 & 100.00 \\
CSV, 5 & Eager exact & 64.06 & 1.894 & 2695.7 & 9.282 & 73.44 & 100.00 \\
CSV, 5 & Program-only & 61.52 & 1.906 & 624.0 & 3.651 & 39.06 & 86.01 \\
CSV, 5 & Protected two-step & 64.06 & 2.147 & 2695.7 & 9.156 & 39.06 & 84.23 \\
CSV, 5 & Generic lazy & 59.18 & 2.017 & 1875.8 & 10.467 & 65.62 & 86.01 \\
CSV, 5 & Program-first expansion & 64.06 & 1.894 & 2581.4 & 9.526 & 73.44 & 100.00 \\
CSV, 5 & Full-row acquisition & 64.06 & 1.894 & 2547.3 & 9.272 & 73.44 & 100.00 \\
CSV, 5 & No tight bounds & 64.06 & 1.894 & 2547.4 & 9.264 & 73.44 & 100.00 \\
JSON, 3 & CAPE & 62.43 & 1.248 & 1550.0 & 12.329 & 73.44 & 100.00 \\
JSON, 3 & Eager exact & 62.29 & 1.247 & 2266.0 & 17.638 & 63.28 & 98.21 \\
JSON, 3 & Program-only & 62.29 & 1.247 & 2080.8 & 17.494 & 60.16 & 98.21 \\
JSON, 3 & Protected two-step & 62.29 & 1.268 & 2263.2 & 17.651 & 60.16 & 92.49 \\
JSON, 3 & Generic lazy & 61.72 & 1.256 & 2007.8 & 15.904 & 71.88 & 98.21 \\
JSON, 3 & Program-first expansion & 62.29 & 1.247 & 2112.3 & 17.492 & 63.28 & 98.21 \\
JSON, 3 & Full-row acquisition & 59.16 & 1.211 & 2268.1 & 17.657 & 64.84 & 90.18 \\
JSON, 3 & No tight bounds & 62.43 & 1.248 & 1551.0 & 12.390 & 73.44 & 100.00 \\
JSON, 5 & CAPE & 62.43 & 1.248 & 1507.7 & 12.402 & 73.44 & 100.00 \\
JSON, 5 & Eager exact & 62.29 & 1.247 & 2259.8 & 17.638 & 63.28 & 98.21 \\
JSON, 5 & Program-only & 62.29 & 1.247 & 2082.1 & 17.466 & 59.38 & 98.21 \\
JSON, 5 & Protected two-step & 62.29 & 1.268 & 2262.7 & 17.624 & 60.94 & 92.49 \\
JSON, 5 & Generic lazy & 61.72 & 1.273 & 1942.6 & 15.782 & 71.88 & 98.21 \\
JSON, 5 & Program-first expansion & 62.29 & 1.247 & 2115.2 & 17.496 & 64.84 & 98.21 \\
JSON, 5 & Full-row acquisition & 59.73 & 1.217 & 1986.2 & 16.801 & 66.41 & 91.96 \\
JSON, 5 & No tight bounds & 62.43 & 1.248 & 1507.7 & 12.422 & 73.44 & 100.00 \\
\bottomrule
\end{tabular}
}
\end{table}
\clearpage

\begin{table}[H]
\centering\small\setlength{\tabcolsep}{3pt}
\caption{Qwen2.5-VL-7B-Instruct: complete panel measurements.}
\fitinput{\begin{tabular}{llrrrrrr}
\toprule
Interface, $C$ & Method & Success (\%) & Calls & Queries & Time (s) & Cert. (\%) & Opt. (\%) \\
\midrule
CSV, 3 & CAPE & 80.47 & 2.060 & 2752.8 & 10.046 & 96.88 & 100.00 \\
CSV, 3 & Eager exact & 80.47 & 2.060 & 3188.6 & 10.449 & 96.88 & 100.00 \\
CSV, 3 & Program-only & 72.95 & 1.903 & 624.0 & 3.596 & 39.06 & 41.07 \\
CSV, 3 & Protected two-step & 79.30 & 2.314 & 3188.6 & 10.401 & 39.06 & 41.07 \\
CSV, 3 & Generic lazy & 79.39 & 2.061 & 3085.9 & 15.851 & 92.97 & 95.98 \\
CSV, 3 & Program-first expansion & 80.47 & 2.060 & 3045.0 & 10.612 & 96.88 & 100.00 \\
CSV, 3 & Full-row acquisition & 80.47 & 2.060 & 2981.7 & 10.486 & 96.88 & 100.00 \\
CSV, 3 & No tight bounds & 80.47 & 2.060 & 3067.1 & 10.549 & 96.88 & 100.00 \\
CSV, 5 & CAPE & 92.19 & 2.199 & 2909.1 & 10.430 & 96.88 & 100.00 \\
CSV, 5 & Eager exact & 92.19 & 2.199 & 3188.6 & 10.649 & 96.88 & 100.00 \\
CSV, 5 & Program-only & 86.82 & 2.213 & 624.0 & 3.613 & 39.06 & 77.46 \\
CSV, 5 & Protected two-step & 92.19 & 2.650 & 3188.6 & 10.452 & 39.06 & 72.77 \\
CSV, 5 & Generic lazy & 79.98 & 2.360 & 2229.5 & 12.839 & 75.00 & 77.46 \\
CSV, 5 & Program-first expansion & 92.19 & 2.199 & 2940.5 & 10.665 & 96.88 & 100.00 \\
CSV, 5 & Full-row acquisition & 92.19 & 2.199 & 2908.9 & 10.388 & 96.88 & 100.00 \\
CSV, 5 & No tight bounds & 92.19 & 2.199 & 2909.1 & 10.434 & 96.88 & 100.00 \\
JSON, 3 & CAPE & 88.78 & 1.412 & 1538.7 & 12.130 & 92.19 & 98.21 \\
JSON, 3 & Eager exact & 88.64 & 1.409 & 2324.0 & 17.879 & 82.03 & 98.44 \\
JSON, 3 & Program-only & 88.64 & 1.409 & 2080.2 & 17.635 & 75.78 & 98.44 \\
JSON, 3 & Protected two-step & 88.64 & 1.428 & 2317.9 & 17.896 & 75.78 & 92.19 \\
JSON, 3 & Generic lazy & 86.93 & 1.479 & 2025.2 & 16.060 & 90.62 & 93.30 \\
JSON, 3 & Program-first expansion & 88.64 & 1.409 & 2170.2 & 17.749 & 82.81 & 98.44 \\
JSON, 3 & Full-row acquisition & 85.09 & 1.369 & 2360.2 & 17.842 & 85.16 & 89.62 \\
JSON, 3 & No tight bounds & 88.78 & 1.412 & 1547.2 & 12.175 & 92.19 & 98.21 \\
JSON, 5 & CAPE & 88.78 & 1.411 & 1530.6 & 12.158 & 94.53 & 100.00 \\
JSON, 5 & Eager exact & 88.64 & 1.409 & 2321.7 & 17.895 & 82.03 & 98.44 \\
JSON, 5 & Program-only & 88.64 & 1.409 & 2081.6 & 17.616 & 75.00 & 98.44 \\
JSON, 5 & Protected two-step & 88.64 & 1.428 & 2318.0 & 17.893 & 76.56 & 92.19 \\
JSON, 5 & Generic lazy & 86.93 & 1.567 & 1957.1 & 15.878 & 90.62 & 93.30 \\
JSON, 5 & Program-first expansion & 88.64 & 1.409 & 2179.2 & 17.749 & 83.59 & 98.44 \\
JSON, 5 & Full-row acquisition & 85.65 & 1.375 & 2053.1 & 16.948 & 85.16 & 91.18 \\
JSON, 5 & No tight bounds & 88.78 & 1.411 & 1534.9 & 12.131 & 95.31 & 100.00 \\
\bottomrule
\end{tabular}
}
\end{table}
\clearpage

\begin{table}[H]
\centering\small\setlength{\tabcolsep}{3pt}
\caption{Qwen3-VL-4B-Instruct: complete panel measurements.}
\fitinput{\begin{tabular}{llrrrrrr}
\toprule
Interface, $C$ & Method & Success (\%) & Calls & Queries & Time (s) & Cert. (\%) & Opt. (\%) \\
\midrule
CSV, 3 & CAPE & 83.40 & 2.062 & 2763.2 & 10.062 & 100.00 & 100.00 \\
CSV, 3 & Eager exact & 83.40 & 2.062 & 3208.0 & 10.484 & 100.00 & 100.00 \\
CSV, 3 & Program-only & 75.68 & 1.904 & 624.0 & 3.583 & 40.62 & 40.62 \\
CSV, 3 & Protected two-step & 82.23 & 2.320 & 3208.0 & 10.436 & 40.62 & 40.62 \\
CSV, 3 & Generic lazy & 82.32 & 2.063 & 3106.0 & 15.878 & 96.09 & 96.09 \\
CSV, 3 & Program-first expansion & 83.40 & 2.062 & 3064.5 & 10.650 & 100.00 & 100.00 \\
CSV, 3 & Full-row acquisition & 83.40 & 2.062 & 3015.0 & 10.545 & 100.00 & 100.00 \\
CSV, 3 & No tight bounds & 83.40 & 2.062 & 3087.0 & 10.580 & 100.00 & 100.00 \\
CSV, 5 & CAPE & 95.31 & 2.197 & 2925.5 & 10.426 & 100.00 & 100.00 \\
CSV, 5 & Eager exact & 95.31 & 2.197 & 3208.0 & 10.661 & 100.00 & 100.00 \\
CSV, 5 & Program-only & 89.94 & 2.210 & 624.0 & 3.594 & 40.62 & 78.12 \\
CSV, 5 & Protected two-step & 95.31 & 2.639 & 3208.0 & 10.502 & 40.62 & 73.44 \\
CSV, 5 & Generic lazy & 83.11 & 2.354 & 2228.0 & 12.745 & 78.12 & 78.12 \\
CSV, 5 & Program-first expansion & 95.31 & 2.197 & 2956.5 & 10.663 & 100.00 & 100.00 \\
CSV, 5 & Full-row acquisition & 95.31 & 2.197 & 2925.3 & 10.398 & 100.00 & 100.00 \\
CSV, 5 & No tight bounds & 95.31 & 2.197 & 2925.5 & 10.432 & 100.00 & 100.00 \\
JSON, 3 & CAPE & 92.33 & 1.407 & 1527.4 & 12.063 & 95.31 & 98.44 \\
JSON, 3 & Eager exact & 92.19 & 1.405 & 2271.4 & 17.818 & 85.16 & 98.44 \\
JSON, 3 & Program-only & 92.19 & 1.405 & 2081.0 & 17.645 & 80.47 & 98.44 \\
JSON, 3 & Protected two-step & 92.19 & 1.416 & 2266.3 & 17.875 & 80.47 & 93.75 \\
JSON, 3 & Generic lazy & 90.48 & 1.473 & 2024.2 & 16.015 & 93.75 & 93.75 \\
JSON, 3 & Program-first expansion & 92.19 & 1.405 & 2168.5 & 17.766 & 85.94 & 98.44 \\
JSON, 3 & Full-row acquisition & 88.64 & 1.366 & 2355.0 & 17.840 & 88.28 & 89.84 \\
JSON, 3 & No tight bounds & 92.33 & 1.407 & 1535.3 & 12.101 & 95.31 & 98.44 \\
JSON, 5 & CAPE & 92.33 & 1.406 & 1514.3 & 12.072 & 97.66 & 100.00 \\
JSON, 5 & Eager exact & 92.19 & 1.405 & 2267.9 & 17.828 & 85.16 & 98.44 \\
JSON, 5 & Program-only & 92.19 & 1.405 & 2083.1 & 17.621 & 79.69 & 98.44 \\
JSON, 5 & Protected two-step & 92.19 & 1.416 & 2265.1 & 17.829 & 81.25 & 93.75 \\
JSON, 5 & Generic lazy & 90.48 & 1.555 & 1952.7 & 15.836 & 93.75 & 93.75 \\
JSON, 5 & Program-first expansion & 92.19 & 1.405 & 2175.8 & 17.761 & 86.72 & 98.44 \\
JSON, 5 & Full-row acquisition & 89.20 & 1.371 & 2046.7 & 16.926 & 88.28 & 91.41 \\
JSON, 5 & No tight bounds & 92.33 & 1.406 & 1518.4 & 12.058 & 98.44 & 100.00 \\
\bottomrule
\end{tabular}
}
\end{table}
\clearpage

\section{Execution and scaling studies}
\label{app:execution-scaling}

\subsection{Execution conditions and feedback controllers}

The execution study uses the 64 retained requests and DeepSeek-V4-Pro
proposals, CSV/JSON interfaces, and $C=3/5$, giving 256 conditions.
The four conditions with the invalid proposal contribute zero completion;
\SVal{e1/valid} conditions have native execution measurements.
CAPE is freshly compiled once per condition.
Its policy attains the independent optimum in
\SVal{e1/optimal}/\SVal{e1/valid} executable conditions and preserves
the successful configurations of the catalog-order initial policy
throughout.
These measurements use Python 3.12.3, pandas 3.0.2, DuckDB 1.5.5,
fsspec 2026.9.0 and pytz 2026.6; timings are kept separate from the
original mechanism, matched and proposal studies.

Catalog retry executes the shared initial policy.
LLM serialization retry starts with program representation zero and,
after failure, selects an untried program representation.
A single remaining representation is executed directly.
The operation is fixed, and each program consumes one tool call.
The controller receives the request, proposed operation, schema, first
five records, available program encodings and its own response history.
CSV reports structured parse/query errors; JSON reports bounded
exception messages. The publication guard and native program responses
are identical across methods.
The DeepSeek-V4-Pro deployment dated 2026-04-23 uses temperature zero,
JSON output and an 8,192-token completion allowance.
Completed decisions are retained without semantic
resampling; identical histories share that response.
The study made \SVal{e1/api_calls} completed model requests, with
\SVal{e1/tokens} observed input and completion tokens.

\begin{table}[H]
\centering\small\setlength{\tabcolsep}{4pt}
\caption{Execution-study paired contrasts with 95\% source-cluster intervals.
Completion gain is CAPE minus the control; calls and time saved are the
control minus CAPE. Costs include the measured common preparation and
method-specific compilation components.}
\label{tab:execution-contrasts}
\fitinput{\begin{tabular}{lllll}
\toprule
Interface/$C$ & Control & Completion gain (pp) & Calls saved & Cold total saved (s) \\
\midrule
CSV/3 & Catalog retry & 15.19 [7.76, 24.61] & -0.084 [-0.157, 0.007] & -9.047 [-11.626, -6.045] \\
CSV/3 & LLM retry & 15.38 [7.81, 25.00] & -0.084 [-0.157, 0.007] & -6.203 [-8.300, -3.899] \\
CSV/5 & Catalog retry & 15.43 [5.47, 27.15] & 0.257 [0.052, 0.508] & -9.486 [-11.817, -6.435] \\
CSV/5 & LLM retry & 16.11 [6.64, 26.66] & 0.260 [0.059, 0.502] & -6.901 [-8.559, -4.738] \\
JSON/3 & Catalog retry & 34.66 [27.56, 41.76] & 0.481 [0.457, 0.521] & -11.229 [-15.785, -6.630] \\
JSON/3 & LLM retry & 34.66 [27.56, 41.76] & 0.481 [0.457, 0.521] & -9.909 [-14.260, -5.581] \\
JSON/5 & Catalog retry & 26.85 [13.92, 39.77] & 1.256 [1.140, 1.361] & -11.188 [-15.722, -6.631] \\
JSON/5 & LLM retry & 21.95 [12.86, 30.89] & 1.206 [1.118, 1.289] & -7.930 [-12.987, -2.886] \\
\bottomrule
\end{tabular}
}
\end{table}

An initial feedback controller also had access to the full program and
probe catalogs. Its bound trajectories are reported in
Table~\ref{tab:feedback-diagnostic}.
After observing its probe usage, we specified the program-first
serialization controller above and froze its protocol before collecting
its responses on the same workload.
Both controllers retain their complete native and model records.

\begin{table}[H]
\centering\small
\caption{Full-catalog feedback-controller diagnostic.
Probe and program counts sum the complete configuration-family
trajectories; identical model contexts share a response.
Lost is the fraction of the initial policy's successful configurations
lost by the controller, averaged within sources.}
\label{tab:feedback-diagnostic}
\begin{tabular}{lrrrrr}
\toprule
Interface/$C$ & Success (\%) & Calls & Lost (\%) & Probe calls & Program calls \\
\midrule
CSV/3 & 15.53 & 2.984 & 81.08 & 9144 & 2888 \\
CSV/5 & 15.58 & 4.753 & 84.08 & 17547 & 1603 \\
JSON/3 & 54.30 & 2.680 & 20.15 & 10044 & 4828 \\
JSON/5 & 46.24 & 4.057 & 39.48 & 17779 & 4701 \\
\bottomrule
\end{tabular}

\end{table}

\subsection{Cold execution and reuse accounting}

A fixed hash selects one endpoint configuration per condition before
execution. It is executed first with an empty decision cache.
We retime CAPE and catalog retry at that same endpoint alongside the
program-first controller, then run repetitions immediately after each
method's cold episode. All methods cache a successfully committed
program representation. Task data, operation, catalog and endpoint
configuration remain fixed for $N=1,8,32$.
The LLM controller also retains its exact-history decision cache.
Calls and elapsed times include failed paths.

Total cost combines the common measured catalog preparation, additional
method-specific compilation and measured execution:
\begin{equation}
 T_{\mathrm{total}}(N)=T_{\mathrm{setup}}+T_{\mathrm{compile}}^{\mathrm{extra}}
 +\sum_{j=1}^{N}T_{\mathrm{execution},j}.
 \label{eq:reuse}
\end{equation}
For CAPE, setup plus additional compilation equals its measured
lifecycle; the retry methods pay the common setup once.
The shared initial proposal is reused.
Adjusted time subtracts client authentication, initialization, rate
pacing and concurrency-slot waiting, retaining model service, tool
execution, controller work and transport retries.
The complete-family quality values and compilation components come from
the verified policy measurements; cold and repeated execution are newly
timed for this comparison.

\begin{table}[H]
\centering\small\setlength{\tabcolsep}{4pt}
\caption{Initial coverage, cold execution and model cost.
Lost is initial-coverage loss (\%); Cold is selected-endpoint completion
(\%) over all requests. Exec.\ is measured execution time; Total includes
preparation and compilation. Adjusted subtracts client waiting from Total.
Time and token means condition on valid proposals.}
\label{tab:execution-costs}
\fitinput{\begin{tabular}{llrrrrrr}
\toprule
Interface/$C$ & Method & Lost (\%) & Cold (\%) & Exec. (s) & Total (s) & Adjusted (s) & Tokens \\
\midrule
CSV/3 & Catalog retry & 0.00 & 64.06 & 0.121 & 1.146 & 1.146 & 0.0 \\
CSV/3 & LLM retry & 2.50 & 67.19 & 2.964 & 3.989 & 2.620 & 1720.2 \\
CSV/3 & CAPE & 0.00 & 82.81 & 0.126 & 10.192 & 10.192 & 0.0 \\
CSV/5 & Catalog retry & 0.00 & 82.81 & 0.106 & 1.191 & 1.191 & 0.0 \\
CSV/5 & LLM retry & 1.95 & 85.94 & 2.691 & 3.776 & 2.655 & 1910.8 \\
CSV/5 & CAPE & 0.00 & 98.44 & 0.135 & 10.677 & 10.677 & 0.0 \\
JSON/3 & Catalog retry & 0.00 & 78.12 & 0.103 & 0.472 & 0.472 & 0.0 \\
JSON/3 & LLM retry & 0.00 & 78.12 & 1.423 & 1.792 & 1.198 & 1039.4 \\
JSON/3 & CAPE & 0.00 & 96.88 & 0.112 & 11.701 & 11.701 & 0.0 \\
JSON/5 & Catalog retry & 0.00 & 68.75 & 0.099 & 0.522 & 0.522 & 0.0 \\
JSON/5 & LLM retry & 0.20 & 73.44 & 3.357 & 3.780 & 2.513 & 2540.7 \\
JSON/5 & CAPE & 0.00 & 90.62 & 0.119 & 11.710 & 11.710 & 0.0 \\
\bottomrule
\end{tabular}
}
\end{table}

\begin{table}[H]
\centering\small\setlength{\tabcolsep}{4pt}
\caption{Fixed-endpoint reuse over all valid proposals.
Success includes failed executions, calls are totals over $N$ executions,
and compilation is charged once. Adjusted time subtracts client waiting.
Means balance data sources.}
\label{tab:execution-reuse}
\begin{tabular}{lrlrrrr}
\toprule
Interface/$C$ & $N$ & Method & Success (\%) & Calls & Total (s) & Adjusted (s) \\
\midrule
CSV/3 & 1 & CAPE & 84.15 & 2.268 & 10.192 & 10.192 \\
CSV/3 & 1 & LLM retry & 68.53 & 2.281 & 3.989 & 2.620 \\
CSV/3 & 1 & Catalog retry & 65.40 & 2.281 & 1.146 & 1.146 \\
CSV/3 & 8 & CAPE & 84.15 & 11.487 & 10.916 & 10.916 \\
CSV/3 & 8 & LLM retry & 68.53 & 13.688 & 4.637 & 3.267 \\
CSV/3 & 8 & Catalog retry & 65.40 & 14.125 & 1.804 & 1.804 \\
CSV/3 & 32 & CAPE & 84.15 & 43.094 & 13.463 & 13.463 \\
CSV/3 & 32 & LLM retry & 68.53 & 52.795 & 7.143 & 5.774 \\
CSV/3 & 32 & Catalog retry & 65.40 & 54.732 & 4.028 & 4.028 \\
CSV/5 & 1 & CAPE & 100.00 & 2.248 & 10.677 & 10.677 \\
CSV/5 & 1 & LLM retry & 87.50 & 2.373 & 3.776 & 2.655 \\
CSV/5 & 1 & Catalog retry & 84.38 & 2.388 & 1.191 & 1.191 \\
CSV/5 & 8 & CAPE & 100.00 & 9.248 & 11.558 & 11.558 \\
CSV/5 & 8 & LLM retry & 87.50 & 12.873 & 4.490 & 3.369 \\
CSV/5 & 8 & Catalog retry & 84.38 & 13.763 & 1.946 & 1.946 \\
CSV/5 & 32 & CAPE & 100.00 & 33.248 & 14.485 & 14.485 \\
CSV/5 & 32 & LLM retry & 87.50 & 48.873 & 6.908 & 5.787 \\
CSV/5 & 32 & Catalog retry & 84.38 & 52.763 & 4.776 & 4.776 \\
JSON/3 & 1 & CAPE & 98.44 & 1.297 & 11.701 & 11.701 \\
JSON/3 & 1 & LLM retry & 79.69 & 1.625 & 1.792 & 1.198 \\
JSON/3 & 1 & Catalog retry & 79.69 & 1.625 & 0.472 & 0.472 \\
JSON/3 & 8 & CAPE & 98.44 & 8.297 & 12.463 & 12.463 \\
JSON/3 & 8 & LLM retry & 79.69 & 11.469 & 2.540 & 1.946 \\
JSON/3 & 8 & Catalog retry & 79.69 & 11.469 & 1.136 & 1.136 \\
JSON/3 & 32 & CAPE & 98.44 & 32.297 & 15.196 & 15.196 \\
JSON/3 & 32 & LLM retry & 79.69 & 45.219 & 5.032 & 4.439 \\
JSON/3 & 32 & Catalog retry & 79.69 & 45.219 & 3.497 & 3.497 \\
JSON/5 & 1 & CAPE & 91.96 & 1.429 & 11.710 & 11.710 \\
JSON/5 & 1 & LLM retry & 74.78 & 2.650 & 3.780 & 2.513 \\
JSON/5 & 1 & Catalog retry & 70.09 & 2.696 & 0.522 & 0.522 \\
JSON/5 & 8 & CAPE & 91.96 & 8.772 & 12.484 & 12.484 \\
JSON/5 & 8 & LLM retry & 74.78 & 16.712 & 4.500 & 3.233 \\
JSON/5 & 8 & Catalog retry & 70.09 & 18.071 & 1.123 & 1.123 \\
JSON/5 & 32 & CAPE & 91.96 & 33.951 & 15.081 & 15.081 \\
JSON/5 & 32 & LLM retry & 74.78 & 64.926 & 6.773 & 5.506 \\
JSON/5 & 32 & Catalog retry & 70.09 & 70.786 & 3.198 & 3.198 \\
\bottomrule
\end{tabular}

\end{table}

The common-success subset comprises \SVal{e1/common} conditions on which
all three methods commit the requested output in the cold episode.
Every subsequent cached execution also succeeds.
Table~\ref{tab:execution-common} reports their costs at equal completion,
with all eight sources represented in each interface/budget group.

\begin{table}[H]
\centering\small\setlength{\tabcolsep}{4pt}
\caption{Reuse costs on the common-success subset.
Cases counts conditions, shared across the three methods and all values
of $N$. Costs are source-balanced seconds with preparation and compilation
charged once. CAPE and catalog retry have no model-client waiting.}
\label{tab:execution-common}
\fitinput{\begin{tabular}{lrrrrrr}
\toprule
Interface/$C$ & $N$ & Cases & CAPE (s) & Catalog (s) & LLM (s) & LLM adjusted (s) \\
\midrule
CSV/3 & 1 & 41 & 8.396 & 1.225 & 3.066 & 2.369 \\
CSV/3 & 8 & 41 & 9.273 & 2.190 & 3.958 & 3.262 \\
CSV/3 & 32 & 41 & 12.316 & 5.386 & 7.581 & 6.884 \\
CSV/5 & 1 & 53 & 9.405 & 1.233 & 2.840 & 2.150 \\
CSV/5 & 8 & 53 & 10.304 & 2.116 & 3.620 & 2.930 \\
CSV/5 & 32 & 53 & 13.318 & 5.430 & 6.273 & 5.582 \\
JSON/3 & 1 & 50 & 12.047 & 0.486 & 0.926 & 0.770 \\
JSON/3 & 8 & 50 & 12.831 & 1.282 & 1.813 & 1.657 \\
JSON/3 & 32 & 50 & 15.608 & 4.153 & 4.846 & 4.690 \\
JSON/5 & 1 & 44 & 11.558 & 0.569 & 1.870 & 1.384 \\
JSON/5 & 8 & 44 & 12.350 & 1.366 & 2.737 & 2.252 \\
JSON/5 & 32 & 44 & 15.073 & 4.114 & 5.475 & 4.990 \\
\bottomrule
\end{tabular}
}
\end{table}

\subsection{Configuration scale, horizon and observation structure}

We select one projection request per source and evaluate five CSV
conditions: 64, 128 and 256 configurations at $C=5$; 64 configurations
at $C=8$; and 64 configurations at $C=5$ with merged probe feedback.
CAPE, eager exact and matched selective AND/OR each have two
fresh-process repetitions per condition, yielding 240 measurements.
The 30-second soft lifecycle allowance, native thread limits and
method definitions are shared with the original compiler studies.
The runtime versions match the execution study above.
Paired intervals use 5,000 source-cluster bootstrap resamples with seed
20261108 for both new studies.

The expanded CSV families use four and eight native date formats.
They retain the original 64 configurations and the effects of the
original action/configuration pairs.
The merged-feedback condition maps successful probe previews to
\texttt{\{parsed: true\}} while retaining error responses, program
effects and publication checks.
Each call budget uses its corresponding catalog-order initial policy
and required success set.
Table~\ref{tab:scaling-structure} gives effective configuration counts
and action-catalog sizes.

\begin{table}[H]
\centering\small\setlength{\tabcolsep}{4pt}
\caption{Complete structural and scaling results.
Success is completion, calls are per configuration, time is compilation
lifecycle, queries are acquired effects, and Cert.\ is the runtime
certificate rate. Each row aggregates eight sources and two repetitions.}
\label{tab:scaling-complete}
\fitinput{\begin{tabular}{llrrrrr}
\toprule
Setting & Method & Success (\%) & Calls & Time (s) & Queries & Cert. (\%) \\
\midrule
64, $C=5$ & Eager exact & 100.00 & 2.875 & 14.170 & 4432.0 & 100.00 \\
64, $C=5$ & Matched AND/OR & 100.00 & 2.875 & 14.874 & 4354.0 & 100.00 \\
64, $C=5$ & CAPE & 100.00 & 2.875 & 15.136 & 4354.0 & 100.00 \\
128, $C=5$ & Eager exact & 66.41 & 3.309 & 26.641 & 8644.3 & 12.50 \\
128, $C=5$ & Matched AND/OR & 74.22 & 3.379 & 26.740 & 7731.1 & 12.50 \\
128, $C=5$ & CAPE & 88.28 & 3.309 & 26.658 & 7559.1 & 12.50 \\
256, $C=5$ & Eager exact & 57.03 & 3.506 & 26.853 & 7745.2 & 12.50 \\
256, $C=5$ & Matched AND/OR & 58.59 & 3.619 & 26.904 & 7198.9 & 12.50 \\
256, $C=5$ & CAPE & 62.60 & 3.479 & 26.903 & 7117.4 & 12.50 \\
64, $C=8$ & Eager exact & 100.00 & 3.047 & 15.830 & 4432.0 & 87.50 \\
64, $C=8$ & Matched AND/OR & 100.00 & 2.875 & 16.123 & 4354.0 & 87.50 \\
64, $C=8$ & CAPE & 100.00 & 2.875 & 16.694 & 4354.0 & 87.50 \\
64, merged feedback & Eager exact & 99.22 & 2.883 & 14.146 & 4432.0 & 100.00 \\
64, merged feedback & Matched AND/OR & 99.22 & 2.883 & 14.982 & 4354.0 & 100.00 \\
64, merged feedback & CAPE & 99.22 & 2.883 & 15.174 & 4354.0 & 100.00 \\
\bottomrule
\end{tabular}
}
\end{table}

\begin{table}[H]
\centering\small\setlength{\tabcolsep}{4pt}
\caption{Scaling-study paired contrasts with 95\% source-cluster intervals.
Completion gain is CAPE minus the control; time and queries saved are
the control minus CAPE.}
\label{tab:scaling-contrasts}
\fitinput{\begin{tabular}{llllr}
\toprule
Setting & Control & Completion gain (pp) & Time saved (s) & Queries saved \\
\midrule
64, $C=5$ & Eager exact & 0.00 [0.00, 0.00] & -0.966 [-1.779, -0.135] & 78.0 \\
64, $C=5$ & Matched AND/OR & 0.00 [0.00, 0.00] & -0.262 [-0.837, 0.357] & 0.0 \\
128, $C=5$ & Eager exact & 21.88 [5.86, 39.45] & -0.017 [-0.036, 0.000] & 1085.2 \\
128, $C=5$ & Matched AND/OR & 14.06 [3.52, 25.00] & 0.082 [0.033, 0.132] & 172.0 \\
256, $C=5$ & Eager exact & 5.57 [1.95, 10.25] & -0.050 [-0.117, -0.012] & 627.8 \\
256, $C=5$ & Matched AND/OR & 4.00 [1.07, 6.93] & 0.001 [-0.019, 0.017] & 81.4 \\
64, $C=8$ & Eager exact & 0.00 [0.00, 0.00] & -0.864 [-1.990, 0.106] & 78.0 \\
64, $C=8$ & Matched AND/OR & 0.00 [0.00, 0.00] & -0.571 [-1.245, -0.009] & 0.0 \\
64, merged feedback & Eager exact & 0.00 [0.00, 0.00] & -1.029 [-1.877, -0.161] & 78.0 \\
64, merged feedback & Matched AND/OR & 0.00 [0.00, 0.00] & -0.192 [-0.630, 0.340] & 0.0 \\
\bottomrule
\end{tabular}
}
\end{table}

\begin{table}[H]
\centering\small\setlength{\tabcolsep}{4pt}
\caption{Structure of the five conditions.
Effective configurations have distinct complete action-effect signatures.
Program, probe and effective-configuration columns give ranges over
sources; Structures counts normalized action structures.}
\label{tab:scaling-structure}
\fitinput{\begin{tabular}{lrrrrr}
\toprule
Setting & Configs & Effective & Programs & Probes & Structures \\
\midrule
64, $C=5$ & 64 & 64--64 & 2--24 & 68--68 & 6 \\
128, $C=5$ & 128 & 128--128 & 2--48 & 100--100 & 6 \\
256, $C=5$ & 256 & 256--256 & 2--96 & 164--164 & 6 \\
64, $C=8$ & 64 & 64--64 & 2--24 & 68--68 & 6 \\
64, merged feedback & 64 & 32--64 & 2--24 & 68--68 & 6 \\
\bottomrule
\end{tabular}
}
\end{table}

\subsection{Compilation allowance}
\label{app:allowance}

The scaling study fixes the compilation allowance at 30 seconds.
We repeat its 64-, 128- and 256-configuration conditions at $C=5$ with
allowances from 5 to 240 seconds.
Requests, proposals, the three compilers, initial policies, the
measurement runner and the environment are those of the scaling study;
the allowance is the only setting that changes.
Each family and allowance has eight requests and two fresh-process
repetitions per method, \AVal{records} measurements in total.

\begin{figure}[H]
\centering
\includegraphics[width=.9\linewidth]{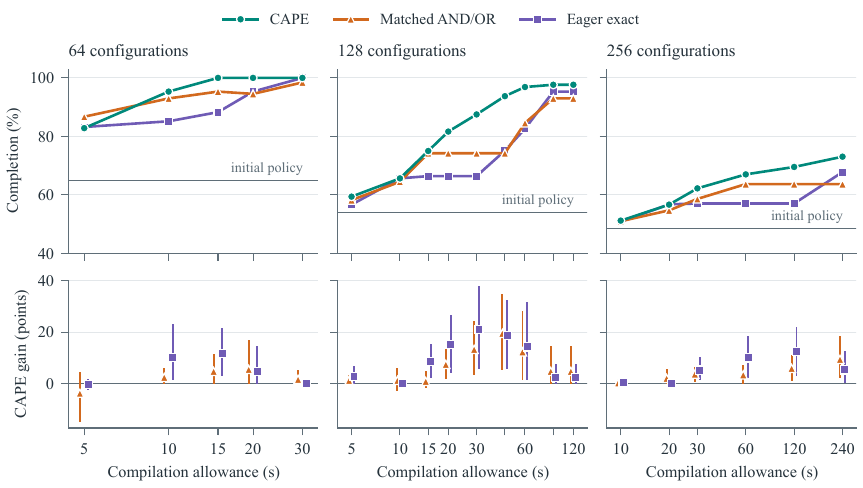}
\caption{\textbf{CAPE's lead is largest when the allowance covers part
of the calibration.}
Top: completion against the compilation allowance, as source-balanced
means; the horizontal line is the initial policy.
Bottom: paired completion gain of CAPE over each control, drawn in the
color of that control, with 95\% source-cluster intervals.}
\label{fig:allowance}
\end{figure}

At 128 configurations CAPE completes at least as many configurations as
both controls at every allowance (Fig.~\ref{fig:allowance},
Table~\ref{tab:allowance}).
Between 15 and 30 seconds eager exact is still acquiring effects and
stays at \AVal{dates-128/30/eager_optimized/completion}\%, and matched
AND/OR stays at \AVal{dates-128/30/lazy_andor_selective/completion}\%
from 15 to 45 seconds, whereas CAPE rises from
\AVal{dates-128/15/partial_native/completion}\% to
\AVal{dates-128/45/partial_native/completion}\%.
Its lead peaks at \AVal{dates-128/30/eager_optimized/gain} points over
eager exact at 30 seconds and at
\AVal{dates-128/45/lazy_andor_selective/gain} points over matched AND/OR
at 45 seconds.
The paired intervals exclude zero from 15 to 60 seconds against eager
exact and from 20 to 60 seconds against matched AND/OR.
At 90 and 120 seconds the leads are
\AVal{dates-128/120/eager_optimized/gain} and
\AVal{dates-128/120/lazy_andor_selective/gain} points and their
intervals reach zero.

The other two families show the same shape at other allowances.
At 64 configurations CAPE leads eager exact by
\AVal{anchor/10/eager_optimized/gain} and
\AVal{anchor/15/eager_optimized/gain} points at 10 and 15 seconds, and
CAPE and eager exact complete every configuration at 30 seconds.
At 256 configurations the lead over eager exact grows to
\AVal{dates-256/120/eager_optimized/gain} points at 120 seconds, and
all eight paired intervals from 30 to 240 seconds exclude zero.
CAPE is below a control in two cells.
At 64 configurations and 5 seconds matched AND/OR completes
\AVal{anchor/5/lazy_andor_selective/completion}\%, eager exact
\AVal{anchor/5/eager_optimized/completion}\% and CAPE
\AVal{anchor/5/partial_native/completion}\%, and at 256 configurations
and 20 seconds eager exact completes
\AVal{dates-256/20/eager_optimized/completion}\% and CAPE
\AVal{dates-256/20/partial_native/completion}\%; the three intervals
include zero.
The 30-second cells agree with the scaling study:
\AVal{dates-128/30/partial_native/completion}\%,
\AVal{dates-128/30/lazy_andor_selective/completion}\% and
\AVal{dates-128/30/eager_optimized/completion}\% at 128 configurations
against 88.28\%, 74.22\% and 66.41\% there.

\begin{table}[H]
\centering\small\setlength{\tabcolsep}{4pt}
\caption{Completion against the compilation allowance $\tau$.
Initial is the completion of the initial policy.
Gains are CAPE minus the control in percentage points, with 95\%
source-cluster intervals. Bold marks CAPE where no control completes
more.}
\label{tab:allowance}
\fitinput{\begin{tabular}{rrrrrrrr}
\toprule
& & & \multicolumn{3}{c}{Completion (\%)} & \multicolumn{2}{c}{CAPE gain [95\% interval]} \\
\cmidrule(lr){4-6}\cmidrule(l){7-8}
$|H|$ & $\tau$ (s) & Initial & Eager & AND/OR & CAPE & vs.\ eager & vs.\ AND/OR \\
\midrule
64 & 5 & 64.84 & 83.20 & 86.72 & 82.81 & $-0.39$ [$-2.34$, $1.17$] & $-3.91$ [$-14.84$, $3.91$] \\
 & 10 & 64.84 & 85.16 & 92.97 & \textbf{95.31} & $10.16$ [$1.56$, $22.66$] & $2.34$ [$0.00$, $5.47$] \\
 & 15 & 64.84 & 88.28 & 95.31 & \textbf{100.00} & $11.72$ [$3.12$, $21.09$] & $4.69$ [$0.00$, $10.94$] \\
 & 20 & 64.84 & 95.31 & 94.53 & \textbf{100.00} & $4.69$ [$0.00$, $14.06$] & $5.47$ [$0.00$, $16.41$] \\
 & 30 & 64.84 & 100.00 & 98.44 & \textbf{100.00} & $0.00$ [$0.00$, $0.00$] & $1.56$ [$0.00$, $4.69$] \\
\midrule
128 & 5 & 53.91 & 56.64 & 58.20 & \textbf{59.38} & $2.73$ [$0.00$, $6.25$] & $1.17$ [$0.00$, $2.73$] \\
 & 10 & 53.91 & 65.62 & 64.45 & \textbf{65.62} & $0.00$ [$0.00$, $0.00$] & $1.17$ [$-2.73$, $5.47$] \\
 & 15 & 53.91 & 66.41 & 74.22 & \textbf{75.00} & $8.59$ [$2.34$, $14.84$] & $0.78$ [$-1.56$, $4.30$] \\
 & 20 & 53.91 & 66.41 & 74.22 & \textbf{81.64} & $15.23$ [$4.30$, $26.17$] & $7.42$ [$1.95$, $12.89$] \\
 & 30 & 53.91 & 66.41 & 74.22 & \textbf{87.50} & $21.09$ [$5.86$, $37.50$] & $13.28$ [$3.52$, $23.83$] \\
 & 45 & 53.91 & 75.10 & 74.22 & \textbf{93.75} & $18.65$ [$6.15$, $32.13$] & $19.53$ [$5.47$, $34.38$] \\
 & 60 & 53.91 & 82.62 & 84.57 & \textbf{96.88} & $14.26$ [$1.76$, $31.45$] & $12.30$ [$1.76$, $27.93$] \\
 & 90 & 53.91 & 95.31 & 92.97 & \textbf{97.66} & $2.34$ [$0.00$, $7.03$] & $4.69$ [$0.00$, $14.06$] \\
 & 120 & 53.91 & 95.31 & 92.97 & \textbf{97.66} & $2.34$ [$0.00$, $7.03$] & $4.69$ [$0.00$, $14.06$] \\
\midrule
256 & 10 & 48.44 & 50.78 & 50.98 & \textbf{51.17} & $0.39$ [$0.00$, $0.98$] & $0.20$ [$0.00$, $0.49$] \\
 & 20 & 48.44 & 56.74 & 54.69 & 56.64 & $-0.10$ [$-0.29$, $0.00$] & $1.95$ [$0.00$, $5.08$] \\
 & 30 & 48.44 & 57.03 & 58.59 & \textbf{62.21} & $5.18$ [$1.56$, $9.87$] & $3.61$ [$0.98$, $6.15$] \\
 & 60 & 48.44 & 57.03 & 63.67 & \textbf{66.99} & $9.96$ [$2.34$, $17.97$] & $3.32$ [$0.20$, $6.84$] \\
 & 120 & 48.44 & 57.03 & 63.67 & \textbf{69.53} & $12.50$ [$3.32$, $21.48$] & $5.86$ [$1.37$, $10.55$] \\
 & 240 & 48.44 & 67.77 & 63.67 & \textbf{73.05} & $5.27$ [$0.39$, $12.30$] & $9.38$ [$2.34$, $17.97$] \\
\bottomrule
\end{tabular}
}
\end{table}

\begin{table}[H]
\centering\small\setlength{\tabcolsep}{4pt}
\caption{Execution calls, compilation time and certificates in the
allowance sweep. Calls are per configuration, time is the compilation
lifecycle in seconds, and Cert.\ is the runtime certificate rate (\%).}
\label{tab:allowance-costs}
\fitinput{\begin{tabular}{rrrrrrrrrrr}
\toprule
& & \multicolumn{3}{c}{Eager exact} & \multicolumn{3}{c}{Matched AND/OR} & \multicolumn{3}{c}{CAPE} \\
\cmidrule(lr){3-5}\cmidrule(lr){6-8}\cmidrule(l){9-11}
$|H|$ & $\tau$ (s) & Calls & Time & Cert. & Calls & Time & Cert. & Calls & Time & Cert. \\
\midrule
64 & 5 & 2.855 & 4.62 & 12 & 3.008 & 4.64 & 12 & 2.895 & 4.63 & 12 \\
 & 10 & 2.914 & 9.04 & 12 & 2.996 & 9.03 & 12 & 2.922 & 9.02 & 12 \\
 & 15 & 2.988 & 13.15 & 38 & 2.969 & 12.96 & 38 & 2.891 & 13.02 & 38 \\
 & 20 & 2.906 & 14.80 & 88 & 3.012 & 14.84 & 81 & 2.883 & 14.94 & 81 \\
 & 30 & 2.875 & 14.70 & 100 & 2.887 & 15.98 & 94 & 2.875 & 16.10 & 100 \\
\midrule
128 & 5 & 3.477 & 4.77 & 12 & 3.406 & 4.80 & 12 & 3.504 & 4.73 & 12 \\
 & 10 & 3.215 & 9.14 & 12 & 3.426 & 9.13 & 12 & 3.145 & 9.12 & 12 \\
 & 15 & 3.309 & 13.97 & 12 & 3.379 & 13.51 & 12 & 3.270 & 13.55 & 12 \\
 & 20 & 3.309 & 17.88 & 12 & 3.379 & 17.98 & 12 & 3.305 & 17.90 & 12 \\
 & 30 & 3.309 & 26.66 & 12 & 3.379 & 26.71 & 12 & 3.309 & 26.69 & 12 \\
 & 45 & 3.537 & 38.34 & 38 & 3.379 & 37.90 & 38 & 3.285 & 37.72 & 38 \\
 & 60 & 3.508 & 47.62 & 50 & 3.621 & 46.40 & 50 & 3.277 & 46.32 & 50 \\
 & 90 & 3.363 & 52.38 & 88 & 3.246 & 55.81 & 88 & 3.254 & 56.52 & 88 \\
 & 120 & 3.363 & 56.41 & 88 & 3.246 & 62.42 & 88 & 3.254 & 61.68 & 88 \\
\midrule
256 & 10 & 3.463 & 9.37 & 12 & 3.645 & 9.45 & 12 & 3.467 & 9.38 & 12 \\
 & 20 & 3.506 & 18.16 & 12 & 3.604 & 18.10 & 12 & 3.314 & 18.15 & 12 \\
 & 30 & 3.506 & 26.90 & 12 & 3.619 & 26.90 & 12 & 3.409 & 26.90 & 12 \\
 & 60 & 3.506 & 53.23 & 12 & 3.572 & 53.34 & 12 & 3.484 & 53.19 & 12 \\
 & 120 & 3.506 & 105.86 & 12 & 3.572 & 105.02 & 38 & 3.469 & 105.72 & 19 \\
 & 240 & 3.664 & 174.49 & 50 & 3.494 & 169.22 & 50 & 3.438 & 174.14 & 50 \\
\bottomrule
\end{tabular}
}
\end{table}

Every returned policy is replayed by fresh native execution, and all
preserve the configurations completed by the initial policy.
All \AVal{certificates} runtime certificates, counted over original and
repeated measurements, equal the independent optimum established for
the same request and family in the scaling study.

The measurement machine is shared, and in the first run some worker
processes were suspended for long periods.
We call a measurement disturbed when its non-CPU time, the lifecycle
minus the CPU seconds of the single-threaded worker, exceeds one second
plus 1\% of the lifecycle.
The largest non-CPU time in the scaling study is 1.09 seconds, so the
rule marks none of its measurements.
It uses process timing only and was fixed after the first run, before
any re-measurement and before replayed results were available.
It marks \AVal{disturbed} of the \AVal{records} measurements
(\AVal{disturbed/partial_native} of CAPE,
\AVal{disturbed/lazy_andor_selective} of matched AND/OR and
\AVal{disturbed/eager_optimized} of eager exact).
These were re-measured with identical specifications in three rounds of
\AVal{round/1/measurements}, \AVal{round/2/measurements} and
\AVal{round/3/measurements} measurements, after which none was
disturbed.
The results above use the first undisturbed measurement of each job;
Table~\ref{tab:allowance-unfiltered} repeats Table~\ref{tab:allowance}
with the \AVal{records} original measurements.

\begin{table}[H]
\centering\small\setlength{\tabcolsep}{4pt}
\caption{Table~\ref{tab:allowance} computed from the original
measurements, without re-measurement of disturbed ones.}
\label{tab:allowance-unfiltered}
\fitinput{\begin{tabular}{rrrrrrrr}
\toprule
& & & \multicolumn{3}{c}{Completion (\%)} & \multicolumn{2}{c}{CAPE gain [95\% interval]} \\
\cmidrule(lr){4-6}\cmidrule(l){7-8}
$|H|$ & $\tau$ (s) & Initial & Eager & AND/OR & CAPE & vs.\ eager & vs.\ AND/OR \\
\midrule
64 & 5 & 64.84 & 83.20 & 84.38 & 82.81 & $-0.39$ [$-2.34$, $1.17$] & $-1.56$ [$-7.81$, $3.91$] \\
 & 10 & 64.84 & 85.16 & 92.97 & \textbf{95.31} & $10.16$ [$1.56$, $22.66$] & $2.34$ [$0.00$, $5.47$] \\
 & 15 & 64.84 & 82.81 & 95.31 & \textbf{100.00} & $17.19$ [$3.12$, $32.81$] & $4.69$ [$0.00$, $10.94$] \\
 & 20 & 64.84 & 95.31 & 94.53 & \textbf{100.00} & $4.69$ [$0.00$, $14.06$] & $5.47$ [$0.00$, $16.41$] \\
 & 30 & 64.84 & 100.00 & 98.44 & 96.88 & $-3.12$ [$-9.38$, $0.00$] & $-1.56$ [$-9.38$, $4.69$] \\
\midrule
128 & 5 & 53.91 & 55.86 & 57.42 & \textbf{58.01} & $2.15$ [$0.00$, $6.05$] & $0.59$ [$-2.93$, $4.69$] \\
 & 10 & 53.91 & 63.28 & 64.45 & \textbf{65.62} & $2.34$ [$0.00$, $7.03$] & $1.17$ [$-2.73$, $5.47$] \\
 & 15 & 53.91 & 66.41 & 70.70 & \textbf{75.00} & $8.59$ [$2.34$, $14.84$] & $4.30$ [$-1.37$, $11.91$] \\
 & 20 & 53.91 & 65.43 & 74.22 & \textbf{81.64} & $16.21$ [$5.66$, $26.76$] & $7.42$ [$1.95$, $12.89$] \\
 & 30 & 53.91 & 66.41 & 74.22 & \textbf{87.50} & $21.09$ [$5.86$, $37.50$] & $13.28$ [$3.52$, $23.83$] \\
 & 45 & 53.91 & 75.10 & 74.22 & \textbf{93.75} & $18.65$ [$6.15$, $32.13$] & $19.53$ [$5.47$, $34.38$] \\
 & 60 & 53.91 & 82.62 & 83.40 & \textbf{96.88} & $14.26$ [$1.76$, $31.45$] & $13.48$ [$1.76$, $31.45$] \\
 & 90 & 53.91 & 95.31 & 92.09 & \textbf{97.66} & $2.34$ [$0.00$, $7.03$] & $5.57$ [$0.00$, $14.94$] \\
 & 120 & 53.91 & 95.31 & 92.97 & \textbf{97.66} & $2.34$ [$0.00$, $7.03$] & $4.69$ [$0.00$, $14.06$] \\
\midrule
256 & 10 & 48.44 & 50.78 & 50.98 & 50.49 & $-0.29$ [$-1.46$, $0.59$] & $-0.49$ [$-1.76$, $0.29$] \\
 & 20 & 48.44 & 54.39 & 51.37 & \textbf{56.59} & $2.20$ [$-0.44$, $7.03$] & $5.22$ [$0.00$, $14.60$] \\
 & 30 & 48.44 & 56.35 & 58.59 & \textbf{62.21} & $5.86$ [$1.95$, $10.64$] & $3.61$ [$0.98$, $6.15$] \\
 & 60 & 48.44 & 57.03 & 63.67 & \textbf{66.99} & $9.96$ [$2.34$, $17.97$] & $3.32$ [$0.20$, $6.84$] \\
 & 120 & 48.44 & 57.03 & 63.67 & \textbf{69.53} & $12.50$ [$3.32$, $21.48$] & $5.86$ [$1.37$, $10.55$] \\
 & 240 & 48.44 & 67.77 & 63.67 & \textbf{73.05} & $5.27$ [$0.39$, $12.30$] & $9.38$ [$2.34$, $17.97$] \\
\bottomrule
\end{tabular}
}
\end{table}

\subsection{Independent validation}

Every executed action is checked against its native effect model,
including observations, commit status, publication receipts and output
content. Model decisions are checked against the stored request history
and raw response; warm executions are checked against the cached
successful program. The new execution comparison contains 80,756 native
calls across cold episodes, repeated execution and full-family retry
evaluation.

Complete-table optimization resolves \SVal{e2/resolved} of the 40
scaling-study objectives. The four remaining 256-configuration
projections retain feasible lower and optimistic upper values; their
returned policies carry no runtime certificate.
All \SVal{e2/certificates} positive runtime certificates are independently
verified, and every returned policy passes native replay and initial
coverage checks.

\end{document}